\documentclass[10pt,twocolumn,letterpaper]{article}

\usepackage[pagenumbers]{cvpr} 

\definecolor{shadecolor}{RGB}{255,255,200}
\definecolor{promptcolor}{RGB}{216, 238, 255}

\newcommand{\qwen}{{Qwen2.5-Omni}}
\newcommand{\vllama}{{VideoLLaMA2}}
\newcommand{\audioset}{{AudioSet}}
\newcommand{\qwenSB}{{Qwen2.5-Omni-7B}}
\newcommand{\panda}{PandaGPT}
\newcommand{\gemini}{Gemini-2.0-Flash-Lite}
\newcommand{\geminipro}{Gemini-2.5-Pro}
\newcommand{\geminiflash}{Gemini-2.0-Flash}
\newcommand{\qwenThree}{Qwen3-Omni-30B}
\newcommand{\chatbridge}{ChatBridge}
\newcommand{\bench}{{MMA-Bench}}
\newcommand{\cohend}{{Cohen's-D}}
\usepackage{pifont}
\usepackage{multirow}
\usepackage[normalem]{ulem}
\usepackage{tcolorbox}

\definecolor{cvprblue}{rgb}{0.21,0.49,0.74}
\usepackage[pagebackref,breaklinks,colorlinks,allcolors=cvprblue]{hyperref}
\usepackage{multirow}
\usepackage[table]{xcolor}
\usepackage{array}
\usepackage{float}

\def\paperID{20283} 
\def\confName{CVPR}
\def\confYear{2026}

\title{\textit{Some Modalities are More Equal Than Others}: Decoding and Architecting Multimodal Integration in MLLMs}

\author{
Tianle Chen\textsuperscript{1*} \quad
Chaitanya Chakka\textsuperscript{1*} \quad
Arjun Reddy Akula\textsuperscript{2} \quad
Xavier Thomas\textsuperscript{1} \quad
Deepti Ghadiyaram\textsuperscript{1} \\
\textsuperscript{1}Boston University \quad
\textsuperscript{2}Google DeepMind \\
{\tt\small tianle@bu.edu} \quad
{\tt\small chvskch@bu.edu} \quad
{\tt\small arjunakula@google.com} \\
{\tt\small xthomas@bu.edu} \quad
{\tt\small dghadiya@bu.edu} \\
\small{*Equal contribution}
}

\begin{document}
\twocolumn[{
\maketitle
\begin{center}
\newcommand{\teaserwidth}{0.80\textwidth}
    \includegraphics[width=\teaserwidth]{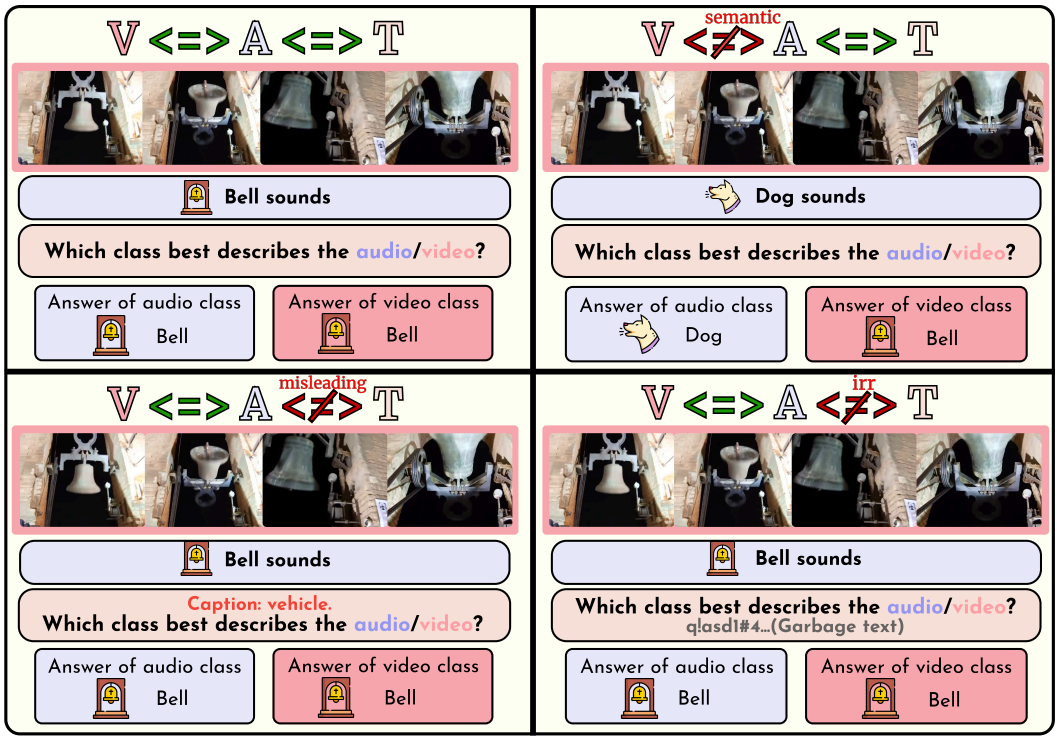}
   \vspace{-2mm}
    \captionof{figure}{We propose 
    \textbf{MMA-Bench to expose how MLLMs behave when sight, sound, and language conflict.}
    Each example presents a controlled modality (e.g., audio, video, or text) conflict and asks two modality-specific questions - one about the video and one about the audio.
    Correct answers differ across modalities, forcing the model to attend to the reliable modality.
    These structured contradictions reveal if MLLMs are truly multi-modal or take shortcuts during cross-modal reasoning tasks.
    }
    \label{fig:pipeline}
\end{center}
}]

\begin{abstract}
Despite remarkable advancements in Multimodal Large Language Models (MLLMs), a fundamental question remains: \emph{are MLLMs robust to contradicting modalities?} To rigorously study this, we introduce {\textbf{\bench}} comprising videos and tasks that probe a model's reliance on specific modalities. Using black-box and white-box interpretability techniques, we provide a critical analysis of the brittleness of both open- and closed-sourced MLLMs. We show that current MLLMs struggle under misaligned audio-visual pairs and simple misleading text, thereby lacking robust multi-modal reasoning. Building on these findings, we propose a modality alignment tuning strategy to teach the model when to prioritize, leverage, or ignore specific modality cues. Through extensive experiments and analysis, we show that our alignment tuning yields demonstrably stronger multimodal grounding. This work provides both interpretability tools and a clear path toward developing MLLMs with intrinsically reliable cross-modal reasoning.
Code and dataset will be available at \href{https://cskyl.github.io/MMA-Bench/}{cskyl.github.io/MMA-Bench/}
\end{abstract}
    
\section{Introduction}
\label{sec:intro}
Consider this scenario: a video of birds chirping. If blindfolded, could you describe what you hear? Similarly, if wearing noise-canceling headphones, could you describe what you see? Now, envision a third case: you experience the video with both sight and sound, but there is distracting and incoherent text descriptions appearing alongside it. In all these situations, most humans, with some effort, can easily describe the events in the video, ignore the distractors, and rely on the available modality. But what about present-day Multimodal Large Language Models (MLLMs) 
~\cite{cheng2024videollama, comanici2025gemini, su2023pandagpt,Qwen2.5-Omni, girdhar2023imagebind}? We investigate this question in our work.

Despite rapid advancements in the design and deployment of large-scale MLLMs, most models~\cite{girdhar2023imagebind, hamilton2024separating, radevski2025dave, sung2024avhbench, chowdhury2025avtrustbench} are trained on datasets that overwhelmingly assume that all available modalities are aligned. This may inadvertently 
make these models struggle when one modality is missing, noisy, or conflicts with another. This in turn makes these models susceptible to simple text-, vision-, and audio-based data poisoning or prompt attacks. Such attacks may coerce an MLLM to generate harmful content, leak sensitive information, or perform unauthorized actions. Given the rapid deployment of MLLMs in robotics~\cite{yang2025magma, zhao2025vlas} and medical diagnostics~\cite{liu2023medical, huang2025towards}, such brittleness may lead to unacceptable risks and unsafe outcomes. 

In our work, we approach this challenging problem from the ground up. First, we aim to understand how current day MLLMs reason when multiple modalities are missing or semantically misaligned. To this end, we propose \textbf{M}ulti-\textbf{M}odal \textbf{A}lignment (MMA)-Bench, to systematically control the presence or alignment of one modality (video, audio, text) at a time. Next, we employ a suite of black-box and white-box techniques to understand MLLMs behavior under these control settings. Our analysis revealed a critical flaw: despite being trained as multimodal systems, \textbf{all MLLMs over-rely on text and collapse when modalities conflict.} As the famous George Orwell's quote goes -- \textit{all animals are equal but some animals are more equal than others} -- we indeed find that though present day MLLMs claim to process all modalities equally, some modalities (like text then visual) are more heavily utilized than others.

Based on these findings, we propose a simple modality-aware tuning method to instill modality grounding. Our fine-tuned models show stronger cross-modal grounding, interpretable attention redistribution, and measurable performance gains on both in-domain and out-of-domain benchmarks. We make the following contributions:
\begin{itemize}
    \item We introduce \textbf{MMA-Bench}, a comprehensive benchmark to probe resilience to modality perturbations of MLLMs. 
    \item We provide \textbf{black-box and white-box diagnostic framework} that reveals why current MLLMs fail under conflicting multimodal inputs.
    \item We propose a \textbf{modality-aware fine-tuning} that substantially improves cross-modal grounding.
\end{itemize}

\section{Related Work}
\paragraph{Audio–Visual Evaluation Benchmarks.}
Prior benchmarks have explored robustness and reasoning in audio–visual large language models (LLMs), yet most evaluate whether the two modalities cooperate correctly rather than whether they can be \emph{selectively} used when cues conflict.  
AVTrustBench~\cite{chowdhury2025avtrustbench} and AVHBench~\cite{sung2024avhbench} focus on global visual-audio consistency and hallucination detection, while fine-grained datasets such as VGGSounder~\cite{zverev2025vggsounder}, DAVE~\cite{radevski2025dave}, and AURA~\cite{samakoush2025aura} assess cross-modal robustness but still assume semantic alignment between sight and sound.  
Other suites (e.g., DAVE, AURA, OmniVideoBench~\cite{li2025omnivideobench}) provide compositional or logical reasoning tests but seldom require distinct answers for visual and auditory content.  
In contrast, \textbf{MMA-Bench} isolates modality sensitivity by pairing each video with a visual- and an audio-focused question whose correct answers may diverge under semantic conflict.  
This design enables disentangled diagnosis of visual and auditory reasoning especially when modalities pose conflicting information while preserving natural multimodal context.  

\noindent \textbf{Mitigating Modality Bias and Misalignment.}
Efforts to reduce modality bias fall broadly into \emph{training-free} and \emph{training-based} approaches.  
Training-free steering methods (e.g., AutoSteer~\cite{wu2025automating}, MC$^2$~\cite{zhang2025evaluating}) adjust attention or residual activations at inference to balance modality preference, whereas training-based strategies impose auxiliary objectives such as temporal ordering or cross-modal consistency (Arrow-of-Time~\cite{xue2025seeing}).  
Complementary decoding schemes like AVCD~\cite{jung2025avcd} and Fork-Merge~\cite{jung2025fork} mitigate hallucination without retraining, and distillation frameworks such as Bridging Ears and Eyes~\cite{jiang2025bridging} align encoder representations across sensory domains.  
Unlike these approaches, our \textbf{modality-aware supervised fine-tuning} directly teaches models to reason selectively using both semantically aligned and deliberately misaligned video–audio pairs.  
This prompt-conditioned supervision yields adaptive attention redistribution toward the queried modality rather than static bias suppression.

\section{Multi-Modal Alignment (MMA) Bench: Debugging Alignment Gaps in MLLMs} \label{sec:mmabench} 
\begin{figure*}[t]
    \centering
    \includegraphics[width=0.7\textwidth]{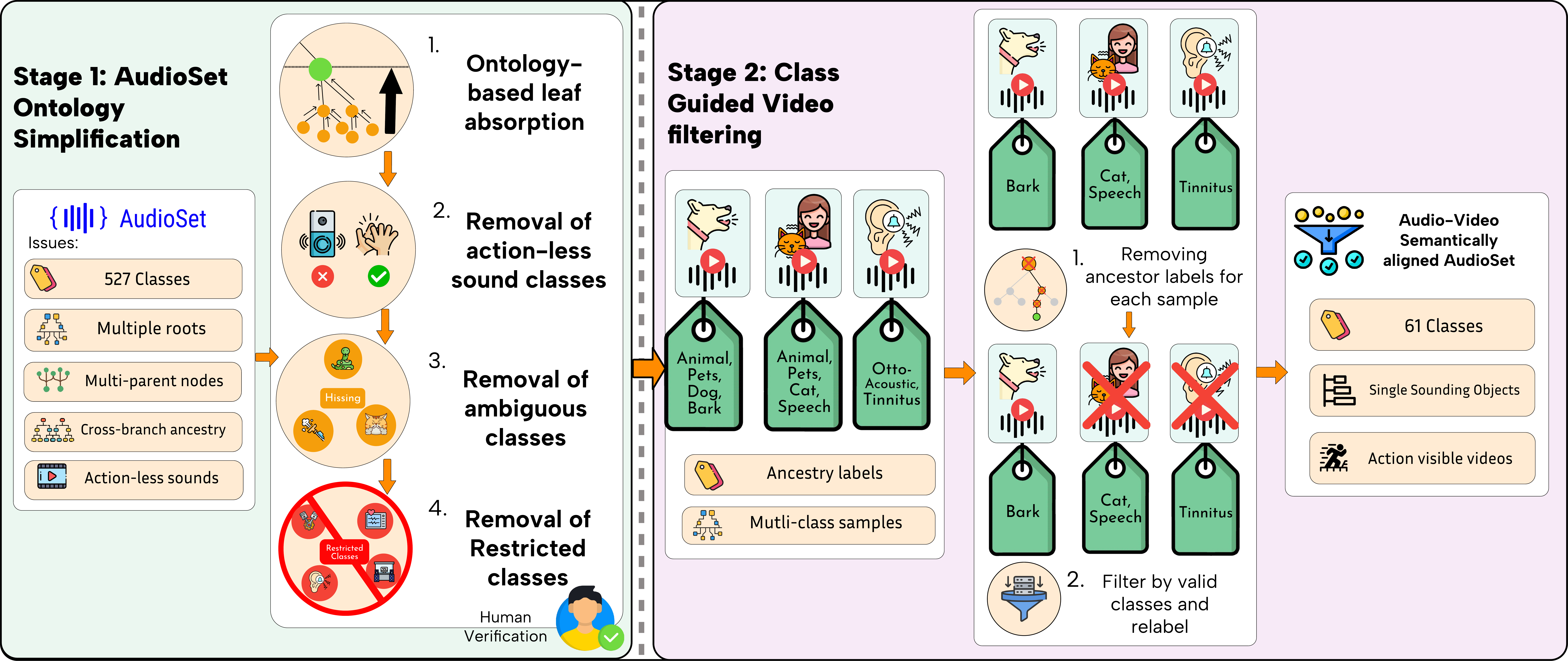}
    \vspace{-2mm}
    \caption{
    \textbf{Automated data curation pipeline for building \bench.}
    Our two–stage pipeline converts raw AudioSet~\cite{gemmeke2017audio} into clean, semantically aligned audio–video samples. 
    \textbf{Stage 1} simplifies the ontology by pruning action-less, ambiguous (e.g., audio event ``hiss" could be associated with ``stream" or ``cat" ), and restricted classes (e.g.,``heart murmur"). 
    \textbf{Stage 2} retains videos based on the simplified audio events. Here, only clips where the audible event is clearly produced by a visible object are retained yielding a high-quality subset which is further post-processed (Sec.~\ref{sec:mmabench}).
    }
    \label{fig:data_curationpipeline}
    \vspace{-4mm}
\end{figure*}
\begin{figure}[t]
    \centering
    \includegraphics[width=0.48\textwidth]{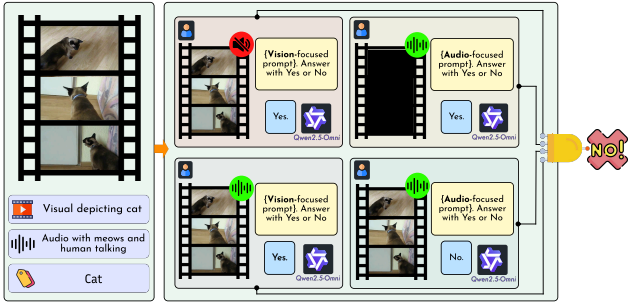}
    \vspace{-4mm}
    \caption{
    \textbf{Verification of audio-visual semantic alignment}. After automated pruning (Fig.~\ref{fig:data_curationpipeline}), each clip undergoes $4$ simple yes/no consistency checks.  
    A sample is kept only if it passes all checks.}
    \label{fig:task_pipeline}
    \vspace{-4mm}
\end{figure}
This work focuses on a critical question: \textit{how do current day MLLMs behave when multiple modalities are semantically misaligned?}
Existing multimodal benchmarks emphasize tasks such as holistic visual-audio judgment~\cite{chowdhury2025avtrustbench} and cross-modal hallucination detection~\cite{sung2024avhbench}, which primarily evaluate whether models can detect or maintain overall visual-audio consistency. 
These datasets only provide binary alignment labels (e.g., ``aligned'' or ``misaligned'') rather than modality-specific semantic annotation for video and audio separately.
By contrast, we wish to probe modality-specific reasoning of MLLMs by asking visual- and audio-focused questions separately (illustrated in Fig.~\ref{fig:pipeline}) 
To this end, we propose \textbf{M}ulti-\textbf{M}odal \textbf{A}lignment (MMA) Bench, the first dedicated benchmark to assess modality sensitivity of MLLMs. We outline the following desiderata while designing MMA-Bench:
\begin{tcolorbox}[
    colback=shadecolor!10,    
    colframe=yellow!45,          
    coltitle=black,           
    title=\textbf{\footnotesize{Selection Criteria for Aligned Samples}}, 
    fonttitle=\bfseries,
    left=3mm, right=3mm, top=1mm, bottom=1mm,
    boxsep=1mm,
    arc=3pt,                  
    outer arc=3pt,
    boxrule=0.8pt             
]
\begin{itemize}[leftmargin=1.3em]
\footnotesize
    \item Sounding object must be clearly visible in the video.
    \item Audio track must correspond to the visible object in the scene.
    \item Video must not contain distracting or composite audio sources.
\end{itemize}
\vspace{-1mm}
\label{fig:sel_creiteria_MMA}
\end{tcolorbox}


\noindent \textbf{Data curation:} We construct MMA-Bench starting from the \textit{evaluation split} of {\audioset}~\cite{gemmeke2017audio}, which contains $20{,}371$ YouTube videos (typically $10$s long) annotated with $527$ audio event classes. However, as noted in \cite{zverev2025vggsounder,dogan2024multi,xie2021zero,xu2024enhancing}, {\audioset} labels are ambiguous and noisy. Critically, it suffers from the following issues: (a) \textbf{missing objects:} many audio events (e.g., ``alarm'') lack a visible object, (b) \textbf{ontology issue:} the ontology of the audio events is excessively fine-grained (e.g., ``crying" is further split into ``whimper," ``baby cry,'' ``infant cry"), and (c) \textbf{label noise:} composite scenes contain imprecise labels (e.g., a video is labeled ``violin" when other instruments are being played). (d) \textbf{multi-label ambiguity:} each clip is annotated with multiple hierarchical labels, many of which do not correspond to a single clear audio–visual event.

We address these issues by designing a rigorous, multi-stage filtering pipeline combining ontology pruning~\cite{tuncay2025hierarchical}
, automated cross-modal verification, and human inspection. 
Specifically, we first remove abstract or overly fine-grained audio class nodes (and their associated videos, since many such fine-grained distinctions cannot be visually grounded, and merging them into parent labels would introduce incorrect audio–visual alignment
) from the eval split of the {\audioset} ontology containing $20,371$ samples in a $2$ stage pipeline as summarized in Fig.~\ref{fig:data_curationpipeline}. 
Next, we filter videos with multiple audio events and with missing sounding objects by passing through {\qwen} with a rigid $4$ question checks as shown in Fig.~\ref{fig:task_pipeline}. After these refinements,
we go through the filtered samples manually to ensure high data quality of aligned samples, resulting in $658$ single-label aligned videos 
addressing the multi-label ambiguity in AudioSet
\begin{table}[t]
\centering
\caption{\textbf{Dataset statistics across all creation stages.}
The table shows the number of training and evaluation samples retained after each curation step, concluding with the final dataset including misaligned counterparts.}
\label{tab:mma_dataset_stats}
\vspace{1mm}
\resizebox{0.45\textwidth}{!}{
\begin{tabular}{lcc}
\toprule
\textbf{Stage} & \textbf{Train} & \textbf{Eval} \\
\midrule
Original AudioSet split & 22,160 & 20,371 \\
After ontology pruning \& downloading & 3892 & 3518 \\
After automated filtering & 1,207 & 1,102 \\
After human verification & 1,207 & 658 \\
Final dataset (with misaligned samples) & 13,277 & 1,316 \\
\bottomrule
\end{tabular}
}
\end{table}

To generate the misaligned samples, we start with an aligned sample (e.g., a church bell video paired with the church bells ringing audio) and we randomly swap the audio track with a different audio event (e.g., dog barking). This procedure creates one misaligned counterpart for every aligned sample, yielding an equal number of aligned and misaligned examples. The final benchmark thus contains \textbf{$1,316$} videos in total -- half aligned and half misaligned -- spanning $49$ representative sound categories.
A summary of all  sizes at each curation step is provided in Table~\ref{tab:mma_dataset_stats}. 

This single-labeled, semantically aligned set forms the foundation for all subsequent fine-tuning and interpretability experiments in our work. (Details in Appendix A.)

\noindent \textbf{Tasks:} Every video, whether the audio and visual streams are aligned or misaligned, in our benchmark is associated with two question-answer pairs. 
\begin{itemize}
    \item \textbf{Visual-object classification:} Which class best describes the visual content of this video?
    \item \textbf{Audio-event classification:} Which class best describes the sound in this video?
\end{itemize}
These paired tasks probe each modality independently: the same video is queried twice, once for visual reasoning and once for auditory reasoning, allowing us to directly assess whether models can attend to the correct modality rather than relying on cues.
Using MMA-Bench, we next analyze how current MLLMs behave under these controlled settings.

\section{Are modern MLLMs Truly ``Multi-modal''?} 

\begin{table}[t]
\scriptsize
\centering
\setlength{\tabcolsep}{2pt}
\renewcommand{\arraystretch}{1.2}
\begin{tabular}{@{}p{4.4cm}p{4cm}@{}}
\toprule
\textbf{Misalignment Type} & \textbf{Example} \\
\midrule
\textcolor{ForestGreen}{Video $ \Leftrightarrow$ Audio $\Leftrightarrow$ Text} (Baseline) & Church bell video paired with church bell ringing audio. \\

\textcolor{red}{Video $ \overset{\text{Semantic}}\nLeftrightarrow$ Audio} \textcolor{ForestGreen}{$\Leftrightarrow$ Text} & Church bell video paired with dog barking audio. \\


\textcolor{ForestGreen}{Video $\Leftrightarrow$ Audio} \textcolor{red}{$\overset{\text{Irrelevant caption}}\nLeftrightarrow$ Text} & Caption describes a dog while video shows a church bell. \\

\textcolor{ForestGreen}{Video $\Leftrightarrow$ Audio} \textcolor{red}{$\overset{\text{Long context}}\nLeftrightarrow$ Text} & Long irrelevant paragraph appended before the question. \\

\textcolor{red}{Video $\overset{\text{Frames Zeroed}}\nLeftrightarrow$ \textcolor{ForestGreen}{Audio} } & \textit{Eyes closed:} all visual frames replaced with black frames. \\

\textcolor{ForestGreen}{Video} \textcolor{red}{$\overset{\text{Audio removed}}\nLeftrightarrow$ Audio} & \textit{Ears shut:} original audio track replaced with silence. \\

\bottomrule
\end{tabular}
\vspace{-0.1in}
\caption{
\textbf{Controlled modality misalignment and ablation scenarios} studied in Sec.~\ref{sec:analysis}.  
We selectively perturb or remove one modality while keeping the others unchanged to isolate specific sources of multimodal brittleness in MLLMs.
}
\label{tab:input_types_studied}
\vspace{-4mm}
\end{table}

\label{sec:analysis}
Our key goal is to systematically understand how MLLMs intrinsically integrate visual, text, and audio representations. To this end, we employ a) \textbf{black-box probing}, where we provide different kinds of inputs and study model performance, and b) \textbf{white-box probing}, where we closely inspect cross-attention interactions of several intermediate attention layers and heads to understand the interplay of multiple modalities. To systematically understand how modalities interact, we study various controlled misalignment input scenarios summarized in Table~\ref{tab:input_types_studied} and discuss next.
\subsection{Black-box Interpretability and Findings}
\label{sec:blackbox}

\subsubsection{Experiment setup} \label{sec:3.1.1}
We use the MMA-Bench constructed in Sec~\ref{sec:mmabench} for the black-box analysis, unless specified otherwise.

\begin{figure}[t]
    \centering

    \begin{subfigure}[t]{0.48\linewidth}
        \centering
        \includegraphics[width=\linewidth]{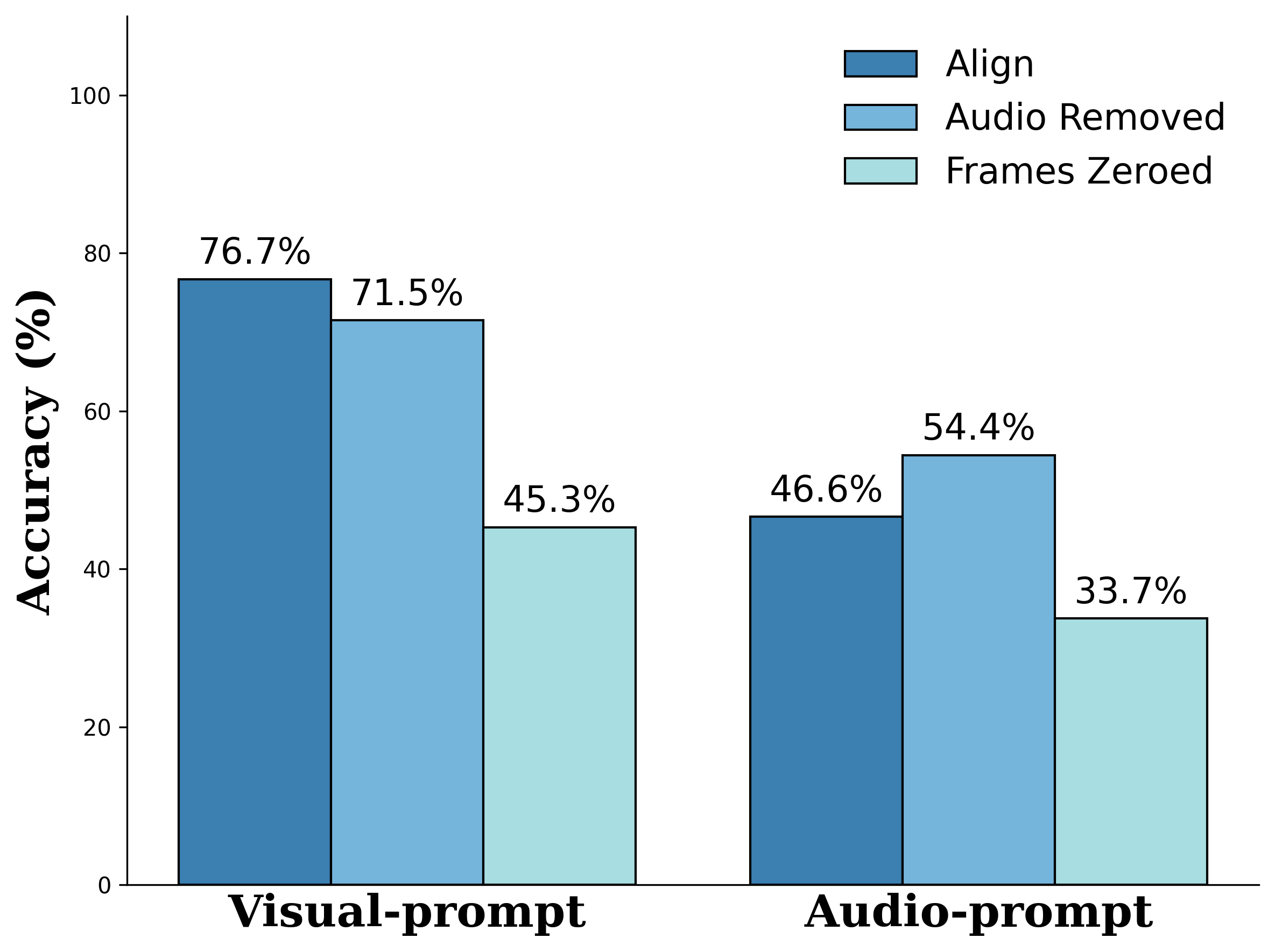}
        \caption{\qwenSB.}
        \label{fig:unimodal_qwen}
    \end{subfigure}
    \hfill
    \begin{subfigure}[t]{0.48\linewidth}
        \centering
        \includegraphics[width=\linewidth]{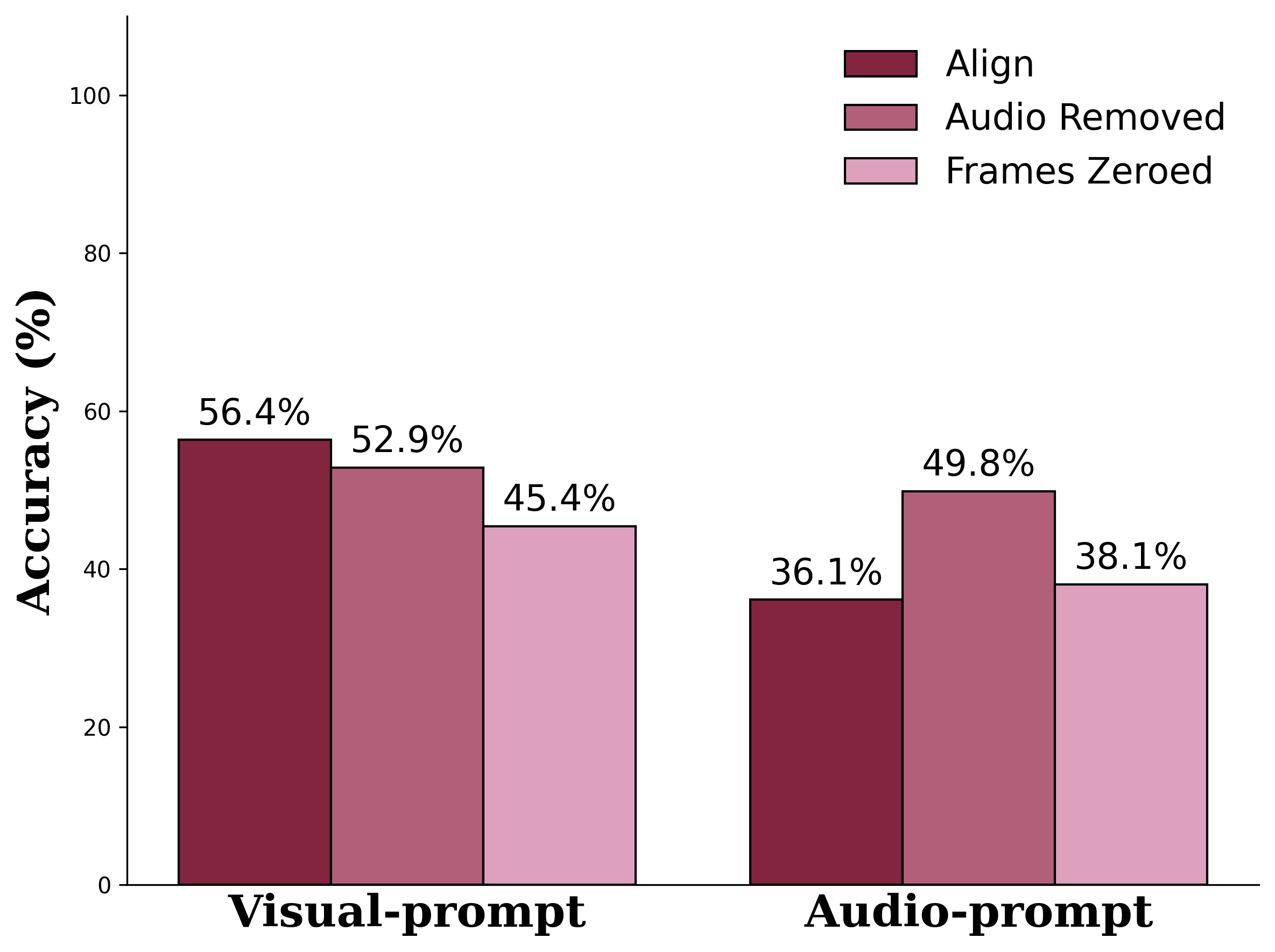}
        \caption{VideoLLaMA2.}
        \label{fig:unimodal_vl2}
    \end{subfigure}

    \vspace{1mm}
    \begin{subfigure}[t]{0.48\linewidth}
        \centering
        \includegraphics[width=\linewidth]{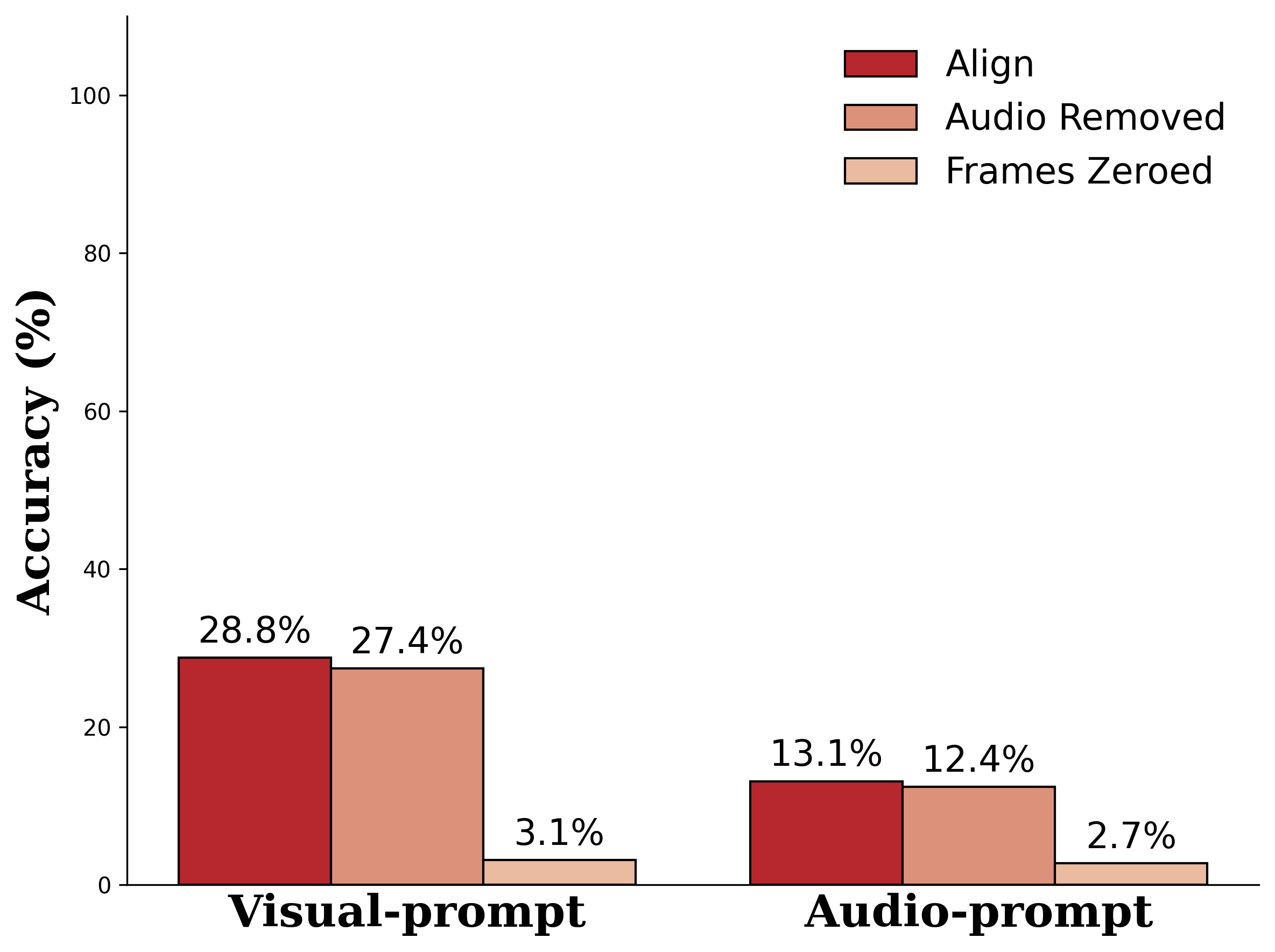}
        \caption{PandaGPT.}
        \label{fig:unimodal_pandagpt}
    \end{subfigure}
    \hfill
    \begin{subfigure}[t]{0.48\linewidth}
        \centering
        \includegraphics[width=\linewidth]{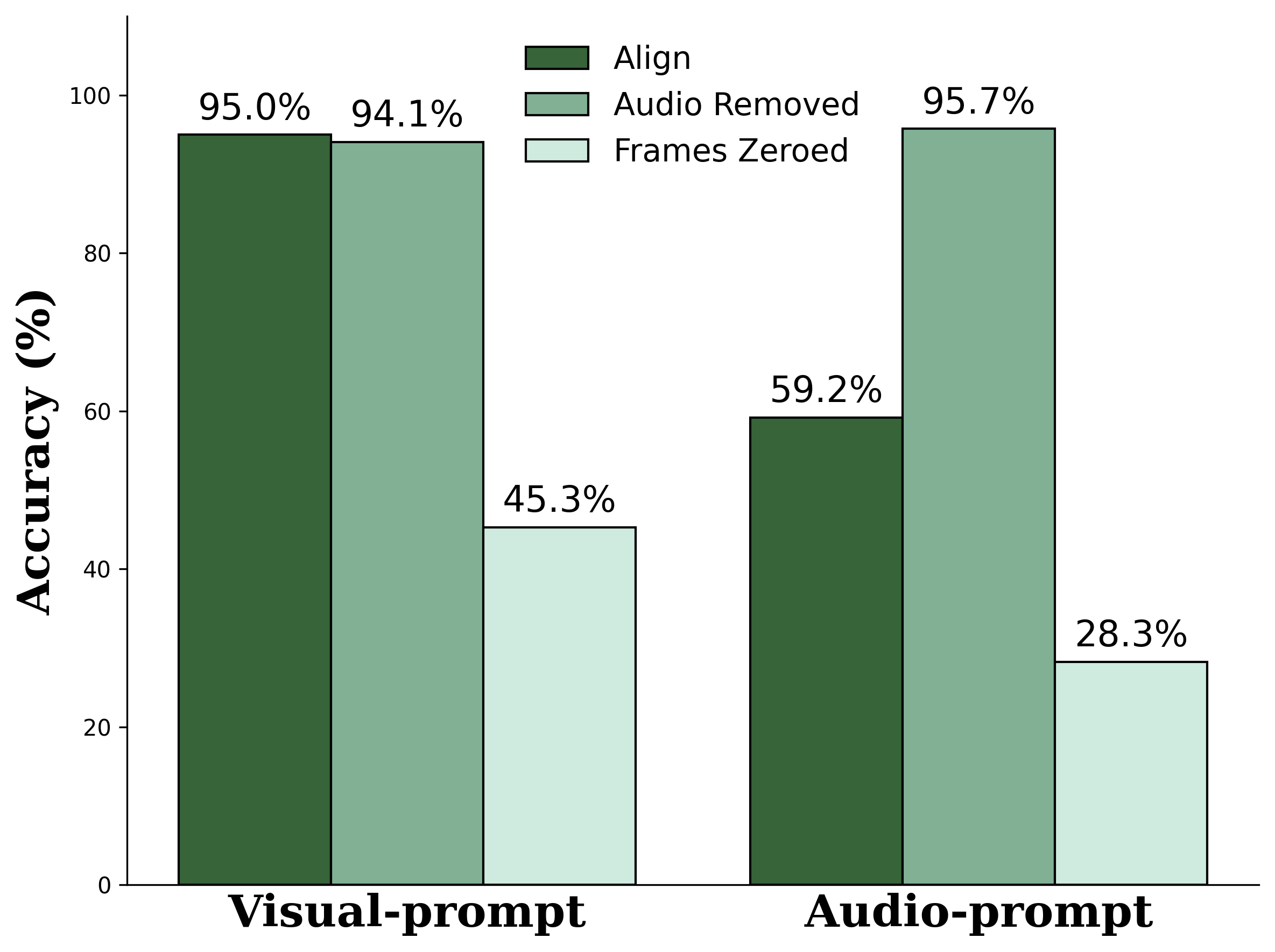}
        \caption{\gemini.}
        \label{fig:unimodal_gemini}
    \end{subfigure}

    \vspace{1mm}
    \caption{
   \textbf{Unimodal probing under visual and auditory ablation}. Each subplot reports classification accuracy under \textbf{visual-focused} (left bars) and \textbf{audio-focused} (right bars) prompts.  
    ``Audio removed'' replaces the sound track with silence, while ``Frames zeroed'' replaces all video frames with black images.  
    }
    \label{fig:unimodal_bias}
    \vspace{-4mm}
\end{figure}

\begin{figure}[t]
    \centering
    \fcolorbox{purple}{promptcolor!10}{
    \parbox{\dimexpr0.9\linewidth-2\fboxsep-2\fboxrule}{
        \scriptsize 
        \noindent\textbf{Visual-focused prompt:} 
        \vspace{-2mm}
        \begin{quote}
         \texttt{"Which class best describes the \emph{visual content} of this video? Options: \{Accordion, Bird, Cat,...\}. Answer using a single word or phrase."}
        \end{quote}
        \vspace{-1mm}
        \noindent\textbf{Audio-focused prompt:} 
        \vspace{-2mm}
        \begin{quote}
         \texttt{"Which class best describes the \emph{audio content} of this video? Options: \{Accordion, Bird, Cat,...\}. Answer using a single word or phrase."}
        \end{quote}
        \vspace{-1mm}
    }}
    \vspace{-2mm}
    \caption{
    \textbf{Visual- and audio-focused prompts used for understanding modality sensitivity of MLLMs .}
    Each prompt encourages the model to focus on one modality and allows a controlled comparison of visual and auditory reasoning within MLLMs.}
    \label{fig:vaprompts}
    \vspace{-4mm}
\end{figure}
 
\noindent \textbf{Unimodal data:} To examine the contribution of each modality for overall reasoning, we construct two unimodal variants of each video: (\textit{i}) \textbf{eyes closed}, where we fill all frames with zeros, resulting in all black frames.
while retaining the original audio, and (\textit{ii}) \textbf{ears shut}, where we strip the audio while retaining the original visual frames of the video. 

\noindent \textbf{Models studied:} We study $3$ open-sourced models: VideoLLaMA2~\cite{cheng2024videollama}, PandaGPT~\cite{su2023pandagpt}, and Qwen2.5-Omni-7B~\cite{Qwen2.5-Omni} and one closed-sourced model: Gemini-Flash-2.0-Lite~\cite{comanici2025gemini}.
Results are averaged across $3$ random seeds.

\noindent \textbf{Prompts:} We use the same prompts detailed in MMA-Bench in Sec~\ref{sec:mmabench} and append different class labels as options to create a classification task, as shown in Fig~\ref{fig:vaprompts}.
Each prompt targets a single modality while the model still receives both modalities as input unless explicitly ablated in the following analyses.


\subsubsection{Performance when one modality is removed} \label{sec:unimodal}

\begin{figure}[t]
    \centering
    \begin{subfigure}[t]{0.48\linewidth}
        \centering
        \includegraphics[width=\linewidth]{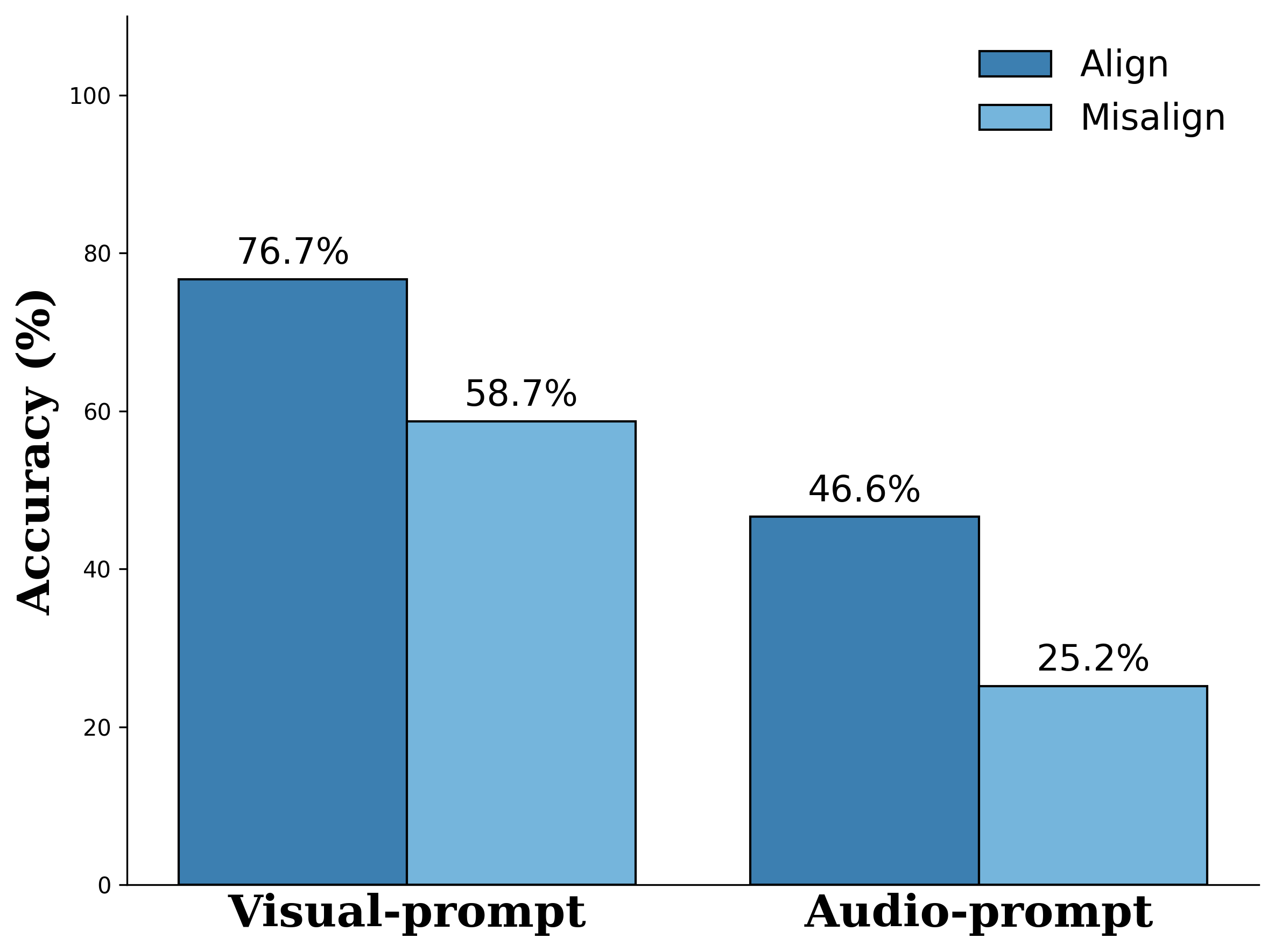}
        \caption{\qwenSB.}
        \label{fig:semantic_misalign_qwen}
    \end{subfigure}
    \hfill
    \begin{subfigure}[t]{0.48\linewidth}
        \centering
        \includegraphics[width=\linewidth]{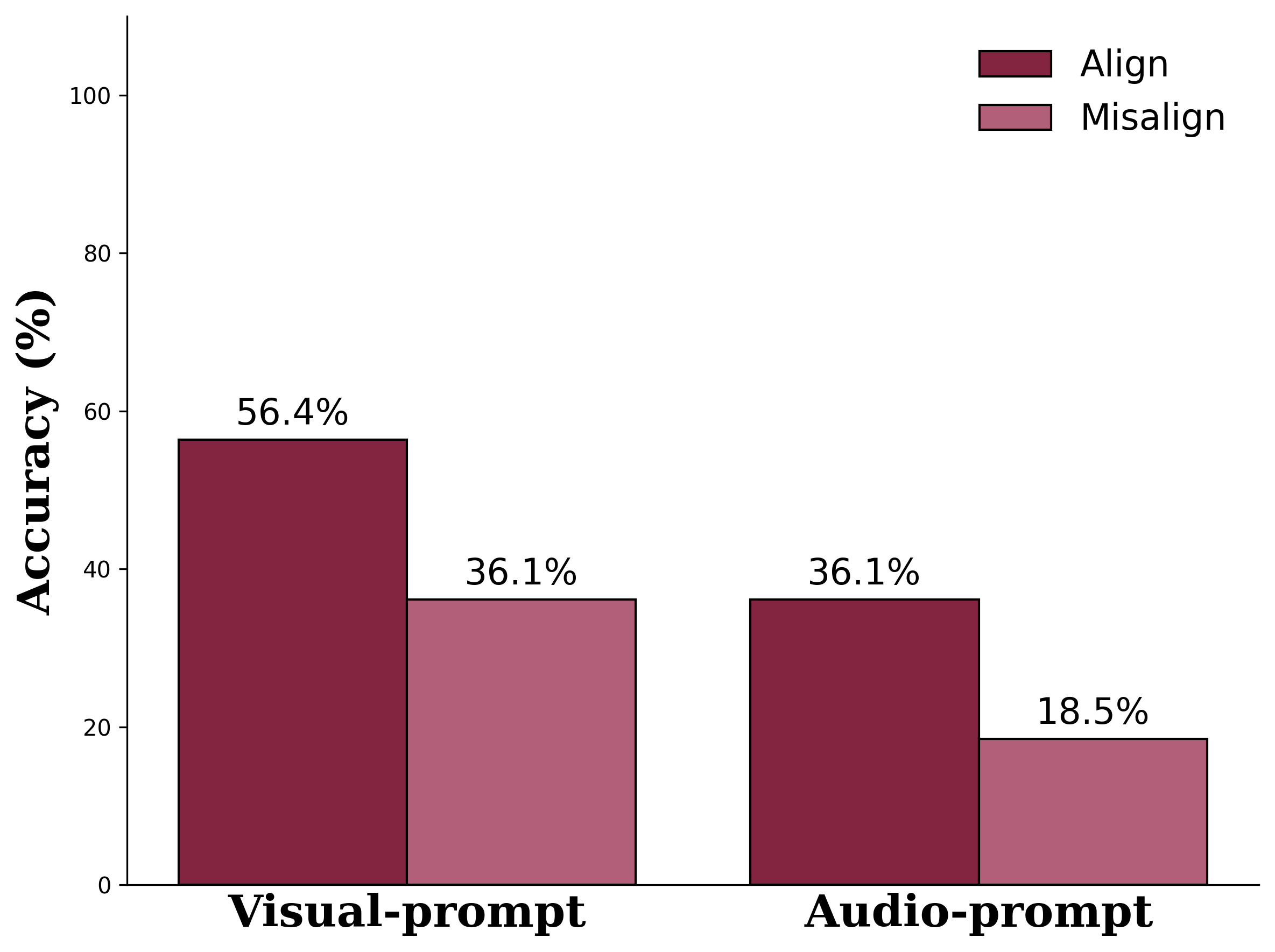}
        \caption{\vllama.}
        \label{fig:semantic_misalign_videollama}
    \end{subfigure}

    \vspace{2mm}
    \begin{subfigure}[t]{0.48\linewidth}
        \centering
        \includegraphics[width=\linewidth]{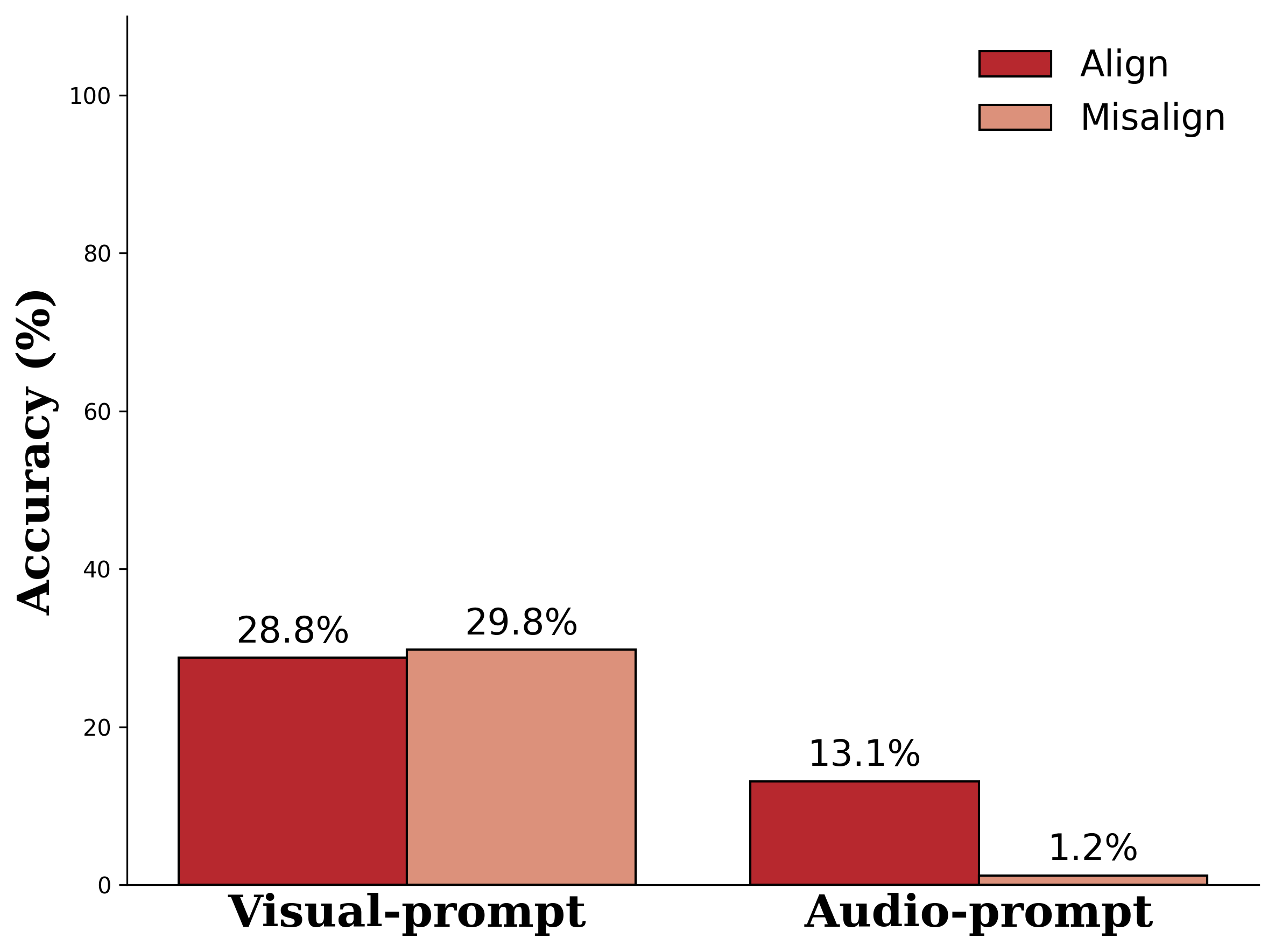}
        \caption{PandaGPT.}
        \label{fig:semantic_misalign_pandagpt}
    \end{subfigure}
    \hfill
    \begin{subfigure}[t]{0.48\linewidth}
        \centering
        \includegraphics[width=\linewidth]{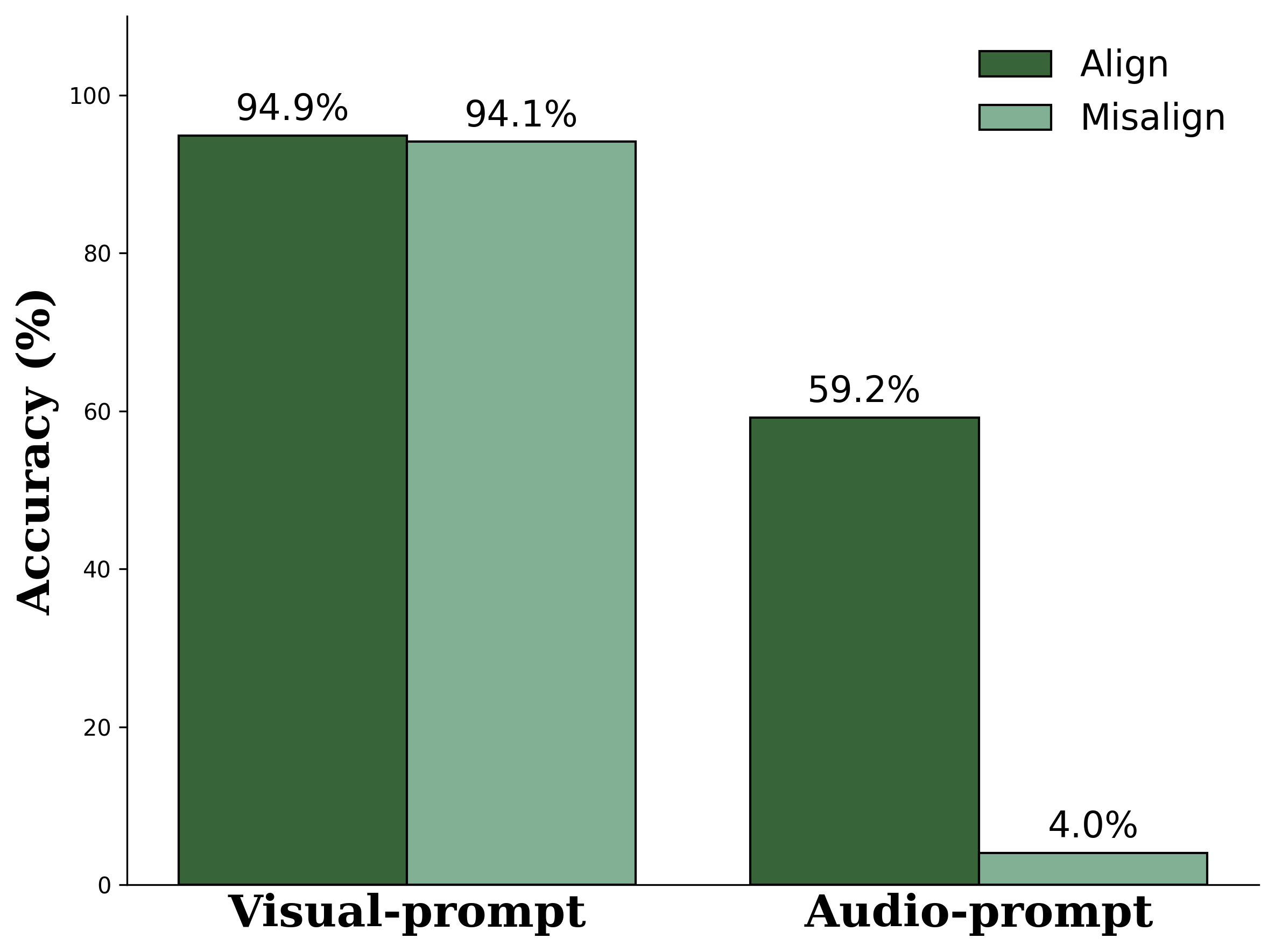}
        \caption{\gemini.}
        \label{fig:semantic_misalign_gemini}
    \end{subfigure}

    \vspace{1mm}
    \caption{
    \textbf{Performance of semantic misaligned} (\textit{Misalign}) and aligned (\textit{Baseline}) video–audio pairs under \textbf{visual-only} (left bar group) and \textbf{audio-only} (right bar group) prompts. 
    }
    \label{fig:semantic_misalignment}
    \vspace{-6mm}
\end{figure}
\noindent \textbf{What do models \textit{see} with ears shut?}  
To assess visual reasoning without auditory input, we replace each clip’s audio with silence and evaluate models using visual-focused prompts in Fig. ~\ref{fig:vaprompts}. 
This corresponds to the \emph{middle bars} (``Audio Removed'') in the visual-prompt columns of Fig.~\ref{fig:unimodal_bias}.
All models retain comparable performance but show a consistent slight drop, indicating that audio offers modest complementary cues.  
{\qwenSB} and {\vllama} exhibit larger declines (4–5\%), suggesting stronger cross-modal use of sound, whereas {\gemini} and {\panda} degrade minimally ($\sim$1\%), relying more on visual content.   

Evaluating with audio-focused prompts under the same silent condition is an inherently ill-posed setting. All models still achieve non-trivial accuracy,
for instance, {\qwenSB} reaches 46.6\% and {\gemini} attains 59.2\% despite receiving no audio at all--by inferring their answers purely from visual cues.
This indicates that when the requested modality is absent, models do not reliably follow the desired modality instructions and instead fall back on whichever modality remains informative--typically the visual stream, (as further analyzed in Sec.~\ref{sec:3.1.3})
, which dominates their learned representations.

\noindent \textbf{What do models \textit{hear} with eyes closed?}
We next test auditory reasoning by zeroing out visual frames while keeping the original audio intact (Fig.~\ref{fig:unimodal_bias}, see the \textit{rightmost bars} in each subplot labeled ``Frames Zeroed'').  
Using audio-focused prompts in Fig. ~\ref{fig:vaprompts}, most models exhibit a significant accuracy drop under this condition, indicating strong dependence on visual context for auditory understanding.  
{\vllama}~\cite{cheng2024videollama}, however, shows a slight improvement, likely benefiting from its audio-only pretraining that promotes more independent audio processing.  
When the task is reversed—using visual-focused prompts with no visual input—the setting becomes ill-posed.  
All models except {\panda} still return high accuracy by inferring the visual answer from audio cues, suggesting that they rely on residual cross-modal correlations rather than abstaining when the requested evidence is absent. 
{\panda} does not exhibit this behavior and instead fails to generate coherent outputs under such out-of-distribution conditions.

\subsubsection{\textcolor{red}{Video $ \overset{\text{Semantic}}\nLeftrightarrow$ Audio} \textcolor{ForestGreen}{$\Leftrightarrow$ Text}} \label{sec:3.1.3}

\begin{figure}[t]
    \centering

    \begin{subfigure}[t]{0.48\linewidth}
        \centering
        \includegraphics[width=\linewidth]{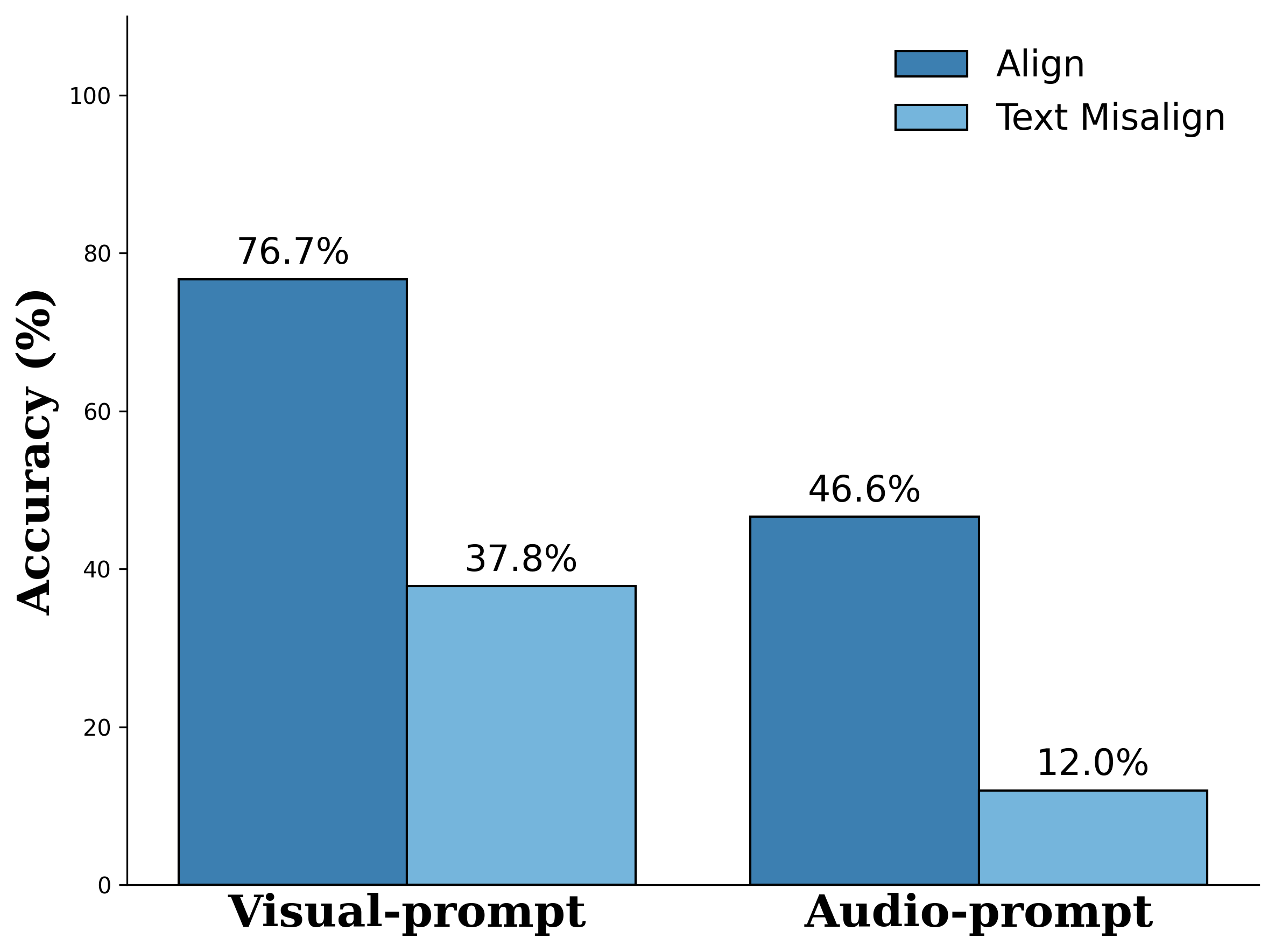}
        \caption{\qwenSB.}
        \label{fig:textmis_qwen}
    \end{subfigure}
    \hfill
    \begin{subfigure}[t]{0.48\linewidth}
        \centering
        \includegraphics[width=\linewidth]{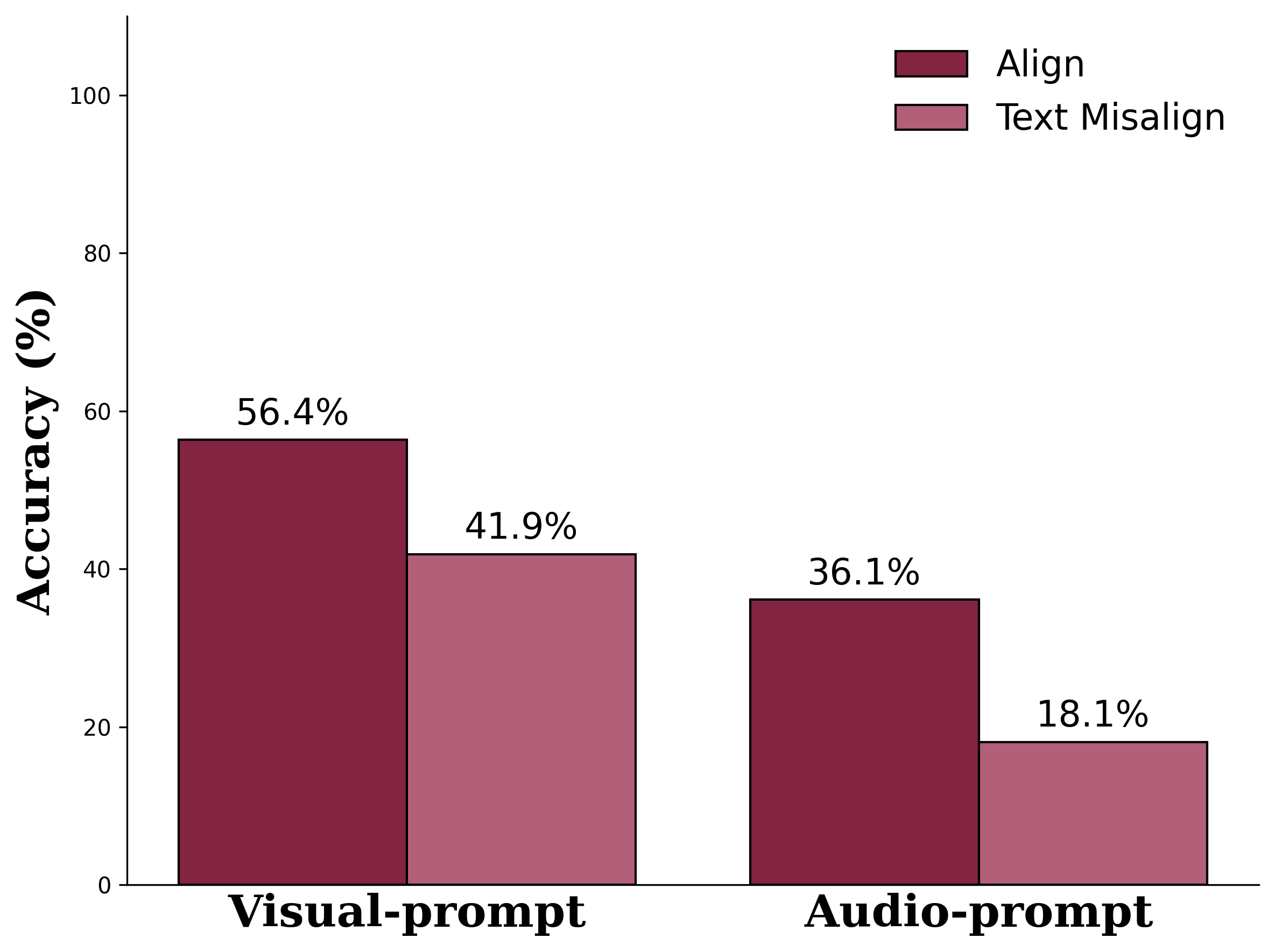}
        \caption{VideoLLaMA2.}
        \label{fig:textmis_vl2}
    \end{subfigure}

    \vspace{1mm}
    \begin{subfigure}[t]{0.48\linewidth}
        \centering
        \includegraphics[width=\linewidth]{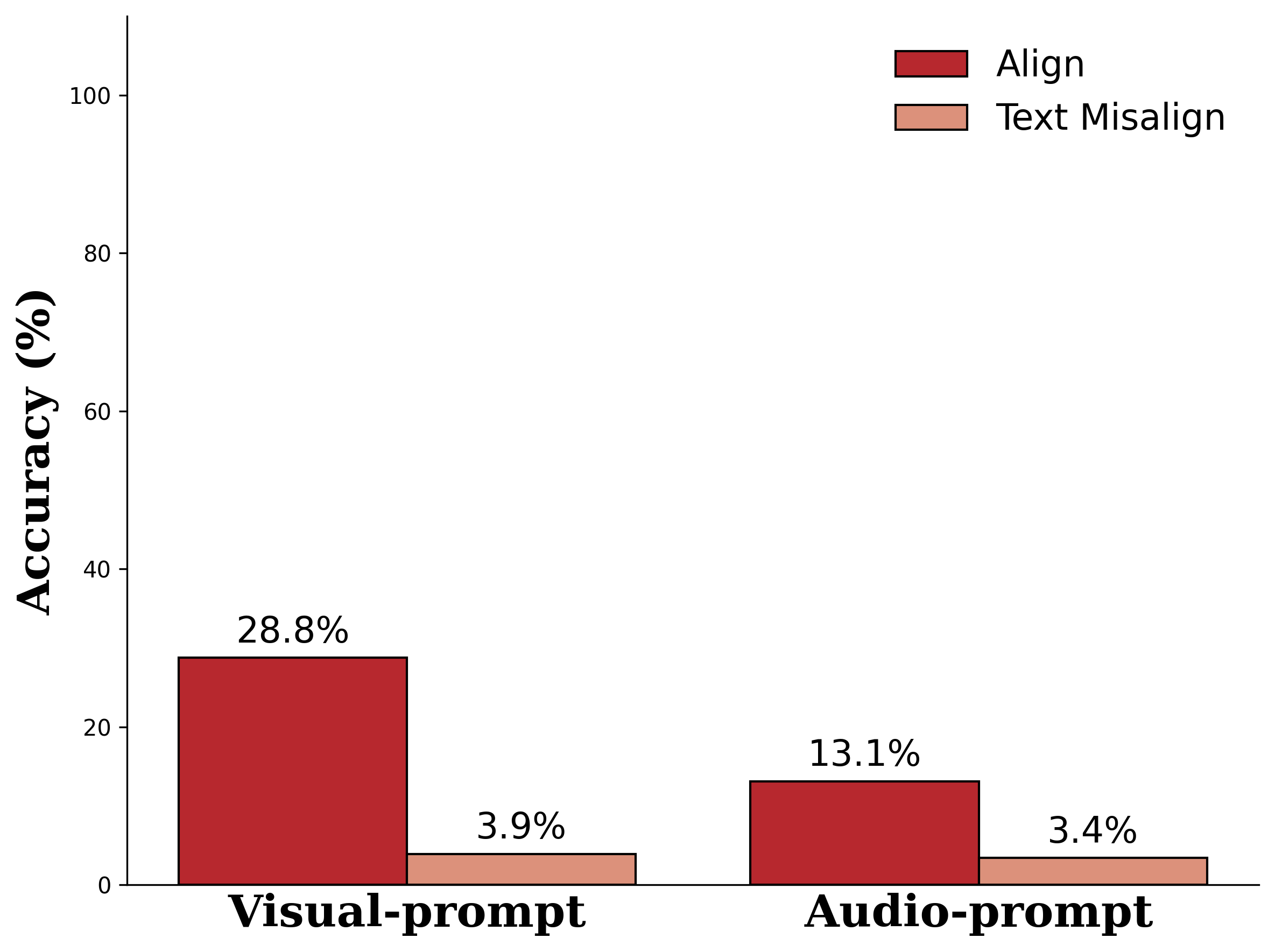}
        \caption{PandaGPT.}
        \label{fig:textmis_pandagpt}
    \end{subfigure}
    \hfill
    \begin{subfigure}[t]{0.48\linewidth}
        \centering
        \includegraphics[width=\linewidth]{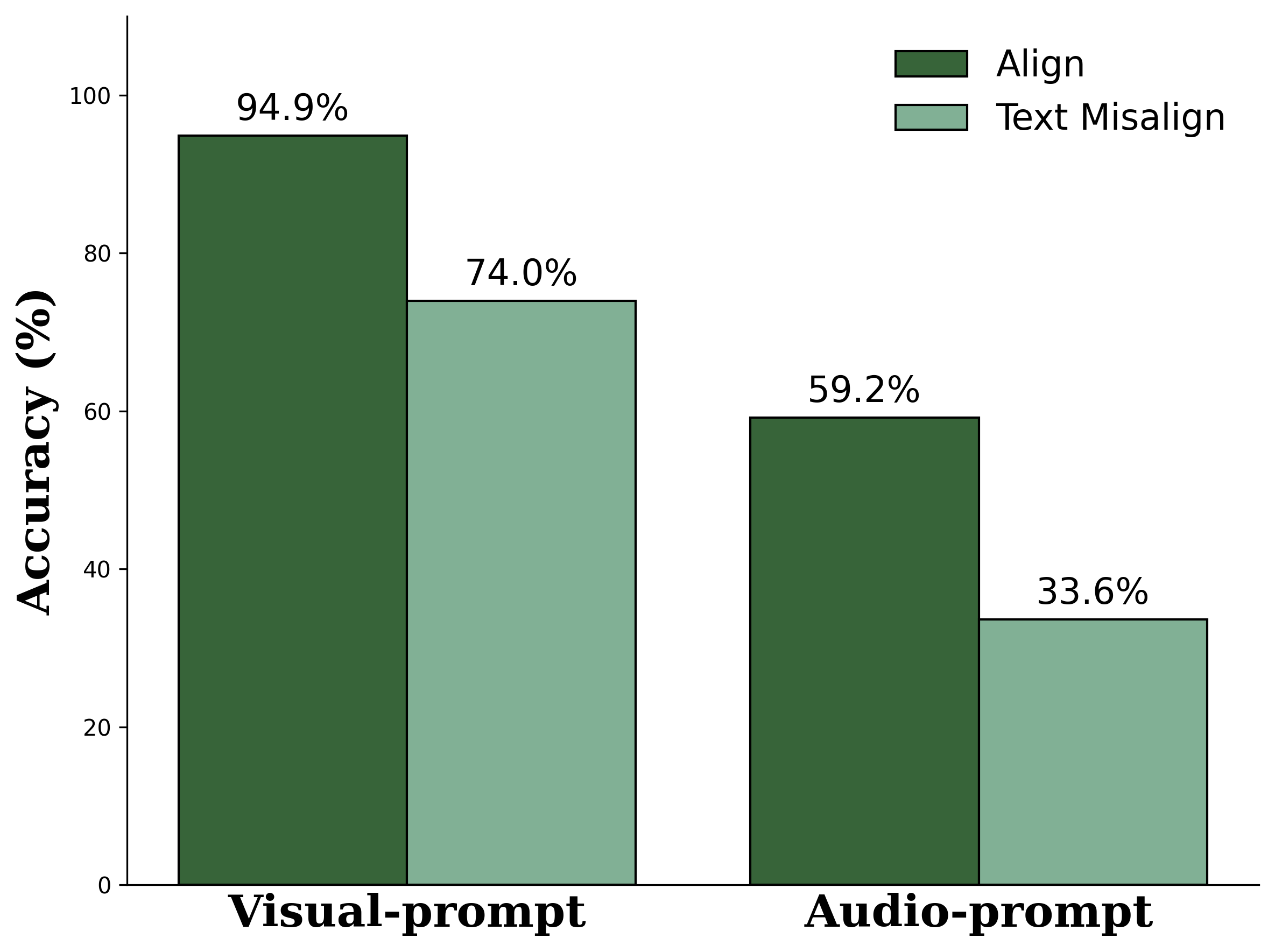}
        \caption{\gemini.}
        \label{fig:textmis_gemini}
    \end{subfigure}

    \caption{
    \textbf{Model sensitivity to misleading textual context.}  
    Each subfigure reports accuracy under \textbf{visual-only} (left bar group) and \textbf{audio-only} (right bar group) prompts, comparing clean (\textit{Baseline}) versus misleading-text (\textit{Text Misalign}) conditions.  
    }
    \label{fig:text_misalignment}
    \vspace{-3mm}
\end{figure}

\begin{figure}[t]
    \centering
    \begin{subfigure}[t]{0.48\linewidth}
        \centering
        \includegraphics[width=\linewidth]{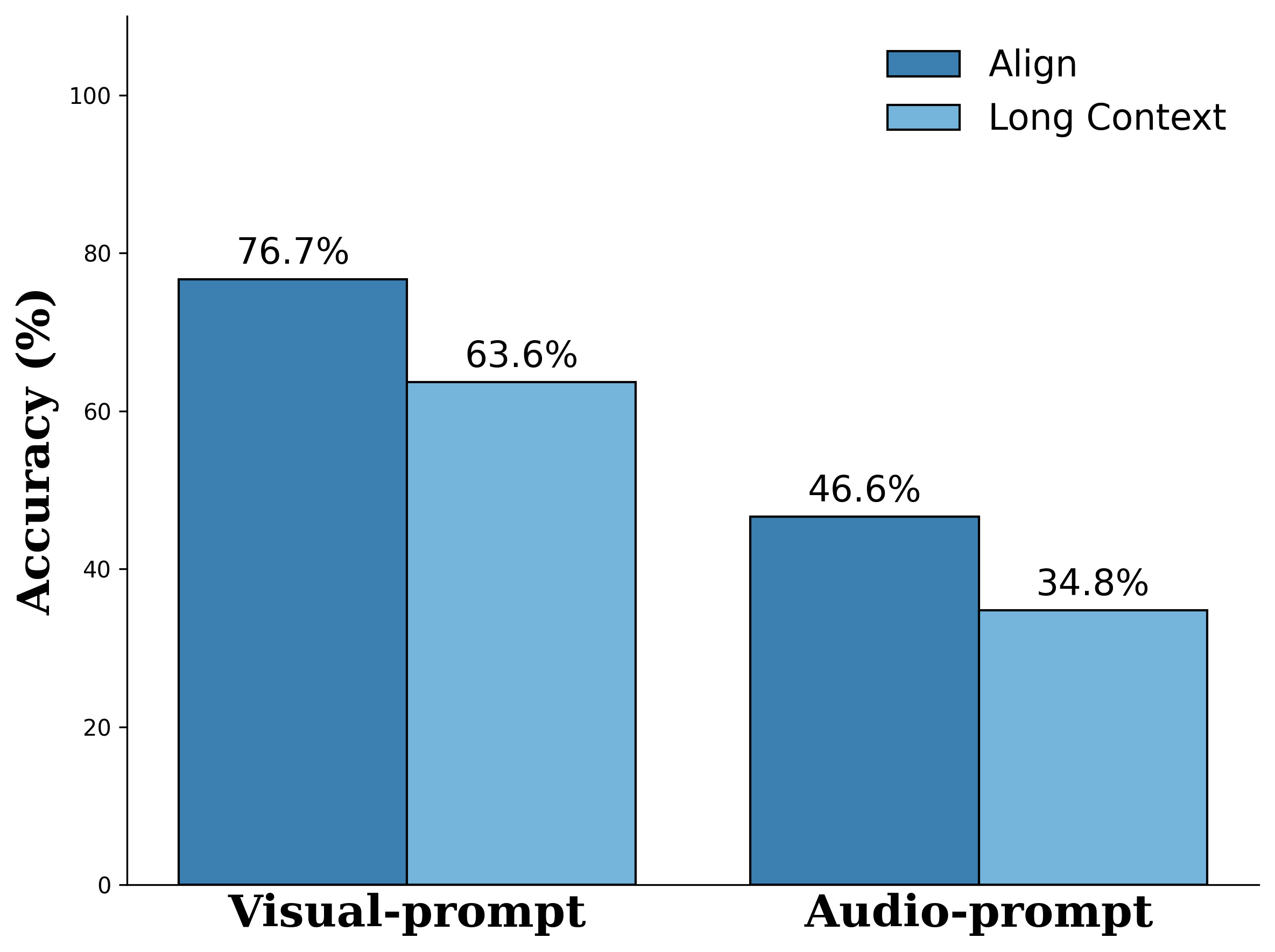}
        \caption{\qwenSB.}
        \label{fig:long_context_qwen}
    \end{subfigure}
    \hfill
    \begin{subfigure}[t]{0.48\linewidth}
        \centering
        \includegraphics[width=\linewidth]{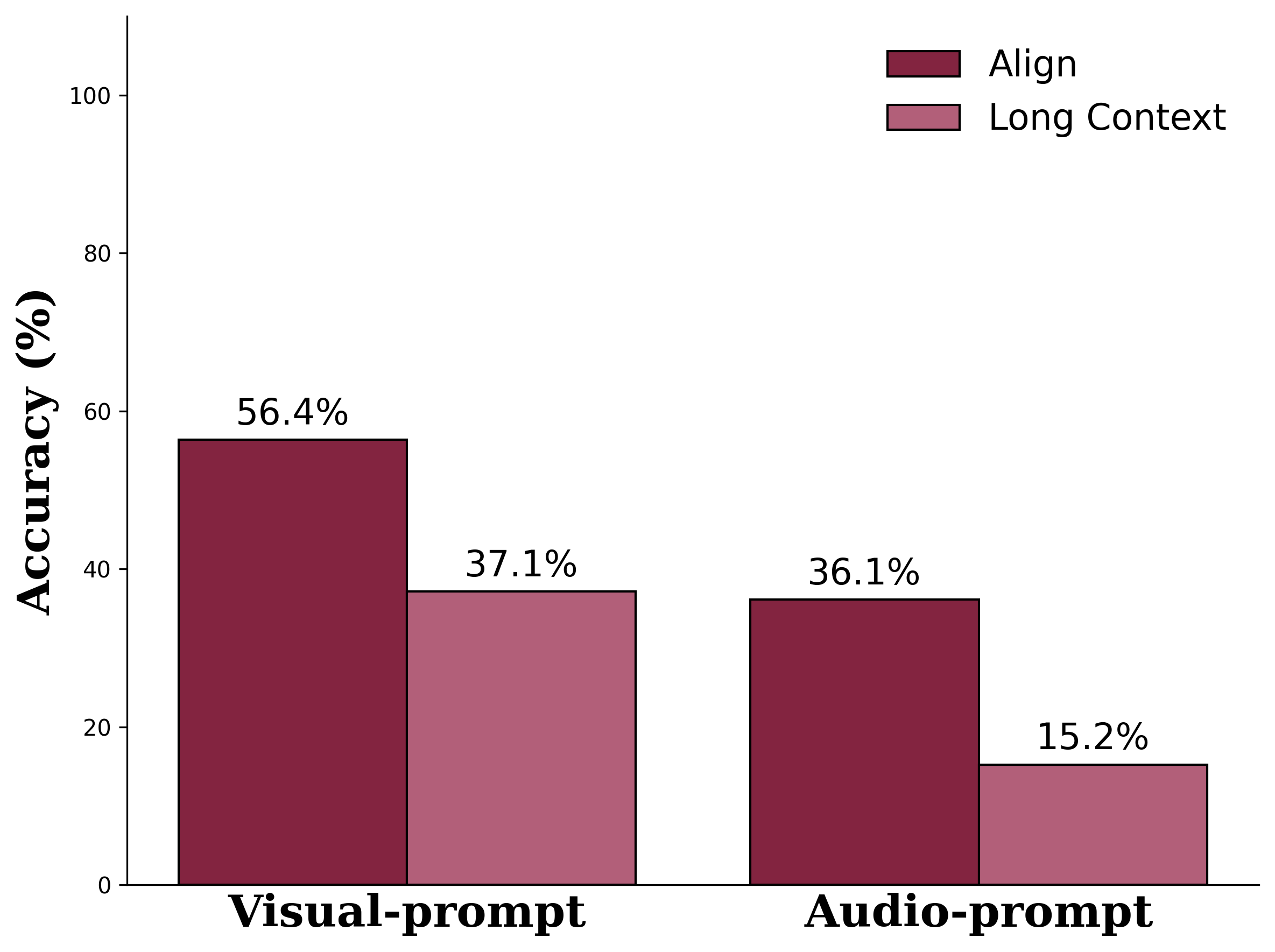}
        \caption{\vllama.}
        \label{fig:long_context_videollama}
    \end{subfigure}

    \vspace{1mm}
    \begin{subfigure}[t]{0.48\linewidth}
        \centering
        \includegraphics[width=\linewidth]{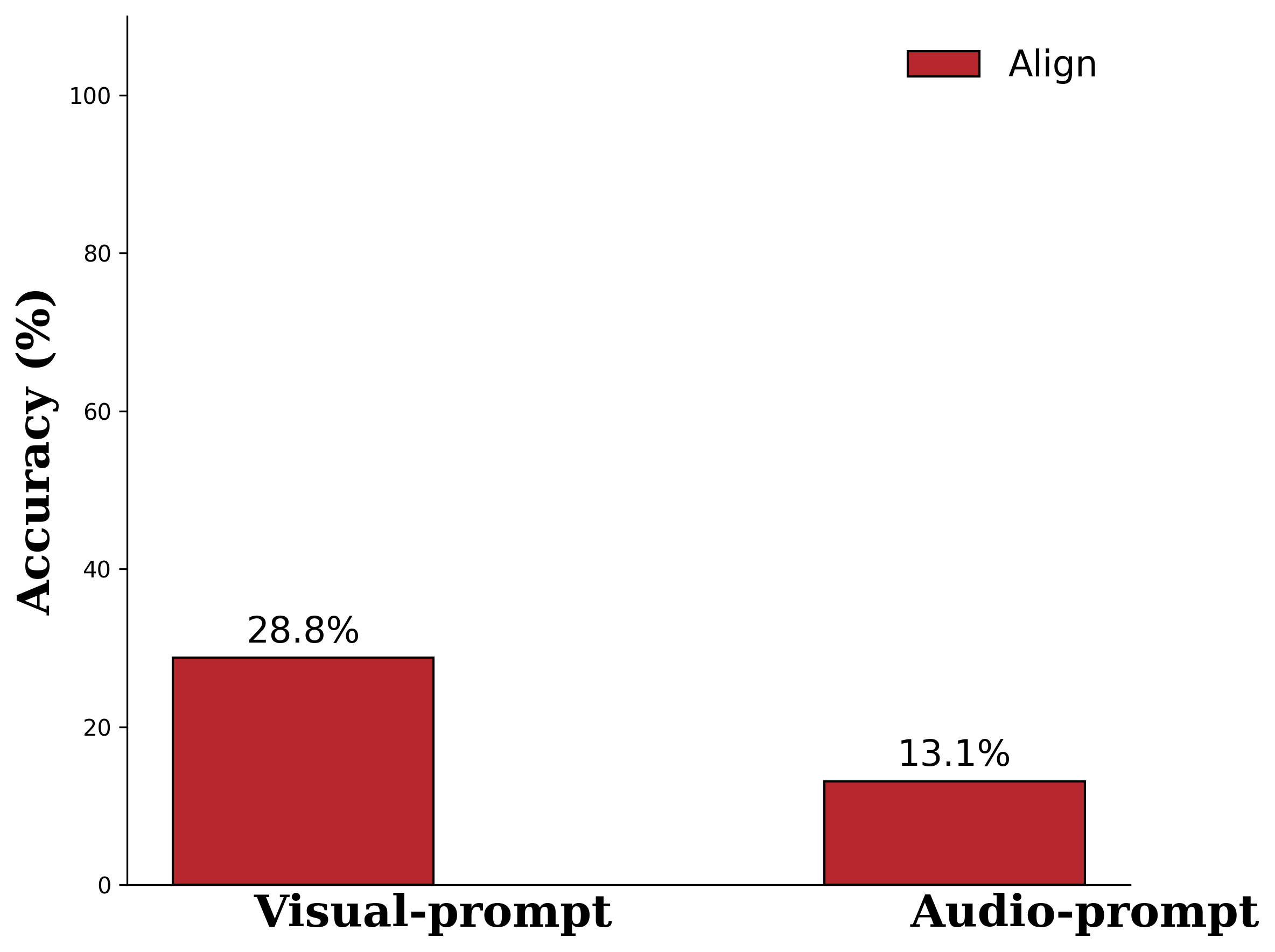}
        \caption{\panda.}
        \label{fig:long_context_pandagpt}
    \end{subfigure}
    \hfill
    \begin{subfigure}[t]{0.48\linewidth}
        \centering
        \includegraphics[width=\linewidth]{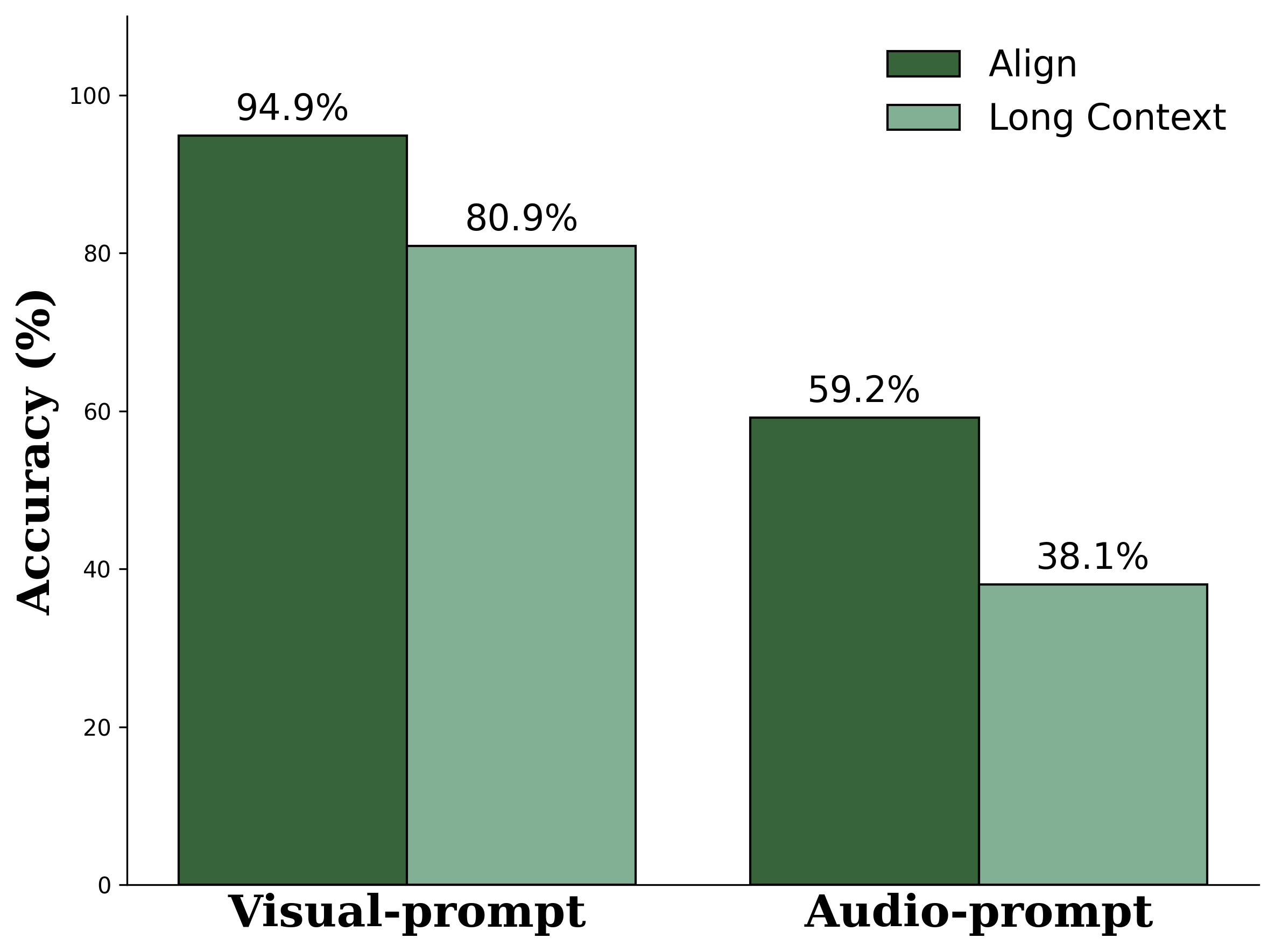}
        \caption{\gemini.}
        \label{fig:long_context_gemini}
    \end{subfigure}

    \caption{
    \textbf{Performance degradation under long-context interference across MLLMs.}  
    Each subfigure shows accuracy under \textbf{visual-only} (left axis) and \textbf{audio-only} (right axis) prompts when irrelevant long-context text is appended.  
    }
    \label{fig:longcontext_results}
    \vspace{-6mm}
\end{figure}

\noindent\textbf{What do models \textit{hear} and \textit{see} when input is semantically misaligned?}
To examine how models handle conflicting modalities, we use the semantically misaligned subset of {\bench} described in Sec.~\ref{sec:3.1.1} where the video and audio carry inconsistent semantics. The aligned setting serves as the baseline.  
As shown in Fig.~\ref{fig:semantic_misalignment}, all models degrade under misalignment, but the extent and pattern differ sharply across architectures.
{\qwenSB} and {\vllama} suffer notable accuracy drops on both visual- and audio-focused prompts. Roughly 20–25\% and 15–20\% declines, respectively, yet still maintain reasonable performance
This pattern indicates that both models integrate information from multiple modalities but \textit{fail to selectively suppress misleading cues} when semantics conflict.  
By contrast, {\gemini} and {\panda} display a distinct shortcut-taking behavior.   
Their visual accuracy remains nearly unchanged, sometimes even slightly improved, while their audio accuracy collapses to near-random levels (1–4\%).  
These models rely almost exclusively on vision, i.e., \textit{seeing without listening} resulting in strong degradation of audio performance when modalities conflict.  


\subsubsection{\textcolor{ForestGreen}{Video $\Leftrightarrow$ Audio} \textcolor{red}{$\nLeftrightarrow$ Text}} \label{sec:blackbox_analsyis_text}
We evaluate models on semantically aligned video–audio pairs but prepend each input text prompt in two ways: (a) a deliberately irrelevant short text and (b) irrelevant long context. We wish to study the role of distracting text cues while performing visual or audio classification tasks.

\noindent \textbf{Misleading short text:}
To evaluate the impact of misleading textual context, we prepend each question with a false caption that contradicts the video’s dominant content (e.g., inserting ``\texttt{Video\_caption: Vehicle.}'' for a non-vehicle scene). (Details prompt in Appendix E). We then query using both visual- and audio-focused prompts to test whether inconsistent text biases their reasoning about the corresponding modality.

As shown in Fig.~\ref{fig:text_misalignment}, all models suffer substantial performance degradation when misleading captions are introduced, demonstrating that models are highly susceptible to misleading textual context in the prompt. 
{\qwenSB} and {\panda} show the most severe drops, larger than those observed under audio–visual misalignment. This indicates that both models depend heavily on textual context.  
{\vllama} exhibits a more moderate decline, suggesting partial resistance to contradictory language but still limited ability to ground its predictions to the desired modality in the presence of misleading text.

\noindent\textbf{Irrelevant Long-Context Caption:} 
To test long-context robustness under  irrelevant long-context, we append $10,000$ random text tokens after each visual- and audio-focused prompt while keeping the original video–audio inputs unchanged.
As shown in Fig~\ref{fig:longcontext_results}, all models experience a degradation across both prompts: {\qwenSB} and {\vllama} drop by roughly 15–20\%, and {\gemini} shows a similar decline despite its larger scale.
{\panda} is excluded due to its limited context window ($\sim$400 tokens).
These results confirm that current MLLMs struggle to retain relevant multimodal grounding once overwhelmed by long, irrelevant text.

\begin{figure}[H]
\centering
\noindent\textbf{Takeaways from Black-box and White-box Interpretability}
\vspace{2mm}
\fcolorbox{purple}{shadecolor}{
\parbox{\dimexpr\columnwidth-2\fboxsep}{
\begin{enumerate}[leftmargin=1.5em]
\setlength\itemsep{0.5em}
    \item \textbf{Brittle Modality Integration:} MLLMs fail ungracefully if any one modality is perturbed.
    \item \textbf{Text triumphs:} Small text distractions cripple models, ignoring clear audio-visual cues.
     \item \textbf{Conflict reflects in attention shifts:} Prompt-driven attention separation increases under misalignment
\end{enumerate}
}}
\vspace{-6mm}
\end{figure}

\subsection{White-box Interpretability and Findings}
\label{sec:attn_analysis}
In this section, we study how multiple modalities interact for reasoning tasks using white-box interpretation techniques. As in Sec.~\ref{sec:blackbox}, our baseline is when all modalities are aligned and we compare them against misaligned scenarios. We use {\qwenSB} and {\vllama}, visual- and audio-only prompts (Fig.~\ref{fig:vaprompts}) and $100$ randomly sampled videos from {\bench} detailed in Sec.~\ref{sec:3.1.1}. Specifically, we study how attention patterns evolve as multimodal inputs traverse through the MLLM decoder using a statistical metric: \textbf{\cohend} ($d$)~\cite{cohen2013statistical}. Given two distributions $D_1$ and $D_2$ capturing $n_1$ and $n_2$ samples, their sample means $\mu_1, \mu_2$, and standard deviations $s_1$ and $s_2$, the {\cohend} of the two distributions is defined as:
    \begin{equation}
    \label{eqn:1}
        d(D_1, D_2) = \frac{\mu_1 - \mu_2}{s_p},
    \end{equation}
     where $s_p$ is the weighted average between $s_1$ and $s_2$:
    \begin{equation}
    \quad s_p = \sqrt{\frac{(n_1 - 1)s_1^2 + (n_2 - 1)s_2^2}{n_1 + n_2 - 2}}
    \end{equation}

A positive $d(D_1, D_2)$  indicates $D_1$ has larger mean than $D_2$. Larger absolute values of $d$ reflect greater separation between the two distributions~\cite{cohen2013statistical}. In our analysis, $D_1$ and $D_2$ correspond to attention distributions under visual- and audio-focused prompts, respectively.

\noindent\textbf{Finding 1: Textual attention dominance.} 
When examining every attention head across all transformer layers
, \textbf{textual tokens consistently exhibit the highest attention magnitudes}. On average, $81.29\%$ of the total attention weights aggregated over all the layers in {\qwenSB} is concentrated within text tokens and around $56.29\%$  in {\vllama}. This echos our observation in Sec.~\ref{sec:blackbox_analsyis_text} where all models show very sharp performance decline under textual misalignment.

\begin{figure}[t]
    \centering

    \fcolorbox{black!15}{black!3}{
        \parbox{0.98\linewidth}{
            \centering
            \small
            \textbf{$D_1$ = Visual-focused prompt \quad\;$D_2$ = Audio-focused prompt}
        }
    }

    \begin{subfigure}[t]{0.48\linewidth}
        \centering
        \includegraphics[width=\linewidth]{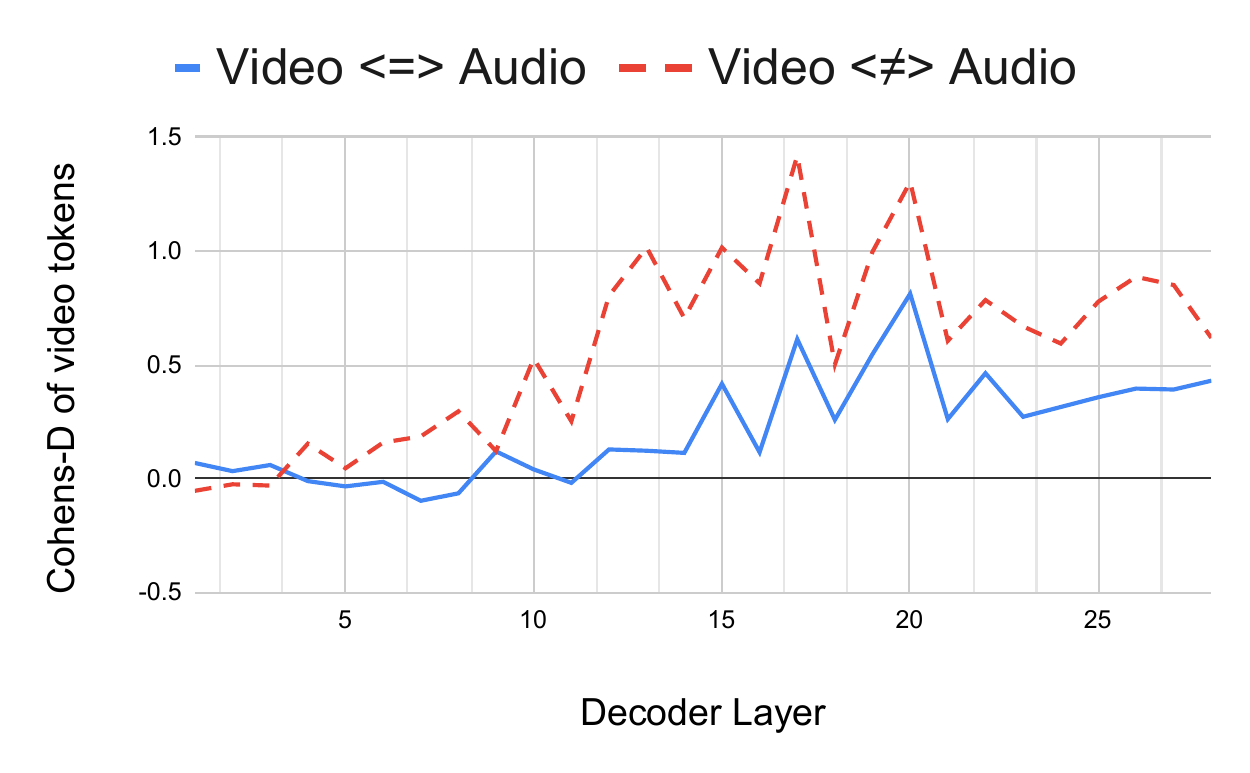}
        \caption{Visual tokens}
        \label{fig:semantic_comparision_cohensd_video_qwen}
    \end{subfigure}
    \hfill
    \begin{subfigure}[t]{0.48\linewidth}
        \centering
        \includegraphics[width=\linewidth]{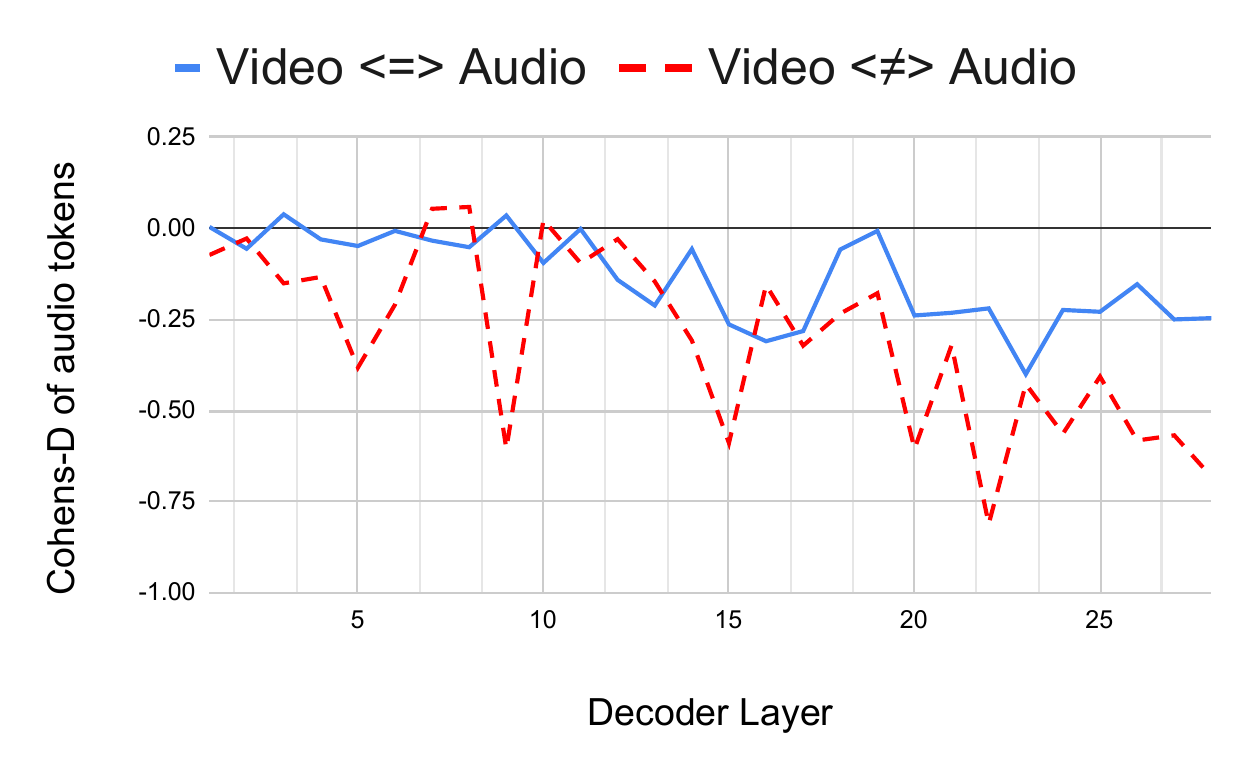}
        \caption{Audio tokens}
        \label{fig:semantic_comparision_cohensd_audio_qwen}
    \end{subfigure}

    \caption{
        \textbf{Layer-wise {\cohend} trends for aligned vs.\ misaligned samples.}
        Misaligned samples (red dotted lines) exhibit consistently higher {\cohend} magnitudes for both visual (left) and audio tokens (right), indicating stronger modality-selective attention shifts under conflict.
    }
    \vspace{-5mm}
    \label{fig:whitebox_semantic_misalignment_qwen}
\end{figure}

\noindent\textbf{Finding 2: Modality Selectivity Under Misalignment}
Next, to focus exclusively on visual–audio interactions, we ignore the attention weights of text tokens and re-normalize the remaining attention over visual and audio tokens.
This enables us to study the relative importance of the two modalities. 
Then, we define $D_1$ as the attention weight distribution of visual tokens under visual-only prompt while $D_2$ is the attention weight distribution of visual tokens under audio-only prompt (Fig.~\ref{fig:vaprompts}).  As shown in Fig.~\ref{fig:whitebox_semantic_misalignment_qwen} (a), {\qwenSB} exhibits consistently positive {\cohend} values for video tokens, indicating higher separation between the attention values under the two prompt settings. Further, in the event that the video and audio are misaligned (red line), {\cohend} values are much higher. This indicates that the model shifts attention more strongly between modalities when the semantic relation between them is disrupted. This strongly corroborates our analysis in Sec.~\ref{sec:3.1.3}.

Conversely, when we repeat this analysis using audio tokens instead, as shown in Fig.~\ref{fig:whitebox_semantic_misalignment_qwen} (b), we see negative {\cohend} values. This suggests that $D_2$, the attention weight distribution of audio tokens under audio-only prompt, has higher mean value compared to $D_1$ during both aligned (blue line) and misaligned (red line) cases. This is intuitive as it reflects model allocating more attention to auditory inputs under audio-focused prompts ($D_2$). These trends confirm that {\qwenSB} tries to attend to the modality emphasized in the prompt.

However, these attention shifts remain relatively weak. We hypothesize that \textbf{stronger, more decisive reweighting is required for robust modality-specific reasoning}. Indeed, we show that after finetuning
, {\cohend} values increase meaningfully ( Sec.~\ref{sec:post_train_attn}).

\noindent\textbf{Finding 3: Modality early layers vs later layers:}  
From Fig.~\ref{fig:whitebox_semantic_misalignment_qwen}, {\cohend} magnitudes increase across layers, peaking around layer 17 for video tokens and 22 for audio in misaligned case. This means that the \textbf{deeper layers change their attention more strongly} when the prompt focus switches from video to audio. Rather than showing separation between modalities, this likely reflects that the higher layers are deciding which modality to trust - a sign of semantic integration rather than raw fusion.



\section{Experiments} \label{sec:exp}
We hypothesize that models like {\qwen} showing larger attention shifts, reflected by higher Cohen's-D, adapt better when finetuned. We proceed to validate this next.
\subsection{Implementation Details}\label{sec:implementation_details}
\noindent \textbf{Data pre-processing:} We start off from the training split of the {\audioset} dataset and perform the $2$ stage automated filtering detailed in Sec~\ref{sec:mmabench}. We then rescale each training video to preserve the aspect ratio~\cite{akbari2021vatt,wang2024internvideo2,cheng2024videollama}, followed by a center crop to size $504\times 504$. After filtering, we retain 1,207 audio–visual aligned samples from the training split of {\audioset} spanning 58 sound classes. To simulate semantic misalignment, each aligned video is paired with $10$ clips from distinct classes. We replace the audio tracks with that of the original aligned video, giving ~$13,277$ videos. Details about video sample pre-processing are in Appendix.


\noindent\textbf{Training objectives:}
We fine-tune {\qwenSB} using the modality-specific semantic classification tasks introduced in Sec.~\ref{sec:mmabench}.  The model is trained to answer the visual-focused and audio-focused questions correctly for each clip, based on the intended modality, even though both modalities are still provided as input.  
This encourages the model to ground its predictions in the appropriate modality while maintaining balanced cross-modal reasoning.

\noindent \textbf{Fine-tuning setup:} We adopt a lightweight parameter-efficient fine-tuning strategy using LoRA adapters~\cite{hu2022lora} applied to all linear projection layers in both the attention and feed-forward modules of {\qwenSB}. This design preserves pretraining knowledge while allowing efficient learning.
All experiments are implemented using the public \texttt{LLaMA-Factory} pipeline~\cite{zheng2024llamafactory}, which provides stable support for multimodal fine-tuning and LoRA configuration management. Detailed configuration in Appendix.


\begin{table*}[t]
    \centering
    \small
    \setlength{\tabcolsep}{5pt}
    \begin{tabular}{l|cc|cc}
    \toprule
    & \multicolumn{2}{c|}{\textbf{Visual Prompt (\%)}} & \multicolumn{2}{c}{\textbf{Audio Prompt (\%)}} \\
    \textbf{Model} & \textbf{Align} & \textbf{Misalign} & \textbf{Align} & \textbf{Misalign} \\
    \midrule
    \textit{Closed-Source Baselines} & & & & \\
    Gemini-2.5-Pro & \textbf{97.90} & \textbf{95.28} & 60.37 & 24.95 \\
    Gemini-2.0-Flash & \underline{96.71} & 91.91 & 57.21 & 9.42 \\
    Gemini-2.0-Flash-Lite & 94.89 & 94.11 & 59.19 & 4.04 \\
    \midrule
    \textit{Open-Source Baselines} & & & & \\
    Qwen3-Omni-30B-Instruct & 92.88 & 83.73 & 57.39 & 14.58 \\
    Qwen2.5-Omni-7B (Base) & 76.68 & 58.72 & 46.60 & 25.16 \\
    VideoLLaMA2 & 56.35 & 36.11 & 36.12 & 18.46 \\
    ChatBridge & 51.64 & 54.71 & 41.61 & 7.07 \\
    PandaGPT & 28.75 & 29.79 & 13.12 & 1.18 \\
    \midrule
    \rowcolor{gray!15} \textbf{Qwen2.5-Omni-7B + Ours} & 94.68 & \underline{94.37} & \textbf{88.14} & \textbf{79.79} \\
    \bottomrule
    \end{tabular}
    \caption{\textbf{Benchmarking against State-of-the-Art.} Comparison of our fine-tuned model against a wide range of baselines. \textbf{Bold} indicates the best performance, and \underline{underline} indicates the second best. Our method (bottom row) achieves the highest audio robustness by a significant margin, outperforming even 30B-parameter and proprietary models in handling conflicting modalities.}
    \label{tab:sota_comparison}
\end{table*}
\subsection{Quantitative Results}
We evaluate the fine-tuned {\qwenSB} on the {\bench} benchmark introduced in Sec.~\ref{sec:mmabench}.
The best-performing checkpoint from the fine-turning step detailed in Sec.~\ref{sec:implementation_details} is selected based on average validation accuracy over a 1K-sample set (details in Appendix).  

\noindent\textbf{Semantic misalignment.}  
As summarized in Table~\ref{tab:sota_comparison}, modality-aware fine-tuning substantially improves {\qwenSB}'s robustness across both aligned and misaligned settings.  
The tuned model achieves 20--40\% absolute gains on visual- and audio-focused prompts, maintaining balanced performance even when modalities conflict.  
This demonstrates that explicit alignment supervision effectively mitigates the model’s visual dominance and strengthens cross-modal integration.

To contextualize these gains, we compare our fine-tuned {\qwenSB} against a broader set of multimodal architectures—including {\vllama}, {\panda}, {\gemini}, {\geminiflash}, {\geminipro}, {\qwenThree}, and {\chatbridge} under identical evaluation protocols.  
Despite its smaller scale (7B vs.\ 30B), our model achieves the highest overall robustness, particularly under audio–visual conflict: while large models like {\geminipro} and {\qwenThree} collapse to 24.95\% and 14.58\%, the fine-tuned {\qwenSB} reaches 79.79\%.  
This $>$50\% margin shows that targeted modality alignment is more effective than scaling alone.  

Moreover, this improvement does not compromise visual reasoning. {\qwenSB} maintains 94.68\% accuracy on visual-aligned benchmarks, comparable to {\geminipro} (97.90\%) and exceeding {\qwenThree} (92.88\%).  
These results confirm that our fine-tuning strategy disentangles modality pathways, allowing selective attention to the correct sensory input without degrading overall multimodal capability.

\noindent \textbf{2. Unimodal Ablation}
\begin{table}[t]
\centering
\caption{\textbf{Unimodal ablation results for Qwen2.5-Omni-7B.}
Accuracy (\%) when one modality is removed, before and after modality-aware fine-tuning.}
\setlength{\tabcolsep}{4pt}
\begin{tabular}{lcc}
\toprule
\textbf{Condition} & \textbf{Visual Prompt} & \textbf{Audio Prompt} \\
\midrule
Aligned             & 76.68 & 46.60 \\
Audio Removed         & 71.49 & 54.39 \\
Frames Zeroed         & 45.28 & 33.74 \\
\midrule
Audio Removed + Ours           & 95.30 & 16.17 \\
Frames Zeroed  + Ours        & 8.51  & 82.52 \\
\bottomrule
\end{tabular}
\label{tab:unimodal_tune_qwen}
\vspace{-4mm}
\end{table}

Table~\ref{tab:unimodal_tune_qwen} reports results when one modality is removed. Fine-tuning improves unimodal reasoning—visual accuracy increases by over 20\% with silent audio, and audio accuracy by nearly 50\% when video frames are zeroed. Conversely, performance declines in scenarios where visual prompts are used without visual input, suggesting a less hallucinating and a more reliable model.

\begin{table}[t]
\centering
\caption{
\textbf{Effect of misleading textual context on {\qwenSB}.}
Accuracy (\%) before and after modality-aware fine-tuning under two settings:
(\textit{a}) misleading captions prepended to the query (\textit{text misalignment}) and
(\textit{b}) 10K random tokens appended after the query (\textit{long context}).
}
\setlength{\tabcolsep}{6pt}
\begin{tabular}{lcc}
\toprule
\textbf{Condition} & \textbf{Visual Prompt} & \textbf{Audio Prompt} \\
\midrule
\multicolumn{3}{l}{\textit{(a) Text Misalignment}} \\
{\qwenSB} & 37.81 & 11.98 \\
+ Ours (Fine-tuned) & \textbf{91.88}\textcolor{ForestGreen}{\,(+54.07)} & \textbf{28.63}\textcolor{ForestGreen}{\,(+16.65)} \\
\midrule
\multicolumn{3}{l}{\textit{(b) Long Context (10K Tokens)}} \\
{\qwenSB} & 63.65 & 34.75 \\
+ Ours (Fine-tuned) & \textbf{78.02}\textcolor{ForestGreen}{\,(+14.37)} & \textbf{28.36}\textcolor{red}{\,(-6.39)} \\
\bottomrule
\end{tabular}
\label{tab:text_longcontext_tune_qwen}
\vspace{-2mm}
\end{table}

\noindent\textbf{3. Misleading textual context.}
Table~\ref{tab:text_longcontext_tune_qwen} shows the effects of irrelevant textual information under two settings: (\textit{a}) prepended false captions and (\textit{b}) appended long irrelevant text. Under misleading captions, modality-aware tuning yields large gains: over $\mathbf{50\%}$ for visual prompts and $\mathbf{16\%}$ for audio prompts. Importantly, the model was never exposed to text corruption during training. \textbf{We stress that this emergent robustness arises from aligning the visual and audio modalities.}

For long-context inputs, tuning improves visual-prompt accuracy but causes a mild drop for audio prompts, indicating that alignment supervision alone cannot fully resolve failures caused by extremely long irrelevant text.

\subsubsection{Attentions After Modality-Aware Tuning}
\label{sec:post_train_attn}
\begin{figure}[t]
    \centering
       \fcolorbox{black!15}{black!3}{
        \parbox{0.98\linewidth}{
            \centering
            \small
            \textbf{$D_1$ = Visual-focused prompt \quad\;$D_2$ = Audio-focused prompt}
        }
    }
    \begin{minipage}[t]{0.98\linewidth}
        \begin{subfigure}[t]{0.48\linewidth}
            \centering
            \includegraphics[width=\linewidth]{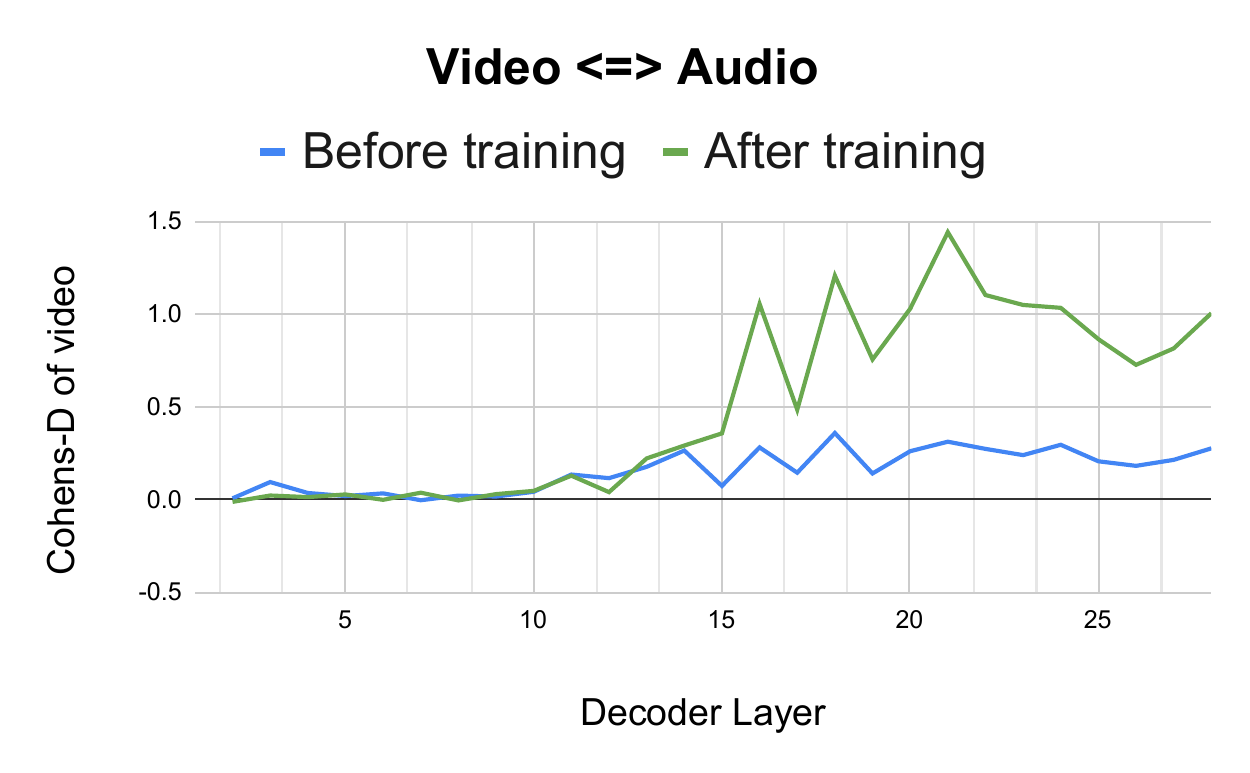}
            \caption{Video tokens (Aligned)}
            \label{fig:aligned_video}
        \end{subfigure}
        \hfill
        \begin{subfigure}[t]{0.48\linewidth}
            \centering
            \includegraphics[width=\linewidth]{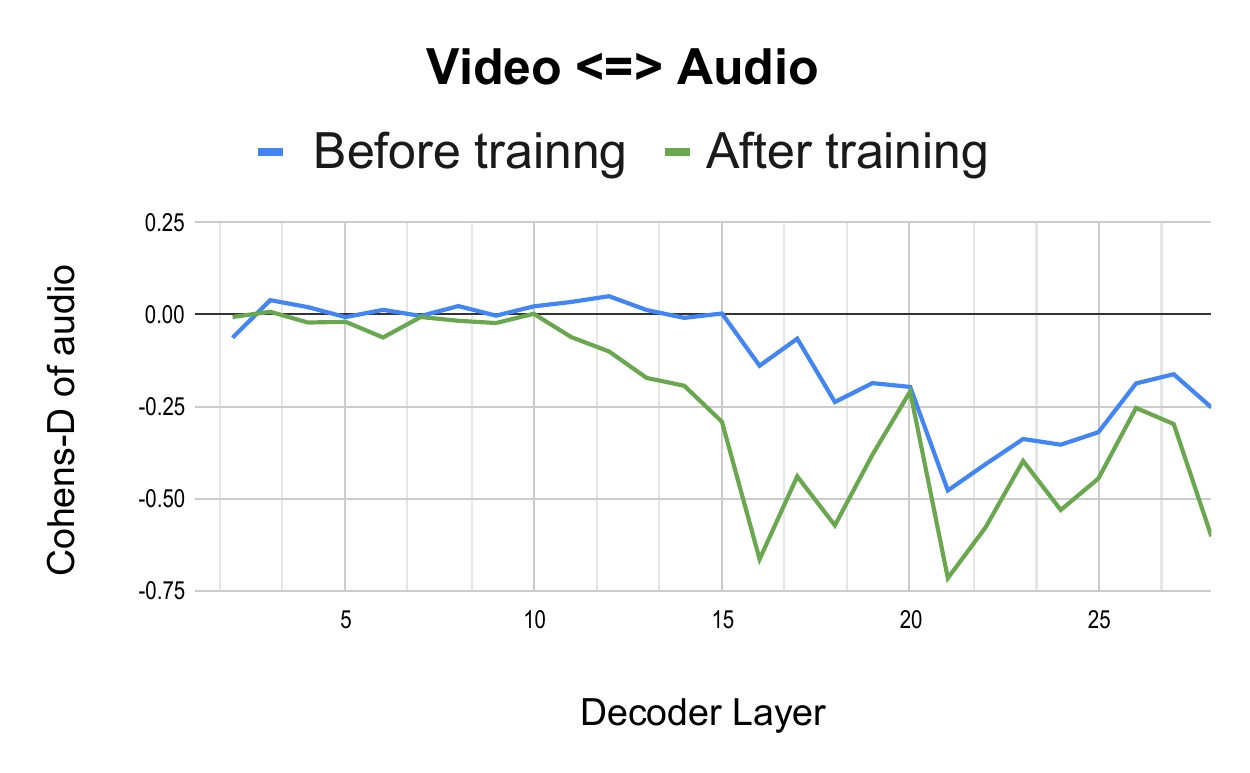}
            \caption{Audio tokens (Aligned)}
            \label{fig:aligned_audio}
        \end{subfigure}
        \\[2mm]

        \begin{subfigure}[t]{0.48\linewidth}
            \centering
            \includegraphics[width=\linewidth]{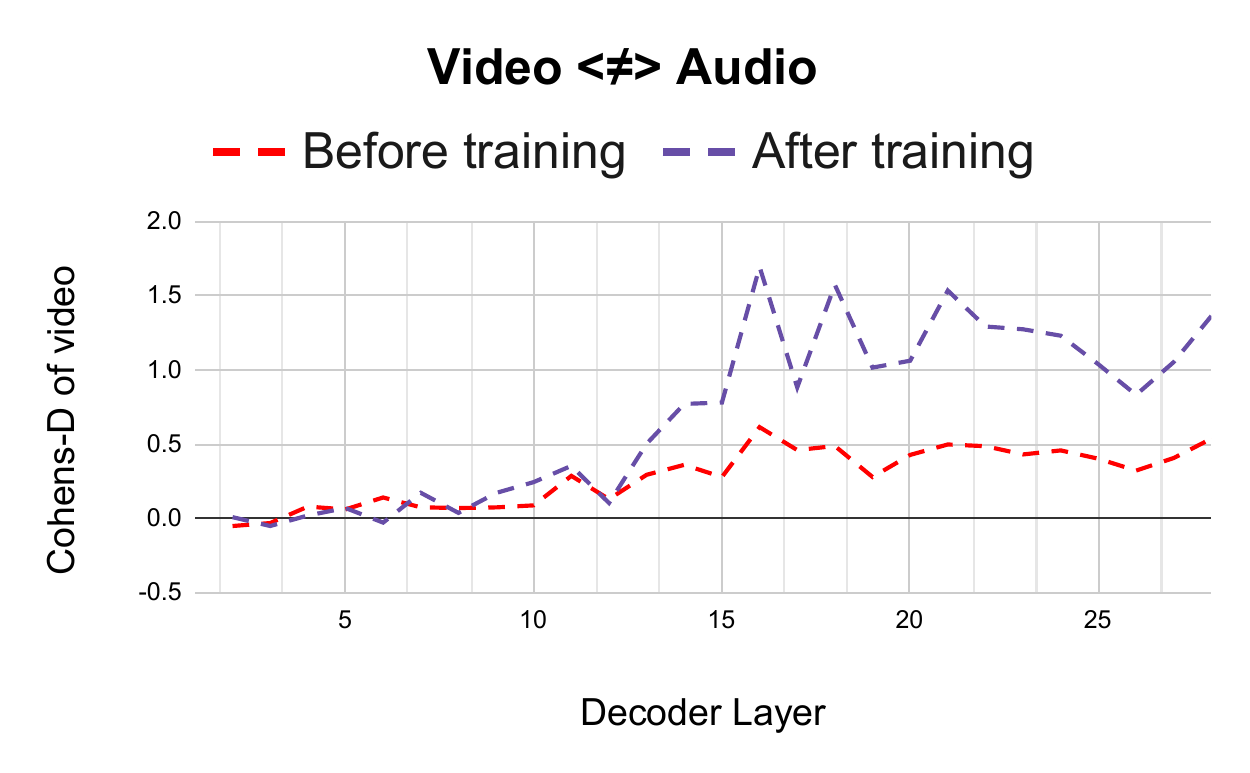}
            \caption{Video tokens (Misaligned)}
            \label{fig:misaligned_video}
        \end{subfigure}
        \hfill
        \begin{subfigure}[t]{0.48\linewidth}
            \centering
            \includegraphics[width=\linewidth]{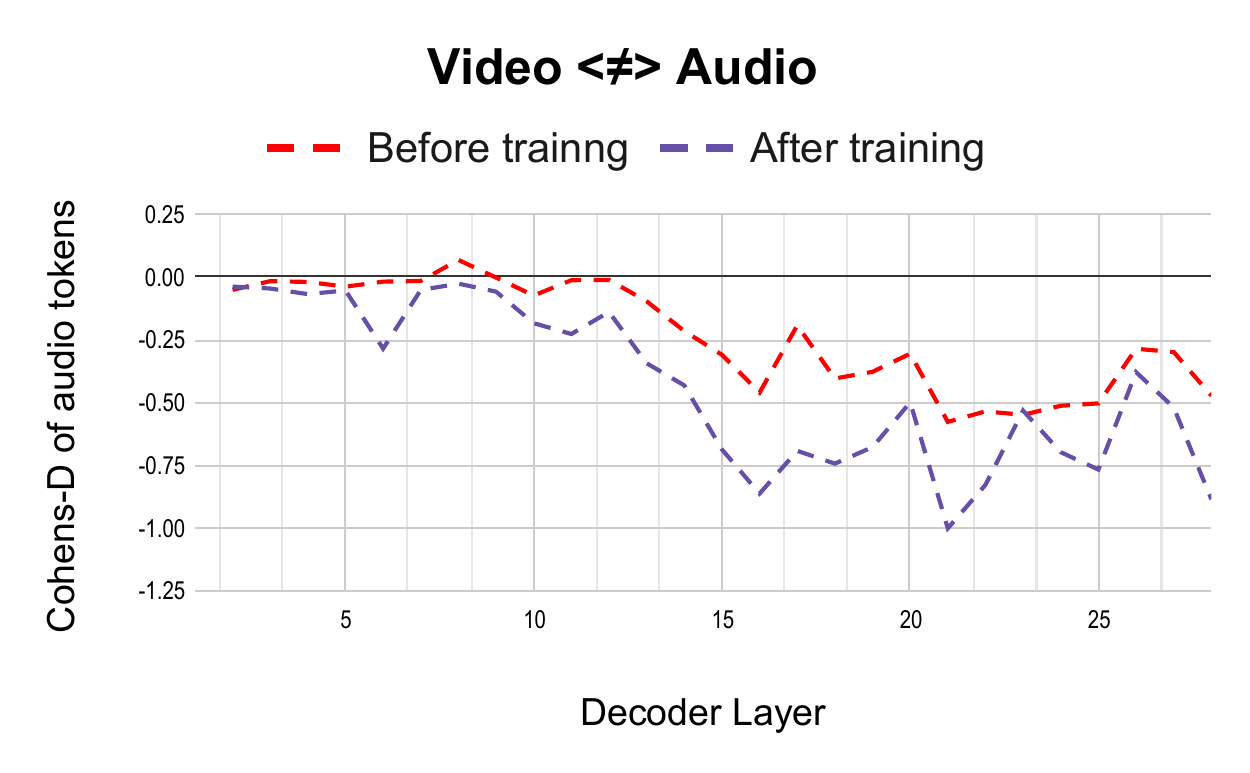}
            \caption{Audio tokens (Misaligned)}
            \label{fig:misaligned_audio}
        \end{subfigure}
    \end{minipage}
    \hfill


    \vspace{1mm}
    \caption{
        \textbf{Layer-wise Cohen's-D before and after modality-aware tuning.}
        Higher absolute values indicate stronger separation between attention distributions under $D_1$ (visual prompt) and $D_2$ (audio prompt).
        After tuning, the model reallocates attention more decisively toward the modality emphasized by the prompt, reflecting improved modality-selective reasoning.
    }
    \label{fig:cohensD_before_after}
    \vspace{-4mm}
\end{figure}

To understand how modality-aware tuning improves the model's multimodal understanding behavior, we analyze layer-wise attention shifts before (solid blue for aligned and red dotted for misaligned) and after training (solid green for aligned and dotted lilac for misaligned) using the Cohen's-D metric introduced in Sec.~\ref{sec:attn_analysis}. 
As shown in the Fig~\ref{fig:cohensD_before_after}, after fine-tuning, the absolute magnitude of Cohen's-D increases across layers, particularly in deeper ones. This suggests that the model now adjusts its attention more decisively toward the modality emphasized by the prompt. We believe that the performance improvements are not merely from memorization, but from a more principled reallocation of attention toward the most relevant modality as noted in the prompt.
\subsection{Zero-shot Evaluation on AVHBench}
\label{sec:avhbench}
\begin{table}[t]
\centering
\caption{\textbf{Zero-shot evaluation on AVHBench.}  
Performance of \textit{Qwen2.5-Omni} before and after modality-aware fine-tuning.  
The first three tasks are binary classification (accuracy in \%), and the last is an open-ended captioning task evaluated using METEOR.
}
\vspace{-2mm}
\setlength{\tabcolsep}{1.8pt}
\begin{tabular}{lcc}
\toprule
\textbf{Task} & \textbf{{\qwen}} & \textbf{+Ours} \\
\midrule
Video-driven Audio Hallucination  & 71.67  & \textbf{79.86} \\
Audio-driven Video Hallucination  & 80.65  & \textbf{85.32} \\
AV Matching                       & 49.57  & 49.57 \\
AV Captioning (METEOR)            & 15.50  & \textbf{15.60} \\
\bottomrule
\end{tabular}
\label{tab:avhbench}
\vspace{-5mm}
\end{table}
We evaluate the zero-shot generalization ability of our modality-aware-tuned model on AVHBench~\cite{sung2024avhbench}, a recent benchmark diagnosing audio–visual hallucination and cross-modal consistency in MLLMs.  
As summarized in Table~\ref{tab:avhbench}, modality-aware fine-tuning yields clear gains on hallucination-oriented tasks, improving accuracy on both Video-driven Audio Hallucination (V2A) and Audio-driven Video Hallucination (A2V) by $\mathbf{+8.2\%}$ and $\mathbf{+4.7\%}$, respectively. These results indicate that modality-aware-tuning helps in reducing hallucinations, a very different task, on unseen data. Performance on \textit{AV Matching} remains unchanged, while \textit{AV Captioning} shows a marginal improvement ($+0.1$ METEOR). This equally important result suggests that modality-aware-tuning improves grounding but does not negatively impact other, broader reasoning or generative tasks. (More details in Appendix.)
\section{Limitations and Future Work}
This work provides a first, crucial step towards systematic interpretation of MLLM robustness to semantic misalignments. Our analysis reveals significant performance instabilities and an over-reliance on text or visual modalities, highlighting key vulnerabilities in current models. While this evaluation is currently limited to {\qwenSB} and classification-style QA, our immediate goal is to extend the analysis to larger Qwen models and other architectures like {\panda}~\cite{su2023pandagpt} and {\vllama}~\cite{cheng2024videollama}. Future work will further broaden this scope to include open-ended reasoning and more fine-grained temporal misalignment tasks.

\section{Acknowledgements}
We thank our collaborators and colleagues for their valuable feedback and discussion throughout this project. 
We also respectfully acknowledge that \textbf{Arjun Akula participated in an advisory capacity only}.
\clearpage
{
    \small
    \bibliographystyle{ieeenat_fullname}
    \bibliography{main}
}

\clearpage
\setcounter{page}{1}
\appendix
\section*{Appendix Table of Contents}
We provide detailed information regarding dataset curation, implementation specifics, and extended experimental analysis in the following sections:
\begin{itemize}
    \item \autoref{appendix:data_pipeline} details the AudioSet ontology simplification process and the rationale for removing ambiguous classes.
    \item \autoref{app:curation} describes the rigorous two-stage sample curation pipeline, including the automated consistency checks and the human verification protocol.
    \item \autoref{app:preprocessing} outlines the video data preprocessing steps, specifically spatial resizing and temporal standardization.
    \item \autoref{app:finetuning} provides the configuration for modality-aware fine-tuning, including LoRA hyperparameters and data augmentation strategies.
    \item \autoref{sec:Cohens-D_supppl} presents extended white-box interpretability results, focusing on head-wise attention statistics and comparisons with VideoLLaMA2.
    \item \autoref{app:heatmaps} visualizes token-level attention heatmaps to quantitatively demonstrate the text dominance in current MLLM architectures.
    \item \autoref{app:text_prompt} displays the specific prompt templates used for textual misalignment and distraction experiments.
    \item \autoref{sec:crossclass} evaluates the cross-class generalization capabilities of the fine-tuned model on unseen categories.
    \item \autoref{app:extended_baselines} extends the black-box analysis to additional baselines (Gemini-Pro, Qwen3-Omni, ChatBridge) across semantic, unimodal, and textual misalignment settings.
    \item \autoref{sec:abstention} introduces the ``None of the Above'' zero-shot abstention evaluation to diagnose hallucination from silence.
    \item \autoref{app:cot} investigates the impact of Chain-of-Thought (CoT) prompting on resolving sensory conflict and alignment drift.
    \item \autoref{app:qualitative} presents a gallery of qualitative examples comparing the improved grounding of our model against baseline failure modes.
\end{itemize}

\section{AudioSet Ontology Details and Dataset Simplification}
\label{appendix:data_pipeline}

The audio event classes in AudioSet~\cite{gemmeke2017audio} are structured as a hierarchical graph ontology, organizing sounds based on semantic granularity. 
This structure captures relationships ranging from broad concepts (roots) to fine-grained events (leaves). 
However, this hierarchy introduces significant label redundancy for our purposes. 
For example, a single video sample is typically annotated with every label along the path from the root to the leaf: a clip might carry the tags ``Animal'' (root), ``Domestic animals, pets'' (intermediate node), and ``Dog'' (leaf). 
While these labels all reference the same acoustic event, treating them as distinct, equal-weighted classes creates multi-label ambiguity.

The complete ontology forest consists of $6$ distinct root categories. 
Furthermore, the graph structure allows nodes to have multiple parents, leading to cross-branch ambiguity. 
For instance, the label ``Hiss'' appears as a child node under both ``Snake'' (within the Animals root) and ``Steam'' (within the Natural Sounds root). 
Additionally, the ontology also provides a restrictions field which marks some of the classes as ``blacklisted" due to being obscure and some as ``abstract" since they are only added to build the tree structure. 

\begin{figure}[t]
    \centering
    \includegraphics[width=\linewidth]{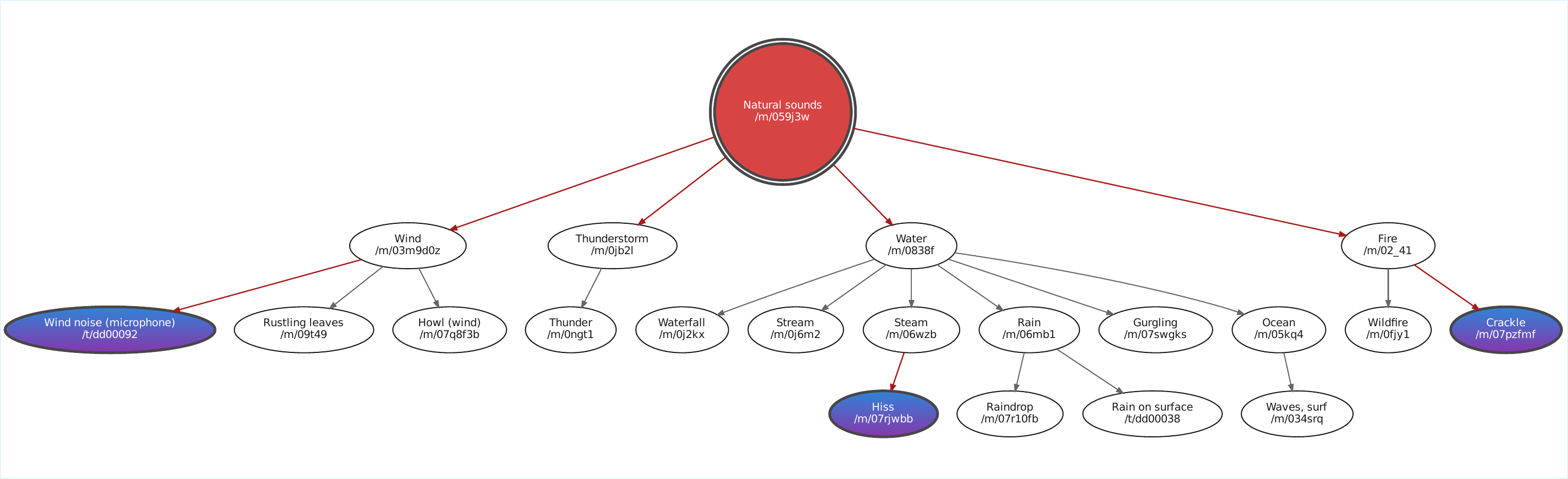}
    \vspace{1mm}
    \caption{
    \textbf{AudioSet Ontology tree of root ``Natural Sounds'' } Nodes colored red are marked ``restricted'' by the original ontology. Nodes with gradient indicate that they have multiple parents from either same or different roots. 
    }
    \label{fig:supp_ontology_tree}
    \vspace{-2mm}
\end{figure}

Consequently, the raw dataset averages $2.7$ class labels per sample. 
This necessitated the rigorous ontology pruning and filtering pipeline described in Section 3 to extract the distinct, single-source events required for precise misalignment analysis. One of the roots of the ontology is shown in Figure~\ref{fig:supp_ontology_tree}
In the original AudioSet ontology (527 annotated classes in the publicly provided version) several issues become salient when the labels are to be used in a multimodal audio-visual classification setting along with above ones:
\begin{itemize}
    \item \textbf{High cardinality of classes} The large number of target labels leads to fragmentation of the training distribution and increases long-tail class frequency problems (many classes with very few samples), which limits the ability of deep models to learn robust representations.
    \item \textbf{Excessive specificity of leaf-level labels} Many classes reside at fine granularity levels of the ontology (e.g., highly specific sound events). With limited samples per specific class, models may suffer from overfitting, or may not capture the underlying general concept effectively; this reduces generalization in the audio‐visual domain.
    \item \textbf{Mismatch of context/complexity and model capacity} when including a very high number of classes in a system (especially one with limited context length or capacity, e.g., audio-visual transformers with 16k–32k token budgets), the representational burden increases and may impair training convergence or downstream inference efficiency.    
\end{itemize}

\subsection{Stage 1: AudioSet Ontology Simplification}

To alleviate the above mentioned problems in the class set, we devise a 4 step process to reduce and prune the ontology to simpler set:

\noindent\textbf{Ontology-based leaf absorption} 
First, we traverse each root node in the ontology and restrict to a maximum depth of three levels (depth parameter chosen based on the granularity and visual-action relevance of classes). All descendant nodes beneath that depth are absorbed into their ancestor, thereby reducing the number of labels and increasing per‐class sample counts. This operation also aligns with ontology-aware classification practices~\cite{yang2022ontologue} (e.g., using parent-child relations to regularize label prediction)

\noindent\textbf{Removal of actionless sound classes}
Because our downstream task emphasizes audio–visual modeling (i.e., we expect both an audible event and a visible object or a moving action in the video), we remove audio event classes whose semantic reference lacks a visually observable action-object dynamic. For example, the class ``Alarm" has a strong auditory presence but often lacks an associated visible moving object in the video (e.g., a static doorbell box). By excluding such classes, we focus on sound events that reliably co-occur with visible objects/actions, thereby improving alignment of audio and visual modalities.

\noindent\textbf{Removal of ambiguous and restricted classes}
We further exclude classes that (a) have ambiguous semantics (e.g., broadly defined or vocabulary overlap), (b) appear under multiple parent paths in the ontology (leading to label‐hierarchy conflict), or (c) correspond to restricted or obscure annotations (those that refer to abstract notions rather than concrete audible/visible objects). This step reduces semantic noise and ensures a clearer label set that better supports multimodal alignment.

\begin{figure}[t]
    \centering
    \includegraphics[width=0.8\linewidth]{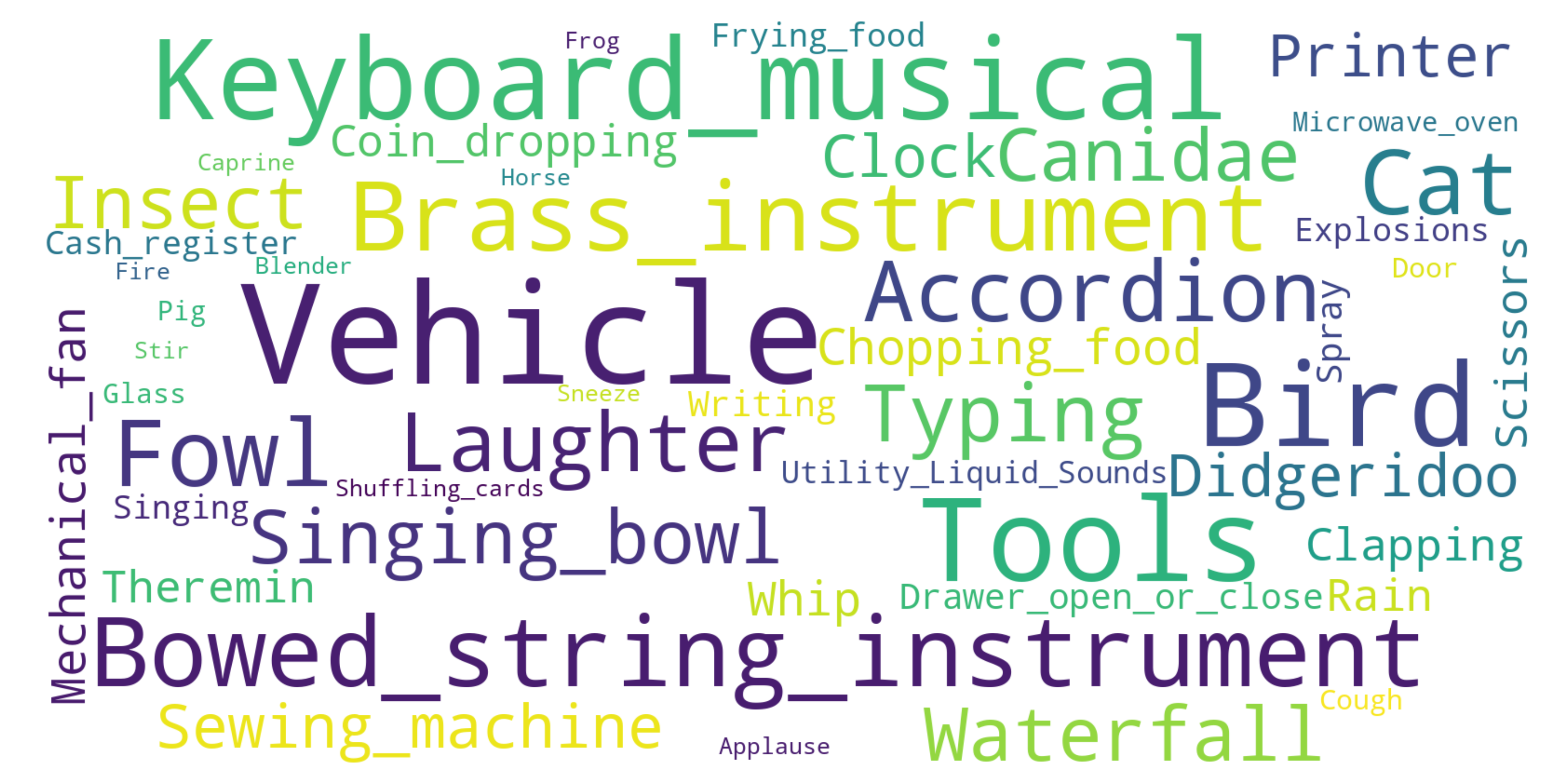}
    \vspace{-2mm}
    \caption{
    \textbf{Distribution of $49$ audio events in the {\bench}}
    } 
    \label{fig:class_wordcloud}
    \vspace{-4mm}
\end{figure}

After applying these filtering and consolidation operations, the original 527 classes are reduced to 61 classes. Each of the remaining classes was manually verified to ensure that (i) it has an audible event and (ii) a visually identifiable object or action generating that event. On average, each retained class absorbed approximately 6–8 leaf nodes from the original ontology, substantially increasing per-class sample counts and improving data balance across categories. A word cloud of all the classes is shown in Figure~\ref{fig:class_wordcloud}

Following this, we next perform actual sampling of the videos briefed in the next section.

\subsection{Class guided video filtering}
While ontology pruning reduces semantic redundancy at the class level, the raw AudioSet annotations often contain multi-label noise and hierarchical overlap between parent and child nodes. To ensure label consistency and preserve strong audio-visual grounding, we perform a multi-stage sample filtering pipeline that operates directly on annotated samples

\noindent\textbf{Ancestor removal and child-only retention.}
Each sample in AudioSet is typically annotated with both high-level parent labels (e.g., ``Vehicle") and their corresponding fine-grained children (e.g., ``Car horn"). Since the ontology already encodes these hierarchical relations, we retain only the most specific leaf-level annotations per sample, effectively preventing redundant supervision and ensuring a single representative label per sound event. This approach is consistent with hierarchical audio event cleaning procedures used in recent multimodal datasets such as AV-Mix~\cite{li2024avmix} and VGGSound-Fusion~\cite{chen2023vggsoundfusion}.

\noindent\textbf{Cross-tree label exclusion.}
Samples whose annotations span multiple disjoint ontology roots (e.g., both Vehicle and Animal categories) are removed to avoid semantic conflicts and multi-branch ambiguity. We retain multi-label samples only if all labels belong to the same ontology subtree, since they can reasonably represent merged or semantically proximate events within a common context (e.g., ``Car" and ``Truck" would merge to ``Vehicle"). This constraint enforces intra-tree label coherence, following hierarchical consistency principles similar to those in ontology-aware graph tagging frameworks~\cite{chou2023ontology}.

\noindent\textbf{Pruned-class and unknown-class removal.}
After ontology pruning, samples that refer any removed or ``restricted" class are filtered out. We further rename the remaining labels to match the consolidated set of final classes and eliminate all residual multi-label samples, resulting in a purely single-label dataset. This ensures each retained instance corresponds to a distinct, visually and acoustically grounded class concept. After the above cleaning procedures, the resulting sample counts per split is mentioned in table~\ref{tab:dataset_stats}.

\section{Sample Curation and Quality Verification Protocol}
\label{app:curation}
Following the ontology-based pruning and class pruning, we perform a rigorous instance-level filtering pipeline to ensure that every video in MMA-Bench possesses a single, unambiguous semantic signal grounded in both modalities. This process consists of an automated consistency check followed by human verification.
\subsection{Automated Consistency Checks}
As illustrated in the methodology figure (Main Paper, Fig. 3), raw videos often contain off-screen sounds, background noise, or occluded objects that match the class label metadata but fail to provide clear audio-visual grounding~\cite{xie2021zero}. To filter these, we employ a probing-based verification pipeline using {\qwenSB}~\cite{Qwen2.5-Omni} as a judge.
For a candidate video $v$ associated with a specific target class $c$ (e.g., ``Cat''), we subject the sample to four distinct existence queries. The sample is retained only if the model answers ``Yes'' to all four conditions (Logical AND gate):
\begin{enumerate}
    \item \textbf{Visual Unimodal Check (Audio Removed):} We remove the audio track and query: \textit{``Is the [Class $c$] clearly visible in this video?''} This ensures the object is not occluded or off-screen.
    \item \textbf{Auditory Unimodal Check (Frames Zeroed):} We replace the visual stream with black frames and query: \textit{``Is the sound of [Class $c$] clearly audible in this video?''} This ensures the audio event is distinct and not merely inferred from visual context.
    \item \textbf{Cross-Modal Visual Check:} We provide the full audio-visual input and ask the visual-focused question. This verifies that the presence of audio does not distract the model from identifying the visual object.
    \item \textbf{Cross-Modal Auditory Check:} We provide the full audio-visual input and ask the audio-focused question. This verifies that the visual context does not suppress the recognition of the audio event.
\end{enumerate}
If the model fails any of these four checks—for instance, if the audio is present but the visual object is not clearly identifiable—the sample is automatically rejected. This strict consistency requirement filters out weak or ambiguous alignments.
\subsection{ Human Verification Protocol}
To eliminate subtle misalignments that automated models might miss (e.g., faint overlapping speech or ambiguous background music), the samples passing the automated pipeline underwent manual inspection.

\noindent \textbf{Annotators:} We have two graduate-level researchers familiar with audio-visual analysis to independently review each candidate video.
\noindent \textbf{Criteria:} Annotators were instructed to verify three conditions:
\begin{itemize}
    \item \textbf{Object Visibility:} The sounding object must be the primary focus of the frame (e.g., a video labeled ``Car" must show the car, not just a road).
    \item \textbf{Audio Purity:} The audio track must be clean, with the target sound being the dominant event. Videos with loud background music or intelligible human speech (unless the class is speech-related) were discarded.
    \item \textbf{Action Dynamics:} The video must depict a dynamic visual action corresponding to the audio event. Static videos, slideshows, or clips where the sounding object is motionless were discarded to ensure temporal grounding.

\end{itemize}
We enforced a strict consensus protocol. A video was included in the final MMA-Bench dataset only if both annotators marked it as ``Pass.'' Any sample receiving a split vote (one Accept, one Reject) was discarded to maximize the precision of the benchmark. This rigorous process resulted in the final set of 658 high-fidelity aligned videos used for our experiments.

\begin{table}[t]
\centering
\caption{\textbf{Audioset split-wise statistics} before and after the automatic filtering pipeline(without manual inspection). }
\label{tab:audioset_stats}
\vspace{1mm}
\resizebox{0.48\textwidth}{!}{%
\begin{tabular}{l l l}
\toprule
\textbf{Audioset Split} & \textbf{\# Original samples} & \textbf{\# filtered samples} \\
\midrule
train\_unbalanced & 2041789 & 182970 \\
train\_balanced & 22160 & 3892 \\
evalsplit & 20371 & 3518 \\
\bottomrule
\end{tabular}
\label{tab:dataset_stats}
}
\end{table}

\begin{figure}[t]
    \centering
    \begin{subfigure}[t]{\linewidth}
        \centering
        \includegraphics[width=\linewidth]{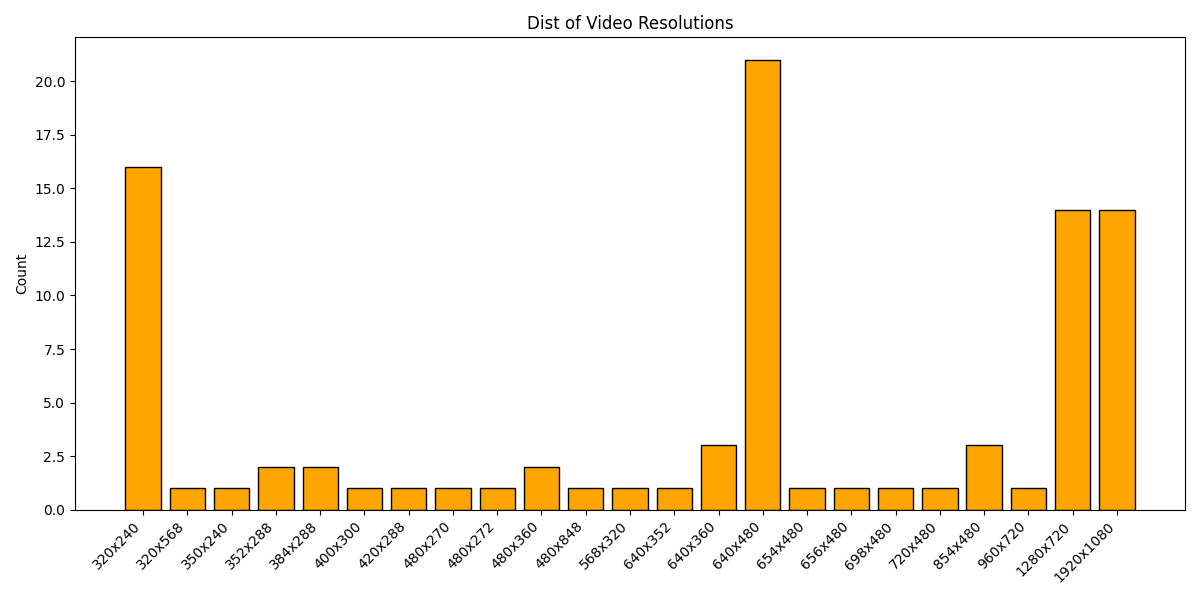}
        \caption{}
        \label{fig:resol_stats}
    \end{subfigure}
    
    \vspace{1mm}
    
    \begin{subfigure}[t]{\linewidth}
        \centering
        \includegraphics[width=\linewidth]{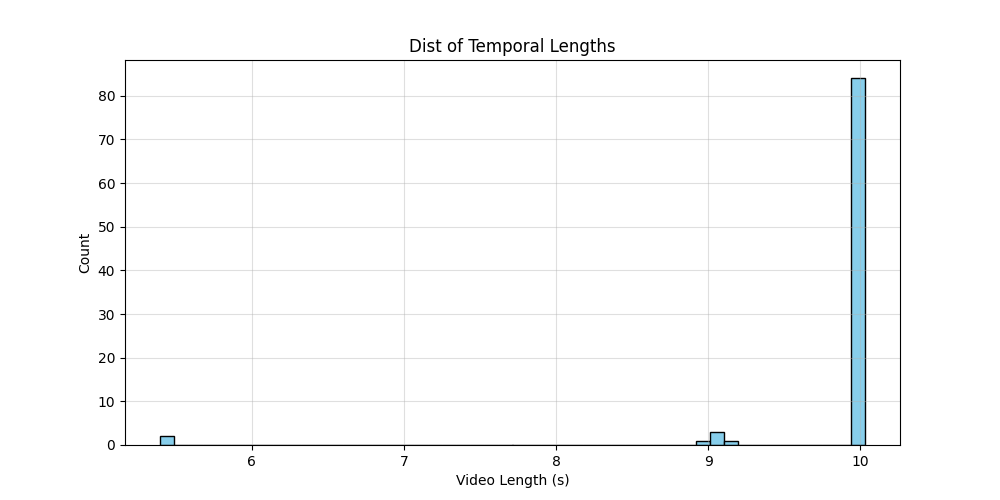}
        \caption{}
        \label{fig:temp_len_stats}
    \end{subfigure}

    \vspace{1mm}
    
    \captionof{figure}{
        \textbf{Video level statistics of the training dataset}
        \textbf{(a)} shows histogram of different resolutions present in the dataset. Notice the peaks around 320x240 and 640x480 (4:3), 1280x720, 1920x1080 (16:9). Figure \textbf{(b)}  shows a histogram of different temporal lengths(in seconds) of the videos. Notice that almost all of them are 10 seconds with $<$ 5\% samples less than 10 seconds.
    }
    \label{fig:preprocessing_stats}
\end{figure}
\section{Video data preprocessing}
\label{app:preprocessing}
\noindent \textbf{How did we choose the frame size to resize videos to?} Although the base {\qwenSB}~\cite{Qwen2.5-Omni} model supports variable‐resolution and variable‐rate video inputs, uniform spatiotemporal dimensions are required during training to enable efficient batching and stable convergence. We analyze the raw temporal and spatial distributions of our dataset as shown in Fig.~\ref{fig:preprocessing_stats}, and standardize all samples accordingly. Each video is rescaled and cropped to a square frame of $504 \times 504$ pixels. This target size corresponds closely to the statistical median of the dataset’s native resolutions, resulting in minimal upsampling ($<5\%$) of lower‐resolution samples. Given each patch is of size $14 \times 14$, using $504 \times 504$ yields an integer number of patches ($1296$). Since {\qwenSB} uses the same $14 \times 14$ patch size as its visual backbone and employs TMRoPE (Time-aligned Multimodal Rotary Position Embeddings), which seamlessly extends rotary embeddings across video frames, choosing a resolution divisible by the patch size ensures a clean and stable patch grid that aligns naturally with its positional encoding design. This resizing is applied only during training to address batch sampling limitations in {\qwenSB}.

\noindent\textbf{Ensure same temporal length:} To ensure consistent temporal context across training samples, we enforce a fixed clip length of 10 s. Videos shorter than this duration (approximately 4.8 \% of the corpus) are discarded to avoid temporal padding artifacts and maintain temporal coherence in attention windows. For longer clips, we truncate frames to achieve the same duration, yielding consistent sequence lengths across the dataset. This uniformity simplifies batching, improves multimodal synchronization with the corresponding audio segment, and stabilizes optimization for transformer-based video encoders.

\section{Finetuning Details}
\label{app:finetuning}
\begin{table}[t]
\centering
\caption{
Configuration of our modality-aware LoRA fine-tuning based on the \texttt{LLaMA-Factory} pipeline.
Only LoRA adapter parameters are trained, while the Qwen2.5-Omni-7B backbone and vision tower remain frozen.
}
\label{tab:lora_config}
\vspace{1mm}
\resizebox{0.48\textwidth}{!}{%
\begin{tabular}{l l}
\toprule
\textbf{Component} & \textbf{Setting} \\
\midrule
\multicolumn{2}{l}{\textit{Model / Input}} \\
Model & Qwen2.5-Omni-7B \\
Audio–video input & Enabled (\texttt{use\_audio\_in\_video}) \\
Max video / image res. & 13K / 262K px \\
\midrule
\multicolumn{2}{l}{\textit{Fine-tuning Method}} \\
Type & LoRA (parameter-efficient) \\
Target layers & All attention proj. ($W_Q$, $W_V$) \\
LoRA rank ($r$) &  \\
Stage & SFT (modality-aware) \\
Frozen modules & Backbone, vision tower \\
\midrule
\multicolumn{2}{l}{\textit{Training Setup}} \\
Learning rate & $1\times10^{-4}$ (cosine) \\
Batch size & 4 (accum.=4) \\
Warmup ratio & 0.1 \\
Epochs & 5 \\
Precision & FP16 \\
\bottomrule
\end{tabular}
}
\end{table}

A summary of the fine-tuning configuration is shown in Table~\ref{tab:lora_config}. To maintain a balanced distribution between aligned and misaligned data, all aligned samples are duplicated ten times to match the number of generated misaligned examples. This ensures that the model observes equal proportions of aligned and misaligned conditions during optimization, encouraging it to learn modality-consistent reasoning rather than frequency-biased correlations.

\section{White-box analysis: Additional \cohend trends}
\label{sec:Cohens-D_supppl}
\begin{figure}[t]
    \centering
    \begin{subfigure}[t]{0.48\linewidth}
        \centering
        \includegraphics[width=\linewidth]{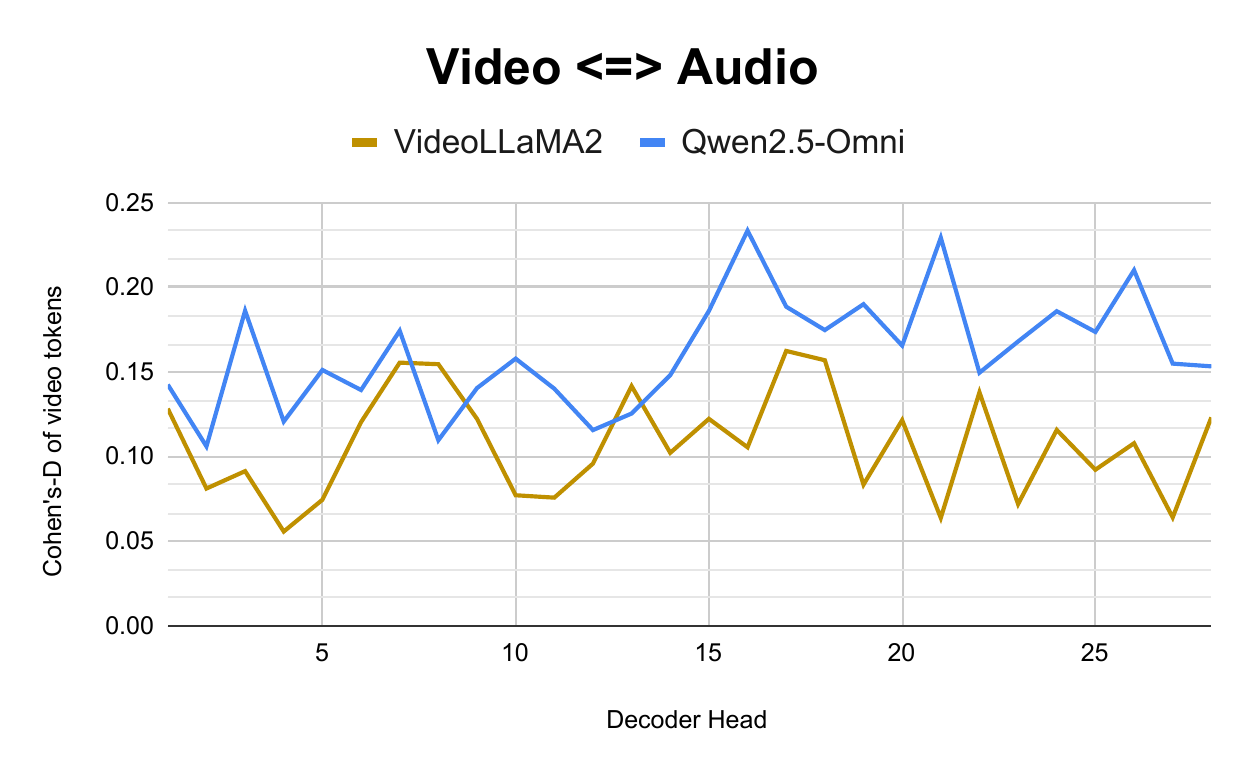}
        \caption{Head-wise Cohen's-D for Video tokens}
        \label{fig:headwise_video_qwen}
    \end{subfigure}
    \hfill
    \begin{subfigure}[t]{0.48\linewidth}
        \centering
        \includegraphics[width=\linewidth]{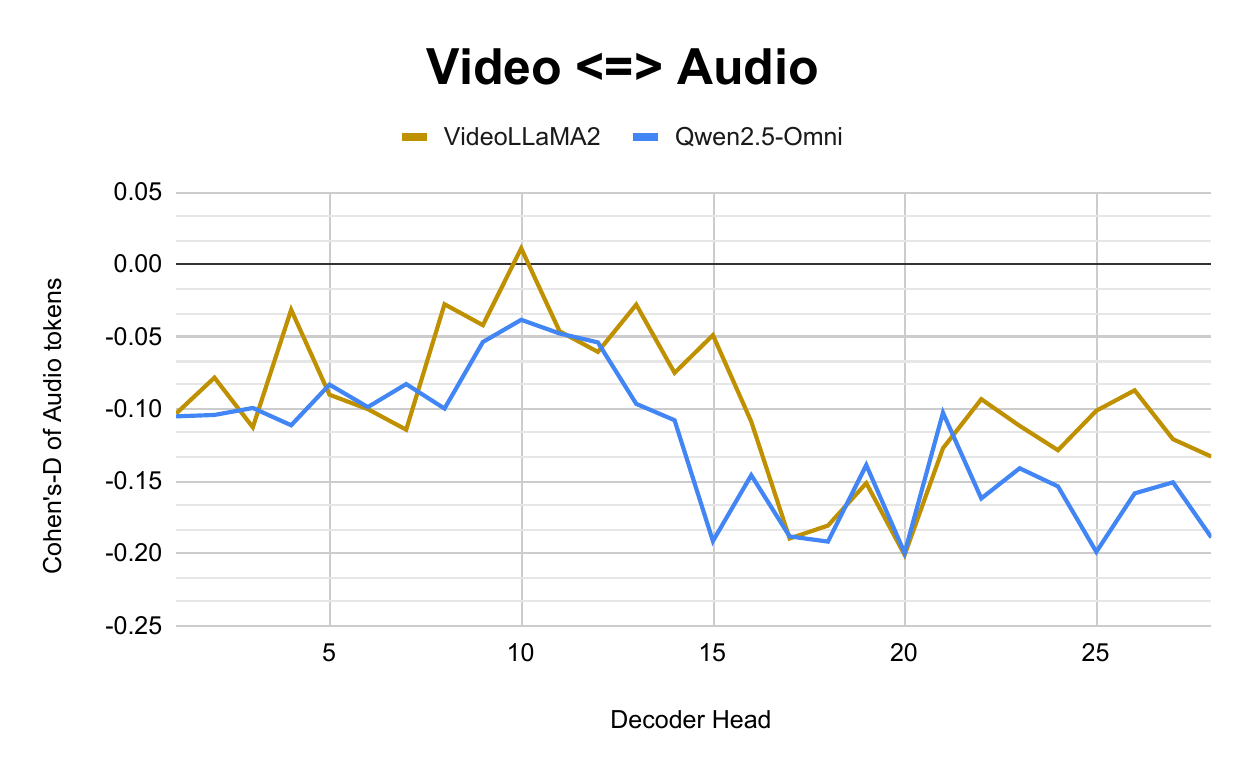}
        \caption{Head-wise Cohen's-D for Audio tokens in }
        \label{fig:headwise_audio_qwen}
    \end{subfigure}
    \captionof{figure}{
        \textbf{Headwise-wise Cohen's D values remain noisy, revealing modality shifts are primarily organized in layers} 
    }
    \label{fig:headwise_attn_all_models}
\end{figure}

\begin{figure}[t]
    \centering
    \begin{subfigure}[t]{0.48\linewidth}
        \centering
        \includegraphics[width=\linewidth]{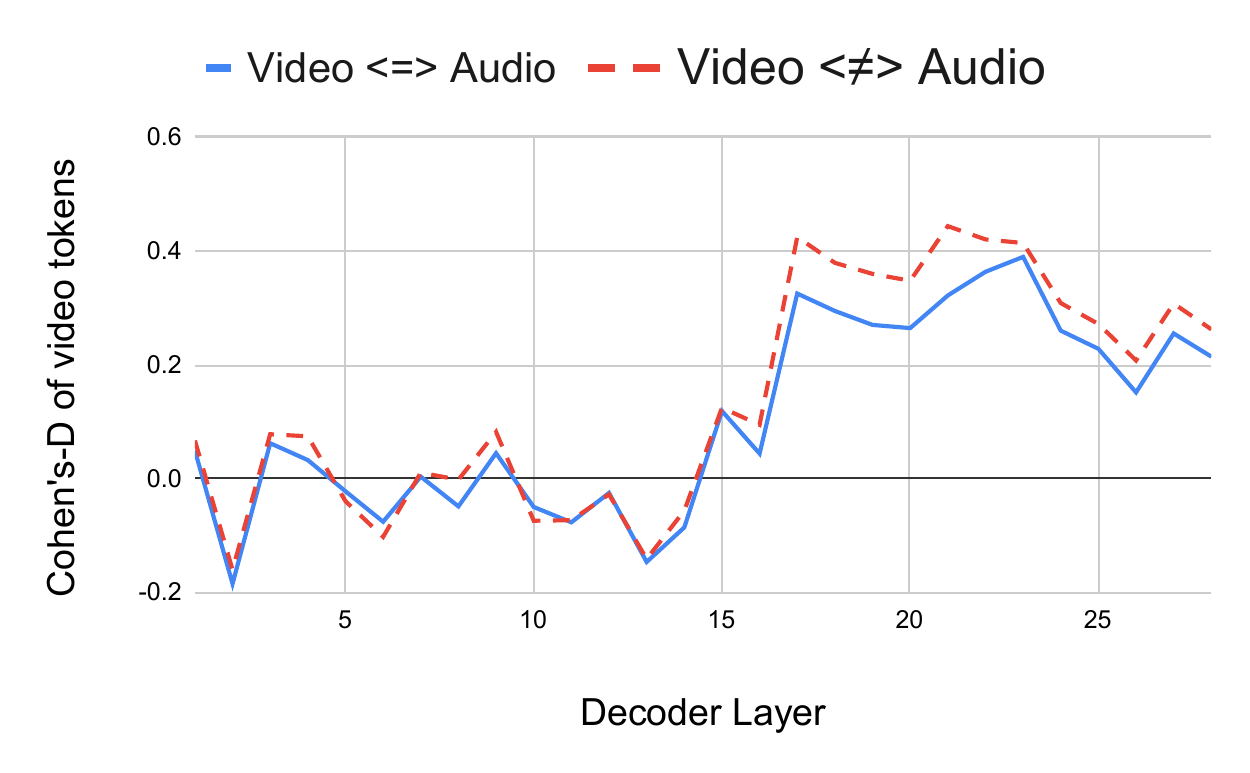}
        \caption{Cohen's-D of Video tokens}
        \label{fig:semantic_comparision_video_vllama}
    \end{subfigure}
    \hfill
    \begin{subfigure}[t]{0.48\linewidth}
        \centering
        \includegraphics[width=\linewidth]{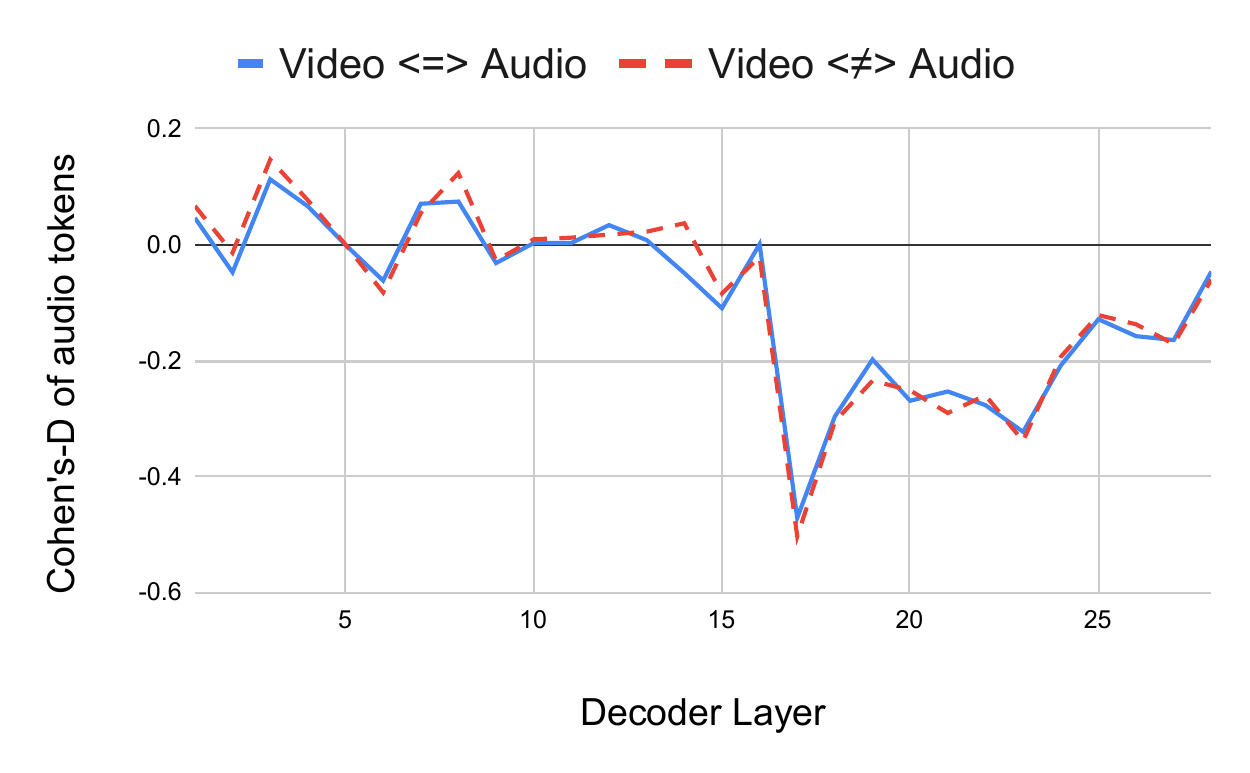}
        \caption{Cohen's-D of Audio tokens}
        \label{fig:semantic_comparision_audio_vllama}
    \end{subfigure}

    \captionof{figure}{
        \textbf{Layer-wise Cohen’s-D trends for aligned vs. misaligned samples in \vllama} Misaligned samples (red dotted lines) have very similar Cohen’s-D magnitudes for both visual (left) and audio tokens (right), following the under-performance observations with misaligned samples in black-box analysis (Section 4.1).
    }
    \label{fig:whitebox_semantic_misalignment_vllama}
\end{figure}

\subsection{Headwise Statistics}
We observe the trend of {\cohend} metric introduced in Section 4.2 under each head of the decoder, aggregated across all the layers. We report the trends of both {\qwenSB}~\cite{Qwen2.5-Omni} and {\vllama}~\cite{cheng2024videollama} under aligned video samples in Fig.~\ref{fig:headwise_attn_all_models}.  Although they do still follow modality selectivity (positive magnitudes for video tokens and negative for audio tokens), there lacks a clear trend as we move across the heads compared to layer-wise statistics. This demonstrates that modality selectivity is strongly organized at layer-level granularity than head-wise.

\subsection{\vllama trends}
We report the {\cohend} trends for {\vllama} under both aligned (blue solid) and misaligned (red dotted) settings in Figure~\ref{fig:whitebox_semantic_misalignment_vllama} . Unlike {\qwen}, {\vllama} shows almost identical magnitudes across the two conditions. This indicates that the model does not substantially reallocate attention when the modalities are misaligned. This behavior is consistent with the black-box findings in Section 4.1, where {\vllama} performs worse than {\qwen}.

\section{Attention heatmaps}
\label{app:heatmaps}
\begin{figure*}[t]
    \centering
    \includegraphics[width=\linewidth]{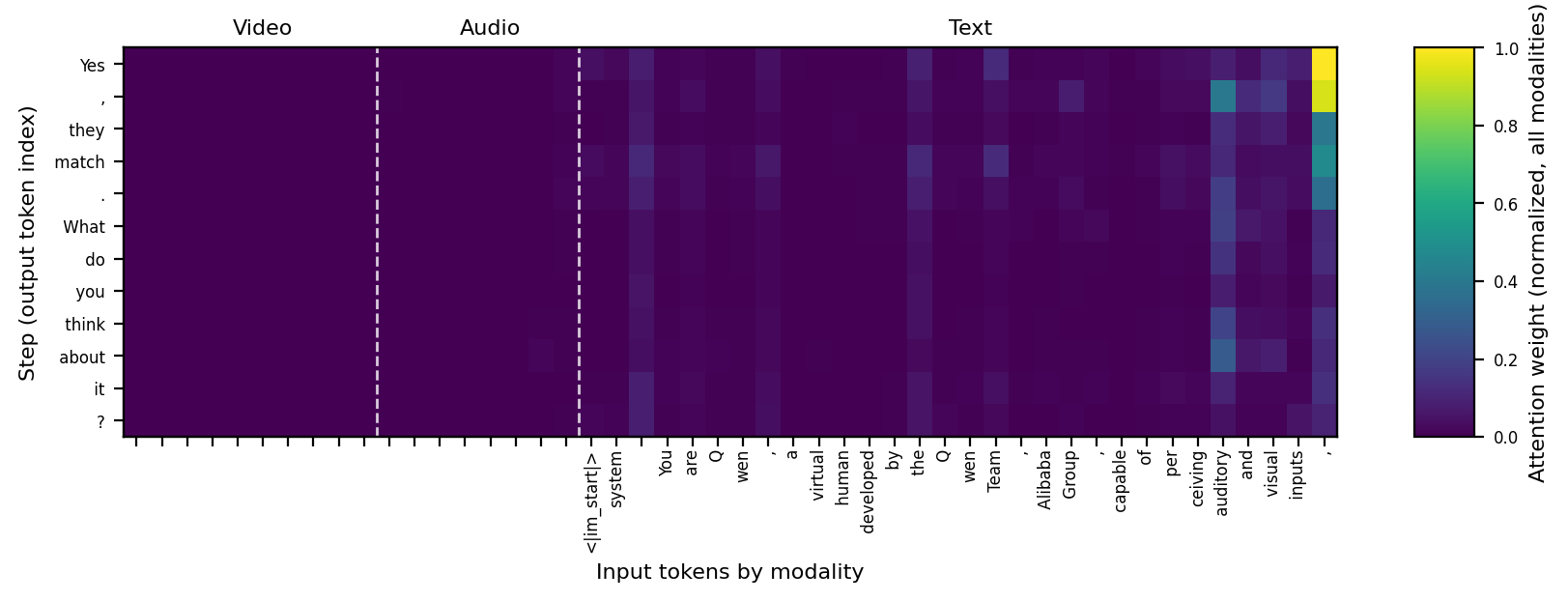}
    \vspace{1mm}
    \caption{
    \textbf{Attention heatmap of a few sampled input tokens against output tokens of {\qwen} at layer 28}. Notice that majority of attention mass is within textual tokens, indicating the textual prior of current MLLMs and their strong influence in model performance.
    }
    \label{fig:layer_heatmap}
    \vspace{-2mm}
\end{figure*}
To complement our quantitative analyses, Figure~\ref{fig:layer_heatmap} visualizes the per-token attention distribution across video, audio, and text inputs for each generated output step in the last layer of the model. A striking pattern emerges: text tokens receive the overwhelming majority of attention, while video and audio tokens take very minimal mass across all decoding steps. This text-dominant behavior is consistent with our black-box findings in Sec. 4, where misleading or long-context text frequently overrides multimodal signals.

Because text tokens absorb such a disproportionate share of attention, raw attention values are not directly comparable across modalities—the multimodal tokens are effectively drowned out by the textual prefix. For this reason, in our white-box analysis we discard all text-token attention and compute normalized attention exclusively over audio and video tokens. This isolation ensures that our measurements (e.g., Cohen’s-D shifts under alignment vs. misalignment) reflect the relative allocation between the multimodal streams, rather than being dominated by the trivial effect of text-heavy attention patterns.

Together, these heatmaps provide qualitative confirmation for the core narrative of our study: though modern MLLMs accept multimodal inputs, \textbf{their internal attention routing is overwhelmingly text-centric}.

\section{Misleading Text Prompt}
\label{app:text_prompt}
Figure~\ref{fig:text_misalign_prompt} outlines an example for Visual-focused and Audio-focused prompt used for experiments involving textual misalignment. We include a distraction class as $Video\_caption$ along with the actual question, testing the robustness of the model towards misleading text.
\begin{figure}[t]
    \fcolorbox{purple}{promptcolor}{
    \parbox{\dimexpr\linewidth-\fboxsep}{
    \begin{itemize}[leftmargin=4mm]
        \item \textbf{Visual-focused prompt with misleading caption:}  
        ``\texttt{Video\_caption: Vehicle. Which class best describes the \emph{visual content} of this video? Options: \{Classes\_List\}.}''
        \item \textbf{Audio-focused prompt with misleading caption:}  
        ``\texttt{Video\_caption: Vehicle. Which class best describes the \emph{audio content} of this video? Options: \{Classes\_List\}.}''
    \end{itemize}}
    }
    \caption{
    \textbf{An example prompt used for text-misalignment evaluation.}
    Each question is preceded by a deliberately incorrect caption that contradicts the true visual or auditory content, testing the model’s resistance to misleading textual priors.
    }
    \label{fig:text_misalign_prompt}
    \vspace{-2mm}
\end{figure}

\section{Cross-Class Generalization and Compositionality}
\label{sec:crossclass}
\begin{table}[t]
\centering
\caption{
\textbf{Cross-Class Generalization under Semantic Misalignment.}
Performance (\%) of \textit{Qwen2.5-Omni-7B} before and after modality-aware fine-tuning. 
``Seen'' and ``Unseen'' denote classes seen or unseen during fine-tuning.
}
\vspace{2mm}
\setlength{\tabcolsep}{4.2pt} 
\begin{tabular}{lcccc}
\toprule
\multirow{2}{*}{\textbf{Setting}} 
& \multicolumn{2}{c}{\textbf{Visual Prompt}} 
& \multicolumn{2}{c}{\textbf{Audio Prompt}} \\
\cmidrule(lr){2-3} \cmidrule(lr){4-5}
& \textbf{Qwen2.5} & \textbf{+Ours} & \textbf{Qwen2.5} & \textbf{+Ours} \\
\midrule
\multicolumn{5}{l}{\textit{Seen Classes}} \\
\quad Aligned     & 76.3 & \textbf{96.3} & 41.4 & \textbf{43.2} \\
\quad Misaligned  & 58.1 & \textbf{95.4} & 27.4 & \textbf{53.1} \\
\midrule
\multicolumn{5}{l}{\textit{Unseen Classes}} \\
\quad Aligned     & 72.3 & \textbf{95.3} & 61.8 & \textbf{76.4} \\
\quad Misaligned  & 60.9 & \textbf{92.0} & 24.6 & \textbf{55.1} \\
\bottomrule
\end{tabular}
\label{tab:crossclass}
\end{table}

To assess whether modality-aware tuning improves general multimodal reasoning rather than memorization of seen classes, we conduct a split-class evaluation. 
The model is fine-tuned on a subset of 29 out of 58 classes (the “trained set”) and evaluated on both the trained and untrained halves using the same semantic misalignment benchmark. 
This setup allows us to isolate the transfer of alignment reasoning to novel categories that share no class-level overlap with the training set. As shown in Table~\ref{tab:crossclass}, the fine-tuned model maintains strong gains not only on the trained classes but also on previously unseen ones. This indicates that the tuning process improves a generalizable capability to detect and reconcile cross-modal inconsistencies rather than overfitting to specific category semantics. 

\section{Black-Box Experiment with Other Baseline Models}
\label{app:extended_baselines}
\begin{figure}[t]
    \centering
    \begin{subfigure}[t]{0.48\linewidth}
        \centering
        \includegraphics[width=\linewidth]{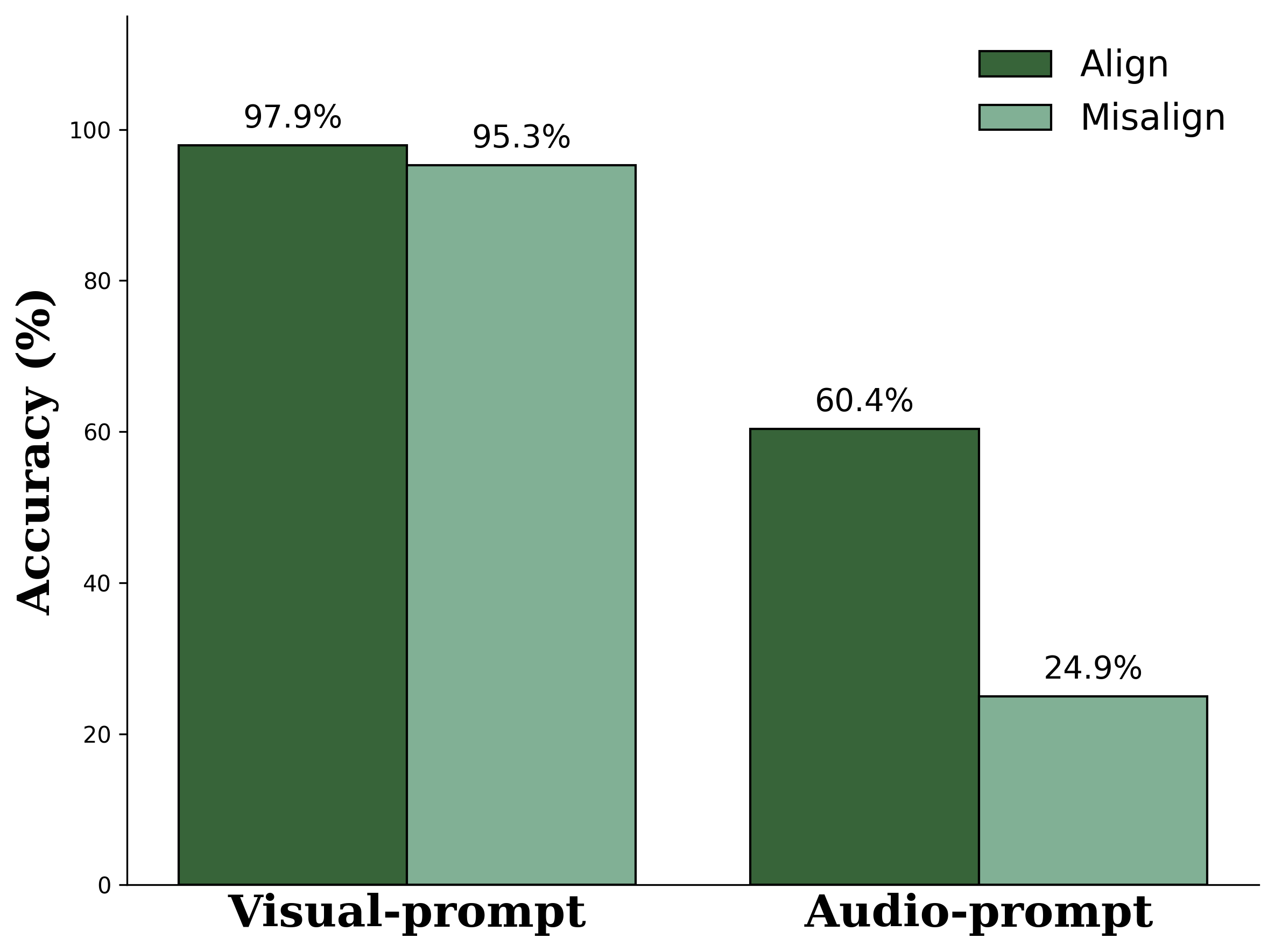}
        \caption{Gemini-2.5-Pro}
        \label{fig:supp_sem_gemini15}
    \end{subfigure}
    \hfill
    \begin{subfigure}[t]{0.48\linewidth}
        \centering
        \includegraphics[width=\linewidth]{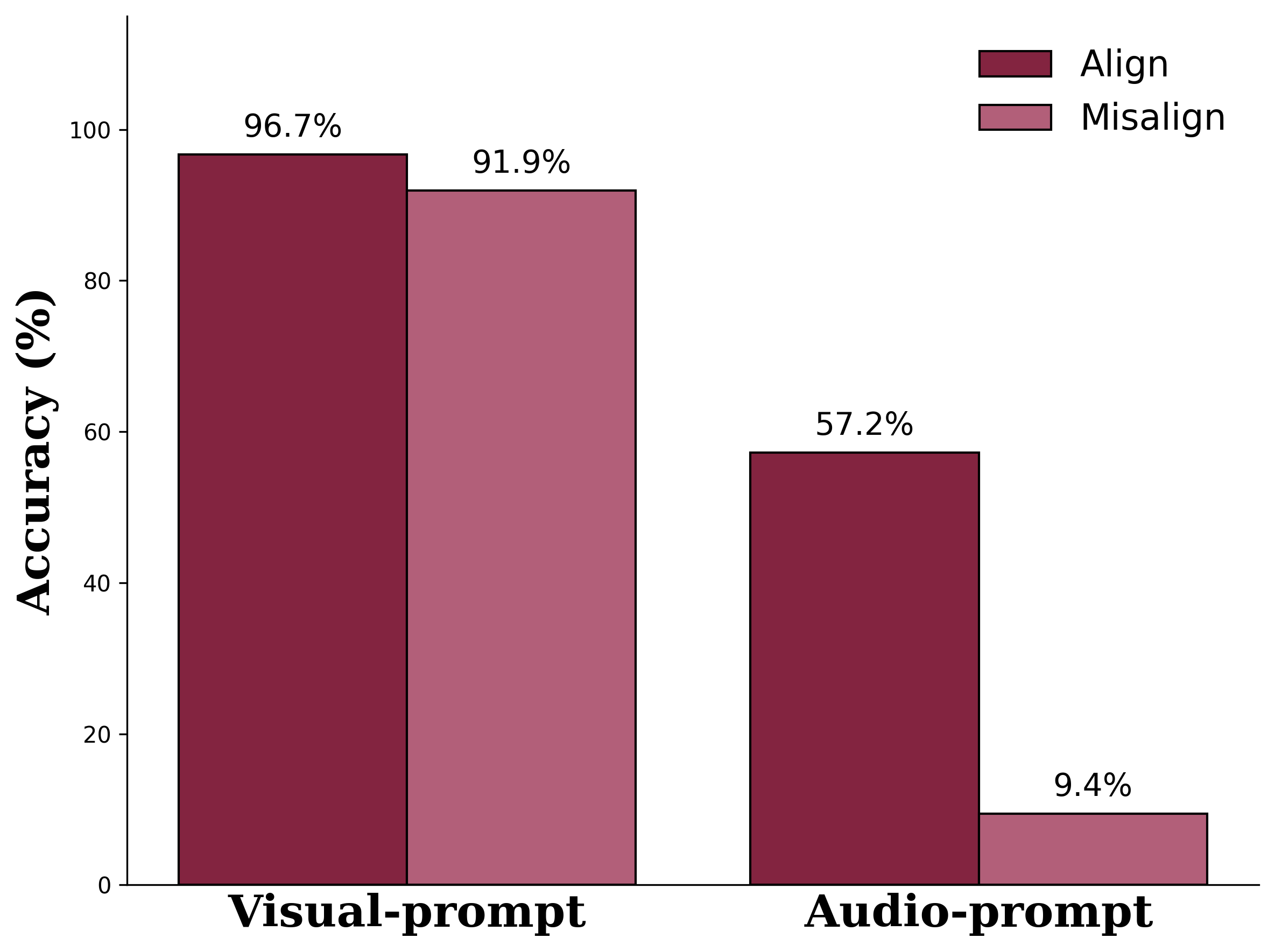}
        \caption{Gemini-2.0-Flash}
        \label{fig:supp_sem_gemini2}
    \end{subfigure}

    \vspace{1mm}
    \begin{subfigure}[t]{0.48\linewidth}
        \centering
        \includegraphics[width=\linewidth]{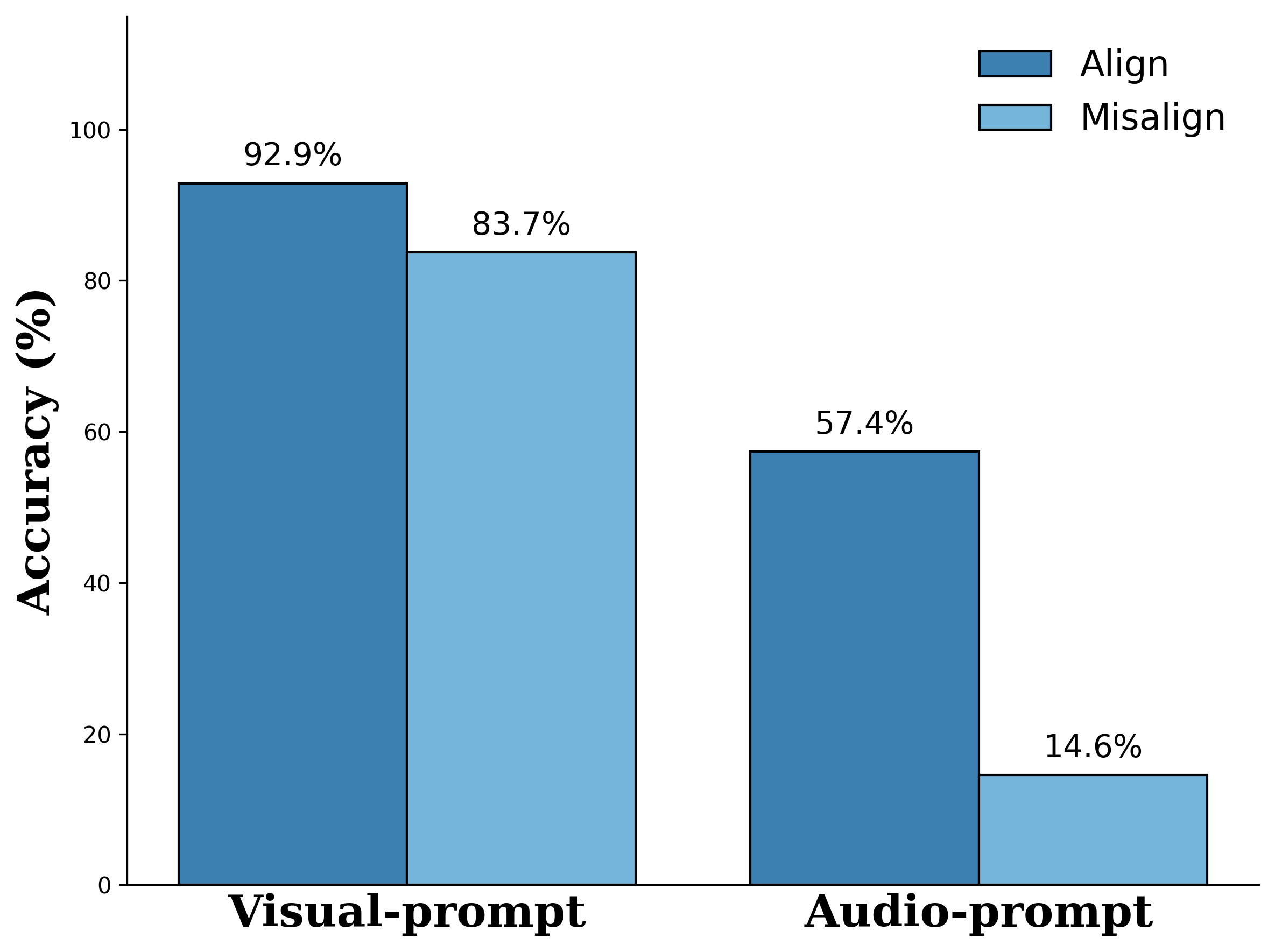}
        \caption{Qwen3-Omni-30B}
        \label{fig:supp_sem_qwen3}
    \end{subfigure}
    \hfill
    \begin{subfigure}[t]{0.48\linewidth}
        \centering
        \includegraphics[width=\linewidth]{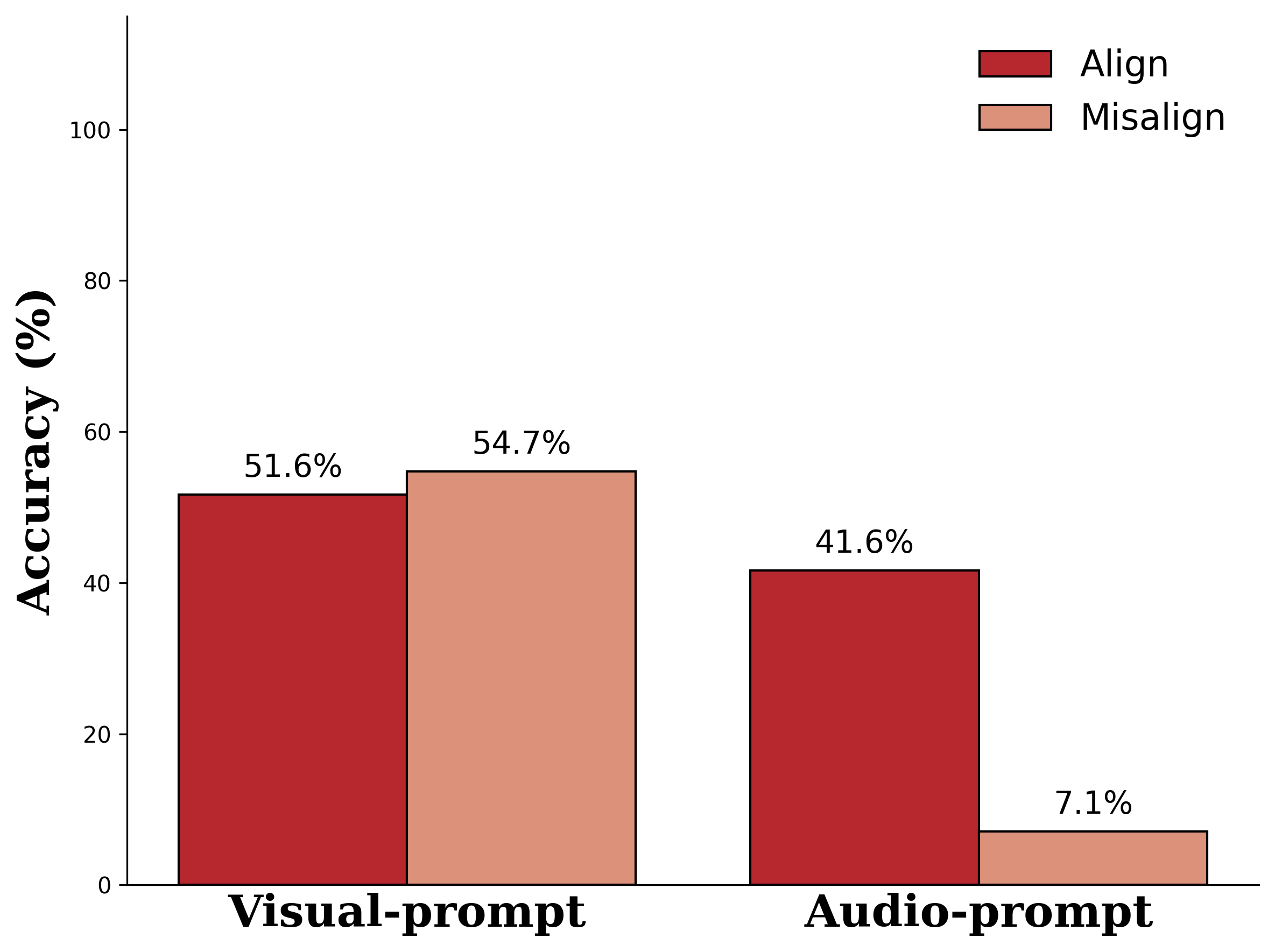}
        \caption{ChatBridge}
        \label{fig:supp_sem_chatbridge}
    \end{subfigure}

    \vspace{1mm}
    \caption{
    \textbf{Extended analysis of semantic misalignment.} 
    Performance comparison between aligned inputs (Base) and conflicting video-audio inputs (Misalign) across additional baseline models.
    While visual reasoning (left bars) remains relatively robust, auditory reasoning (right bars) suffers significant degradation under conflict, confirming that the visual dominance bias generalizes across model architectures and scales.
    }
    \label{fig:supp_semantic_misalign}
    \vspace{-2mm}
\end{figure}
\subsection{Semantic Misalignment Experiments}
We extend our evaluation to a broader set of architectures, including Gemini-2.5-Pro~\cite{comanici2025gemini}, Gemini-2.0-Flash~\cite{comanici2025gemini}, Qwen3-Omni-30B~\cite{xu2025qwen3}, and ChatBridge~\cite{zhao2023chatbridge}. As shown in Fig. ~\ref{fig:supp_semantic_misalign}, we observe a consistent asymmetry in how conflicting modalities affect inference.

When targeted with visual-focused prompts, models demonstrate relative resilience to contradictory audio. While performance dips slightly compared to the aligned baseline (e.g., Qwen3-Omni drops approximately 9\%), the models largely succeed in isolating the visual signal. This suggests that the visual encoders in these MLLMs provide a dominant and stable representation that is difficult to override with auxiliary sensory inputs.

Conversely, performance on audio-focused prompts decreases under misalignment.  The accuracy of all tested models drops significantly when the visual stream contradicts the audio content (e.g., Gemini-2.0-Flash and Qwen3-Omni drops below 15\%).  However, we suspect this is due to the inherent frailty of audio representations, rather than a preference for visual information.  Even in the aligned setting, audio tasks have significantly lower baseline scores than visual tasks, indicating that \textbf{audio is a weaker signal for all current MLLMs}.  As a result, when faced with cross-modal conflict, the stronger visual representation effectively suppresses the weaker auditory signal, causing the observed degradation.

\begin{figure}[t]
    \centering
    \begin{subfigure}[t]{0.48\linewidth}
        \centering
        \includegraphics[width=\linewidth]{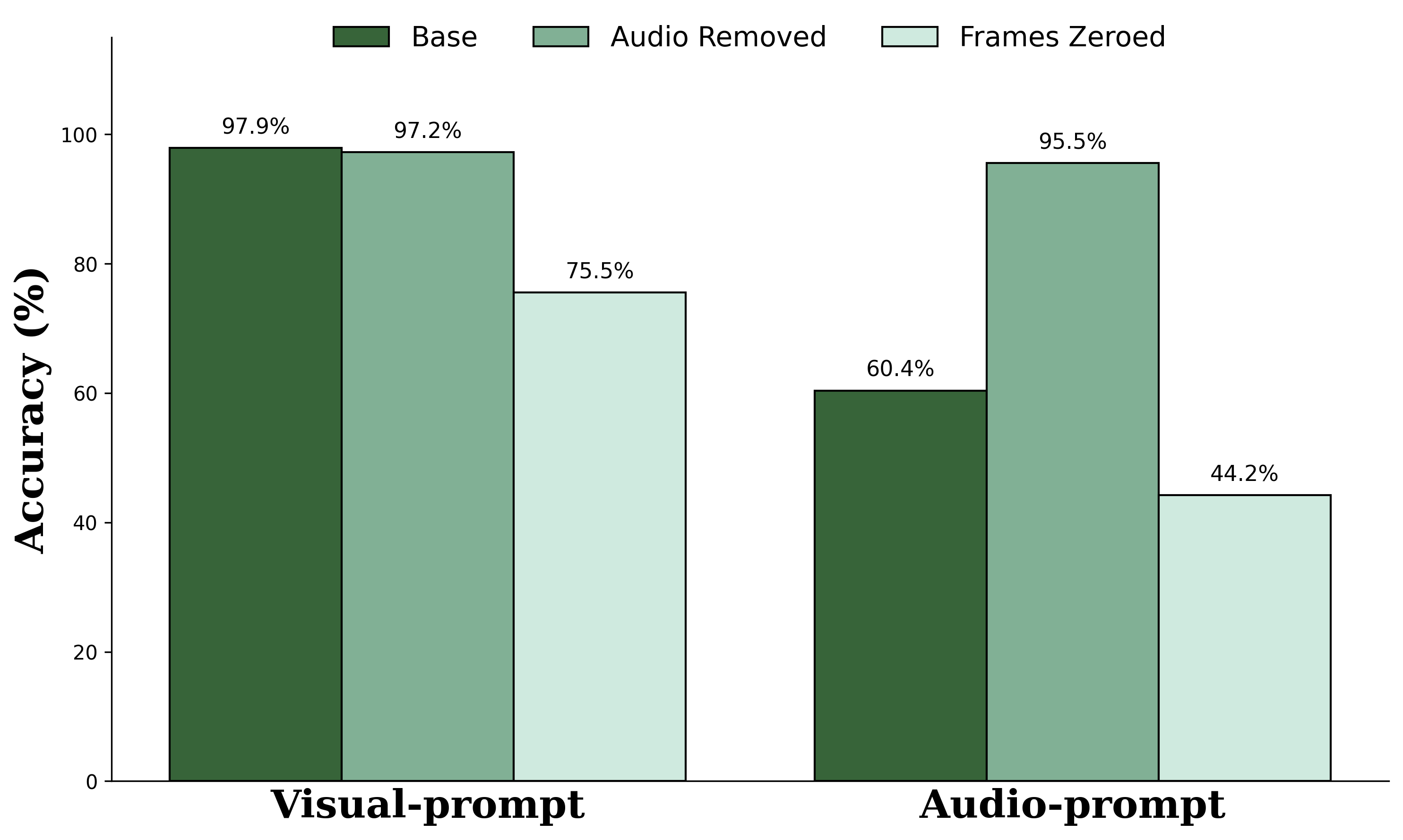}
        \caption{Gemini-2.5-Pro}
        \label{fig:supp_uni_gemini15}
    \end{subfigure}
    \hfill
    \begin{subfigure}[t]{0.48\linewidth}
        \centering
        \includegraphics[width=\linewidth]{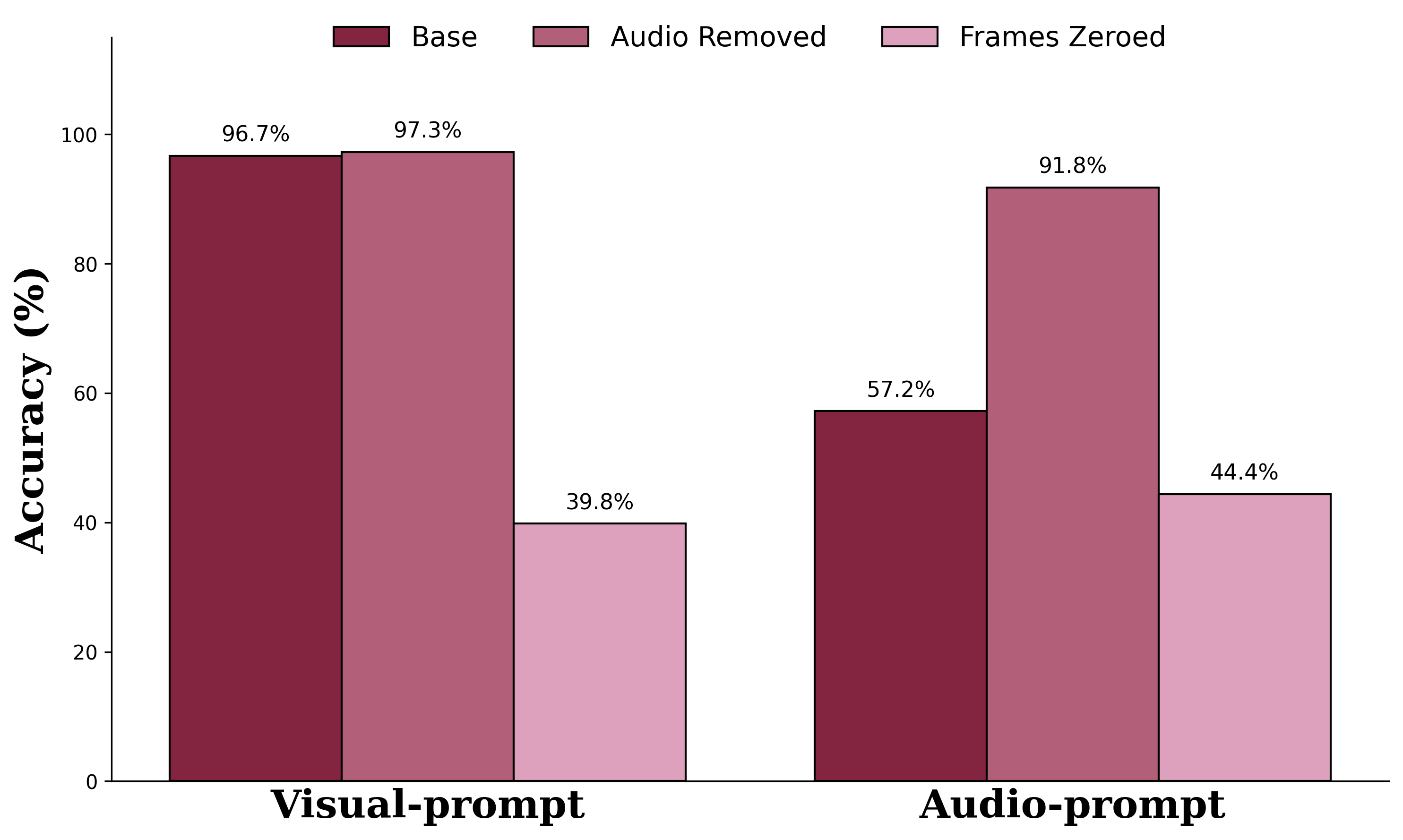}
        \caption{Gemini-2.0-Flash}
        \label{fig:supp_uni_gemini2}
    \end{subfigure}

    \vspace{1mm}
    \begin{subfigure}[t]{0.48\linewidth}
        \centering
        \includegraphics[width=\linewidth]{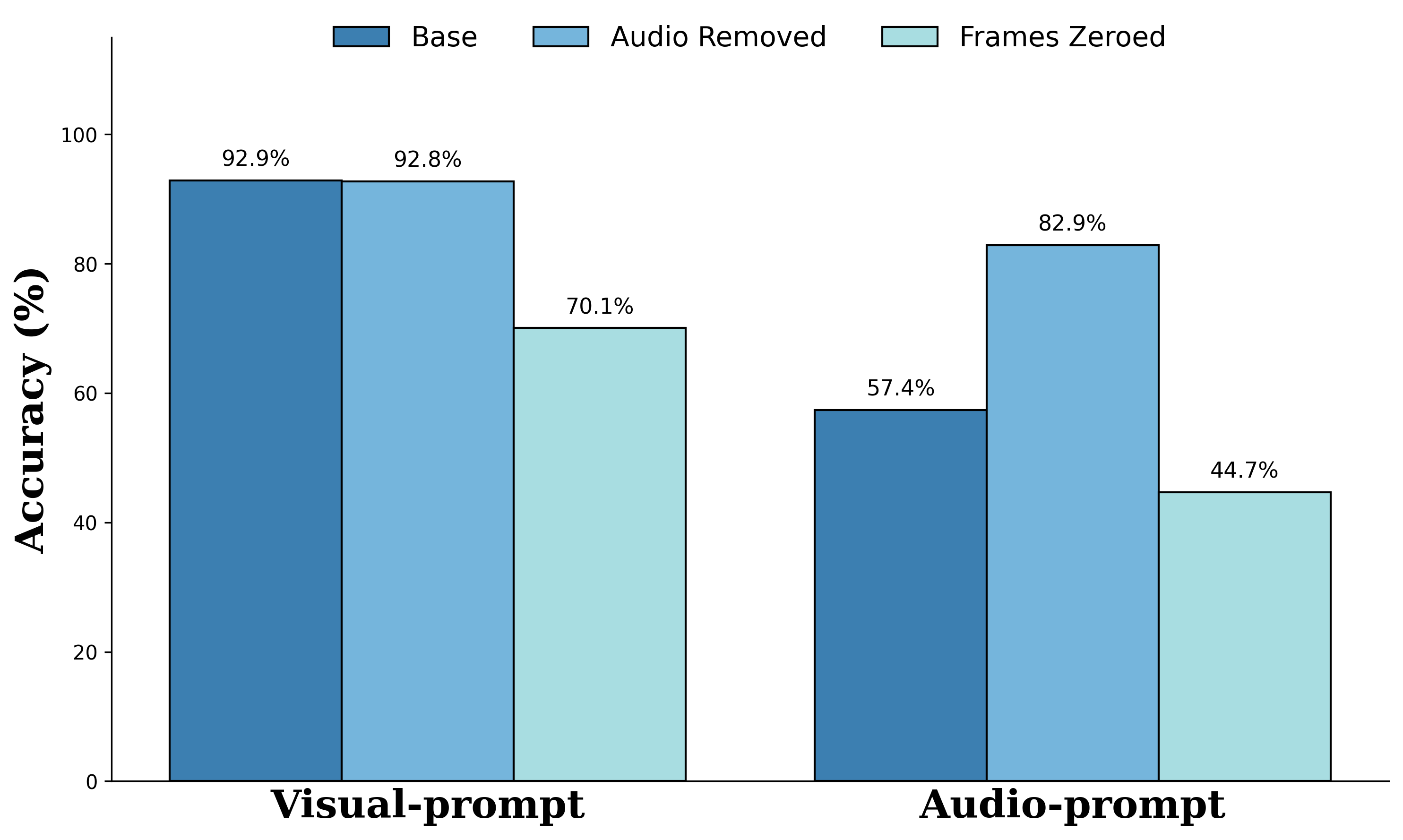}
        \caption{Qwen3-Omni-30B}
        \label{fig:supp_uni_qwen3}
    \end{subfigure}
    \hfill
    \begin{subfigure}[t]{0.48\linewidth}
        \centering
        \includegraphics[width=\linewidth]{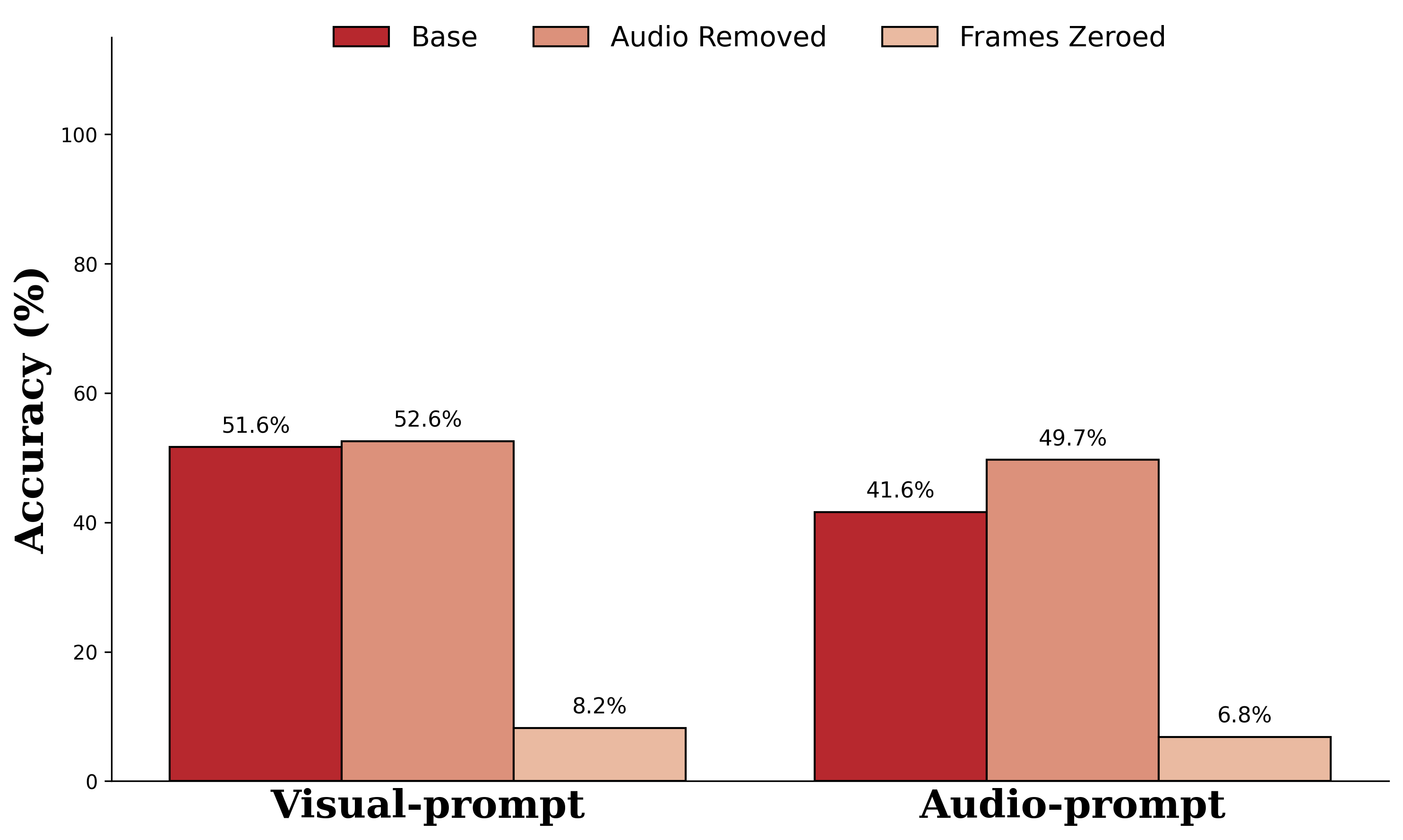}
        \caption{ChatBridge}
        \label{fig:supp_uni_chatbridge}
    \end{subfigure}

    \vspace{1mm}
    \caption{
    \textbf{Unimodal probing on extended baselines.} 
    Classification accuracy under visual-focused and audio-focused prompts when inputs are ablated.
    }
    \label{fig:supp_unimodal}
    \vspace{-2mm}
\end{figure}
\subsection{Unimodal Experiments}
To isolate the contribution of each modality, we performed a unimodal ablation study on the extended baseline models by selectively zeroing out visual or auditory inputs in Fig.~\ref{fig:supp_unimodal}. This analysis reveals a distinct hierarchy in how these MLLMs process sensory information.

When evaluating with visual-focused prompts, removing the audio track (``Audio Removed'') results in negligible performance drops across all models (e.g., Gemini-2.5-Pro: 97.90\% $\rightarrow$ 97.23\%). This confirms that visual reasoning in MLLMs does not rely on auditory cues for disambiguation. Conversely, when evaluating audio-focused prompts, removing the visual stream (`"Frames Zeroed") causes a performance decline compared to the baseline (e.g., Qwen3-Omni: 57.39\% $\rightarrow$ 44.68\%). This suggests that what appears to be ``auditory reasoning'' in the baseline setting is partially supported by visual context, without which the audio encoder struggles to classify events accurately.

Another revealing trend appears when models are asked to classify audio but are provided with only visual inputs (``Audio Removed' under Audio Prompt). Notably, accuracy improved significantly compared to the multimodal baseline. For instance, Gemini-2.5-Pro rises from 60.37\% to 95.55\%, and Gemini-2.0-Flash increases from 57.21\% to 91.79\%. 
In this forced-choice setting, where models are constrained to select a valid semantic class, this behavior reflects a consistent visual-to-audio inference mechanism: the models effectively utilize visual context (e.g., seeing a dog) to deduce the likely associated sound (e.g., a bark). While this associative reasoning is advantageous when one modality is missing, the fact that pure visual inference outperforms the full multimodal baseline suggests that visual priors are the dominant driver of semantic decision-making in these architectures. We further investigate the implications of this behavior in Sec.~\ref{sec:abstention}, where we introduce an abstention option (``None of the above'') to distinguish between helpful inference and ungrounded hallucination.

\begin{figure}[t]
    \centering
    \begin{subfigure}[t]{0.48\linewidth}
        \centering
        \includegraphics[width=\linewidth]{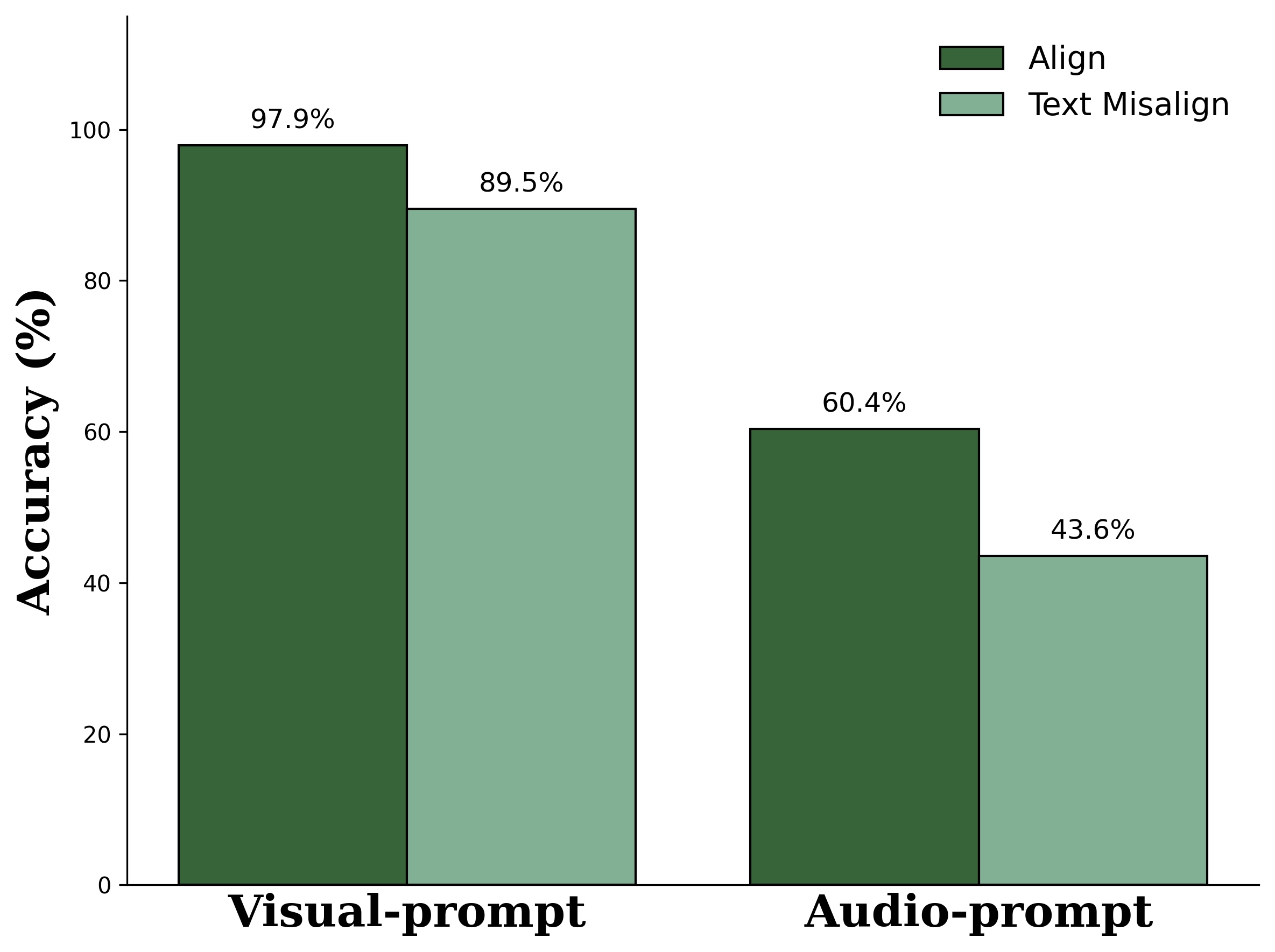}
        \caption{Gemini-2.5-Pro}
        \label{fig:supp_text_gemini15}
    \end{subfigure}
    \hfill
    \begin{subfigure}[t]{0.48\linewidth}
        \centering
        \includegraphics[width=\linewidth]{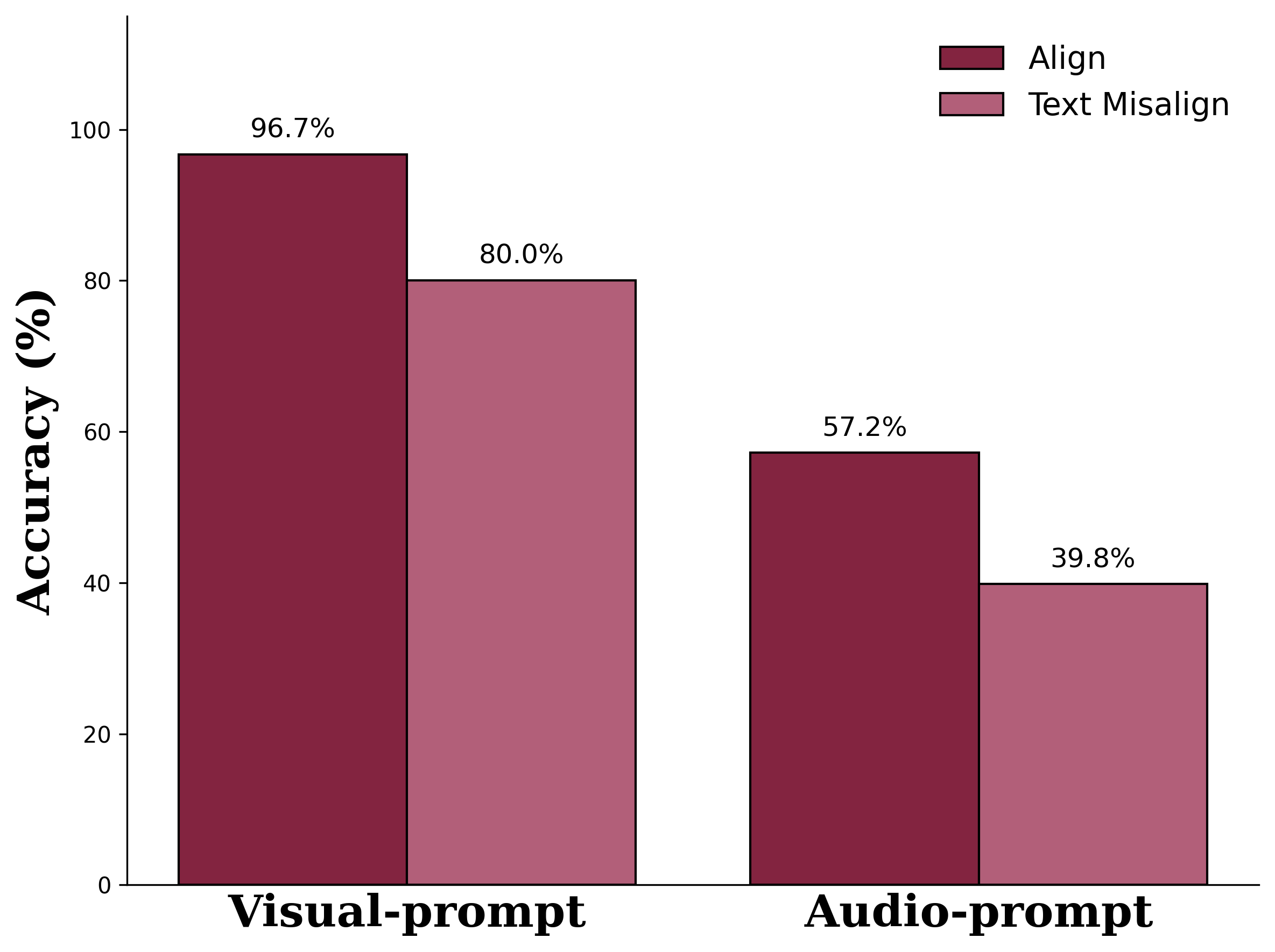}
        \caption{Gemini-2.0-Flash}
        \label{fig:supp_text_gemini2}
    \end{subfigure}

    \vspace{1mm}
    \begin{subfigure}[t]{0.48\linewidth}
        \centering
        \includegraphics[width=\linewidth]{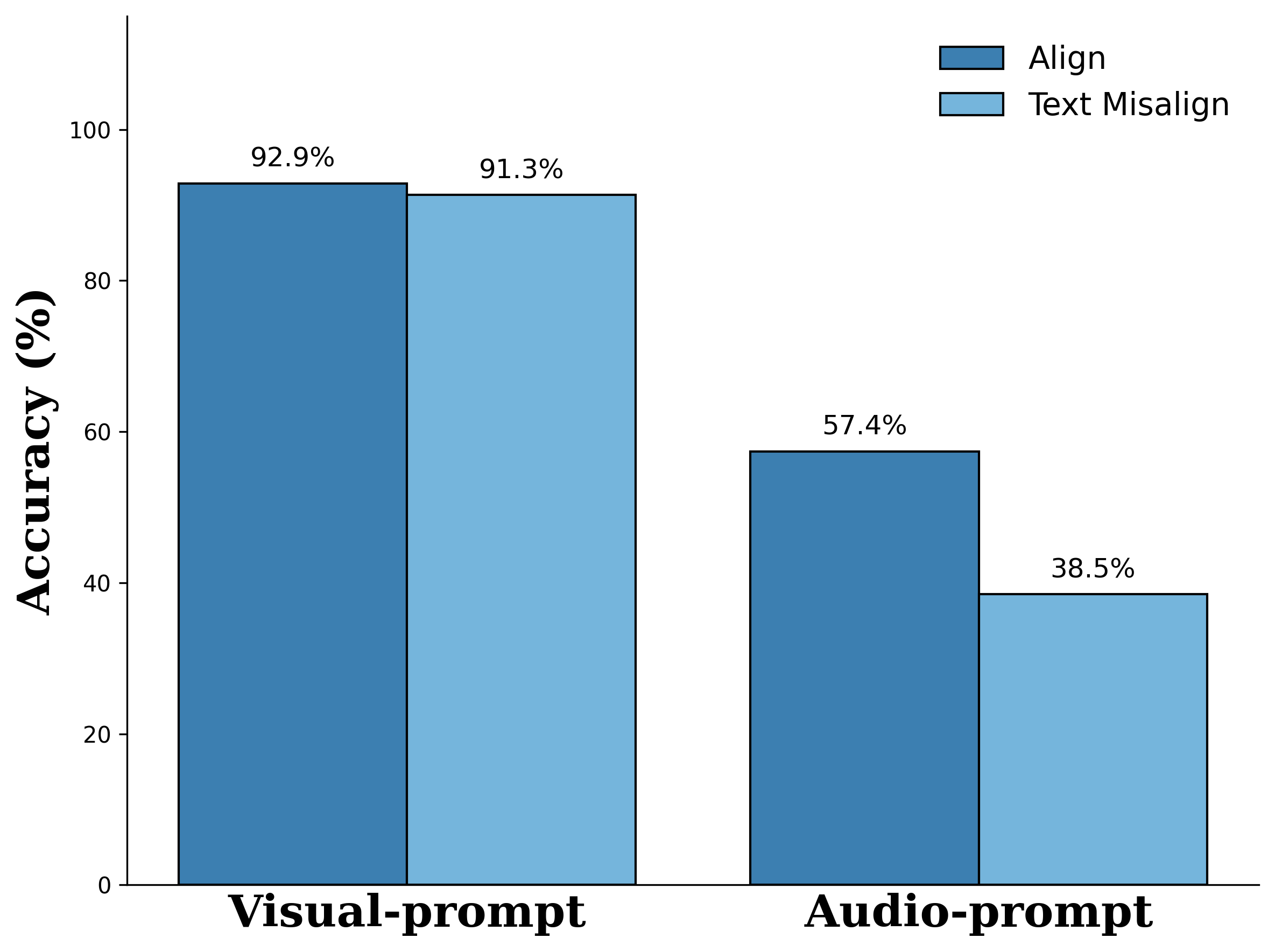}
        \caption{Qwen3-Omni-30B}
        \label{fig:supp_text_qwen3}
    \end{subfigure}
    \hfill
    \begin{subfigure}[t]{0.48\linewidth}
        \centering
        \includegraphics[width=\linewidth]{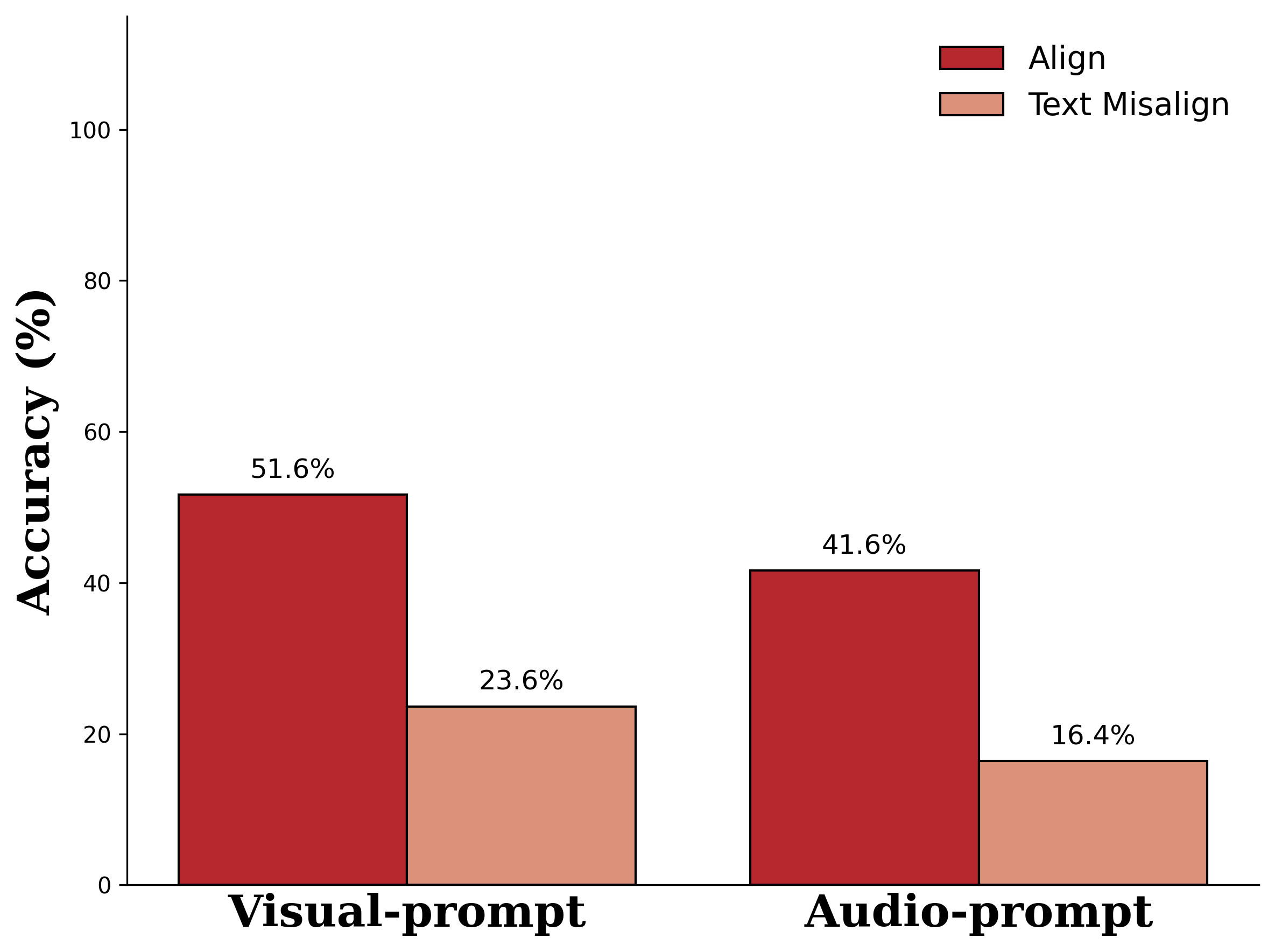}
        \caption{ChatBridge}
        \label{fig:supp_text_chatbridge}
    \end{subfigure}

    \vspace{1mm}
    \caption{
    \textbf{Impact of misleading textual context.} 
    Accuracy comparison when a contradictory caption is prepended to the query. 
    }
    \label{fig:supp_text_misalign}
    \vspace{-2mm}
\end{figure}
\subsection{Misalignment via text}
Using the misleading captions in Figure~\ref{fig:text_misalign_prompt}, we evaluate the baseline models under aligned and text-misaligned conditions (Fig.~\ref{fig:supp_text_misalign}). The results show a clear asymmetry: visual-prompt accuracy is comparatively robust for stronger models, whereas audio-prompt accuracy collapses across all architectures.

Under Visual prompt setting, Gemini-2.5-Pro and Qwen3-Omni-30B remain stable under text misalignment, dropping only $1.9\%$ and $0.6\%$ respectively. In contrast, Gemini-2.0-Flash and ChatBridge show pronounced degradation, falling $16.9\%$ and $29.9\%$. These sharp reductions indicate a higher susceptibility to linguistic interference, implying weaker decoupling between the visual stream and the textual conditioning.

Textual misalignment severely harms all models under Audio-focused prompt with drops as high as $22.7\%$. Even in the aligned setting, audio accuracies begin far below their visual counterparts, pointing to inherently weaker or noisier audio representations. When the caption contradicts the sound, the linguistic modality - typically the strongest and most trusted input channel for modern MLLMs - overrides the fragile auditory signal, resulting in severe degradation.
Overall, text misalignment disproportionately disrupts audio-based reasoning, while visual-based reasoning remains fairly robust only for the strongest models.

\begin{figure}[t]
    \centering
    \begin{subfigure}[t]{0.48\linewidth}
        \centering
        \includegraphics[width=\linewidth]{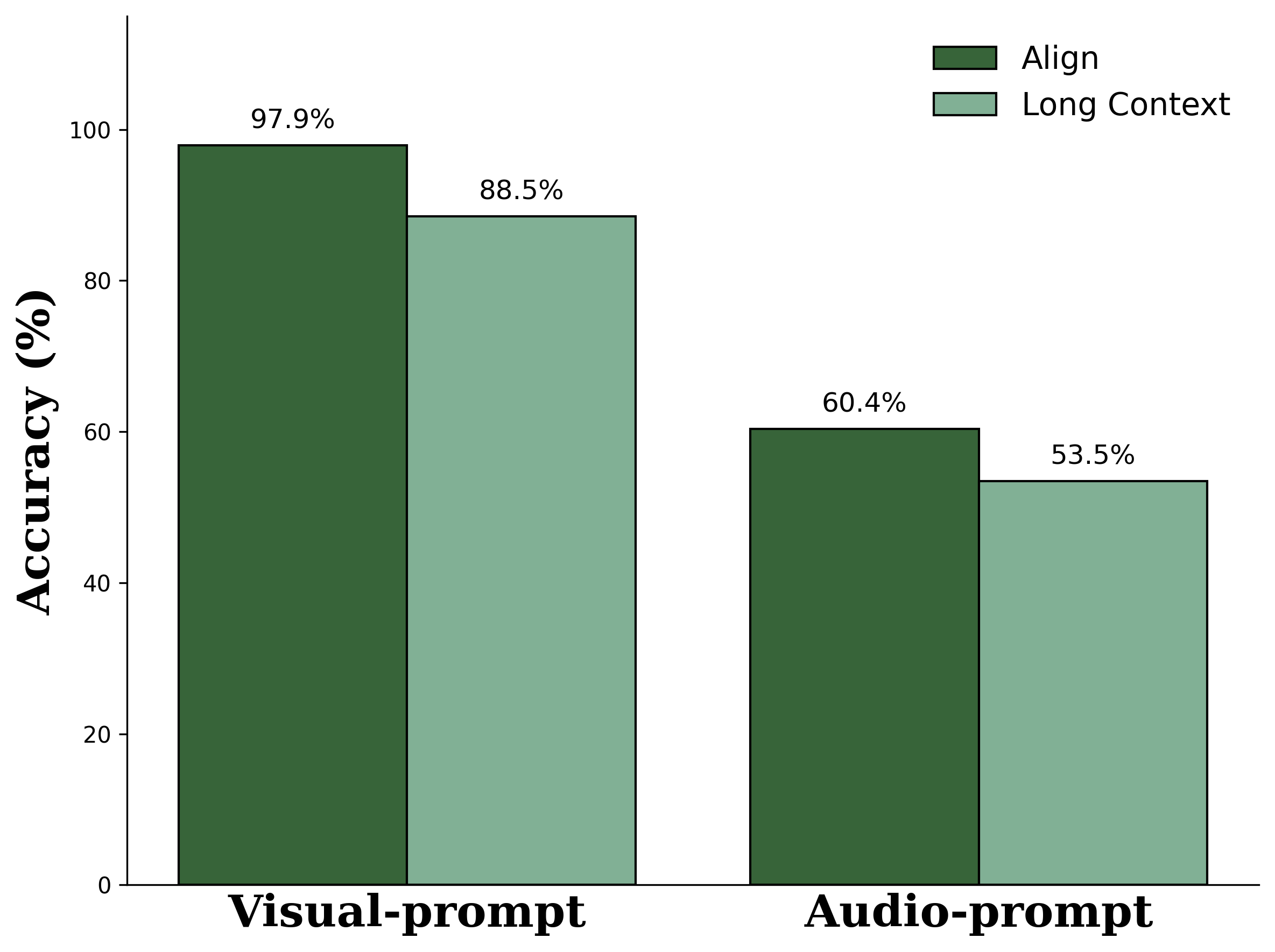}
        \caption{Gemini-2.5-Pro}
        \label{fig:supp_long_gemini15}
    \end{subfigure}
    \hfill
    \begin{subfigure}[t]{0.48\linewidth}
        \centering
        \includegraphics[width=\linewidth]{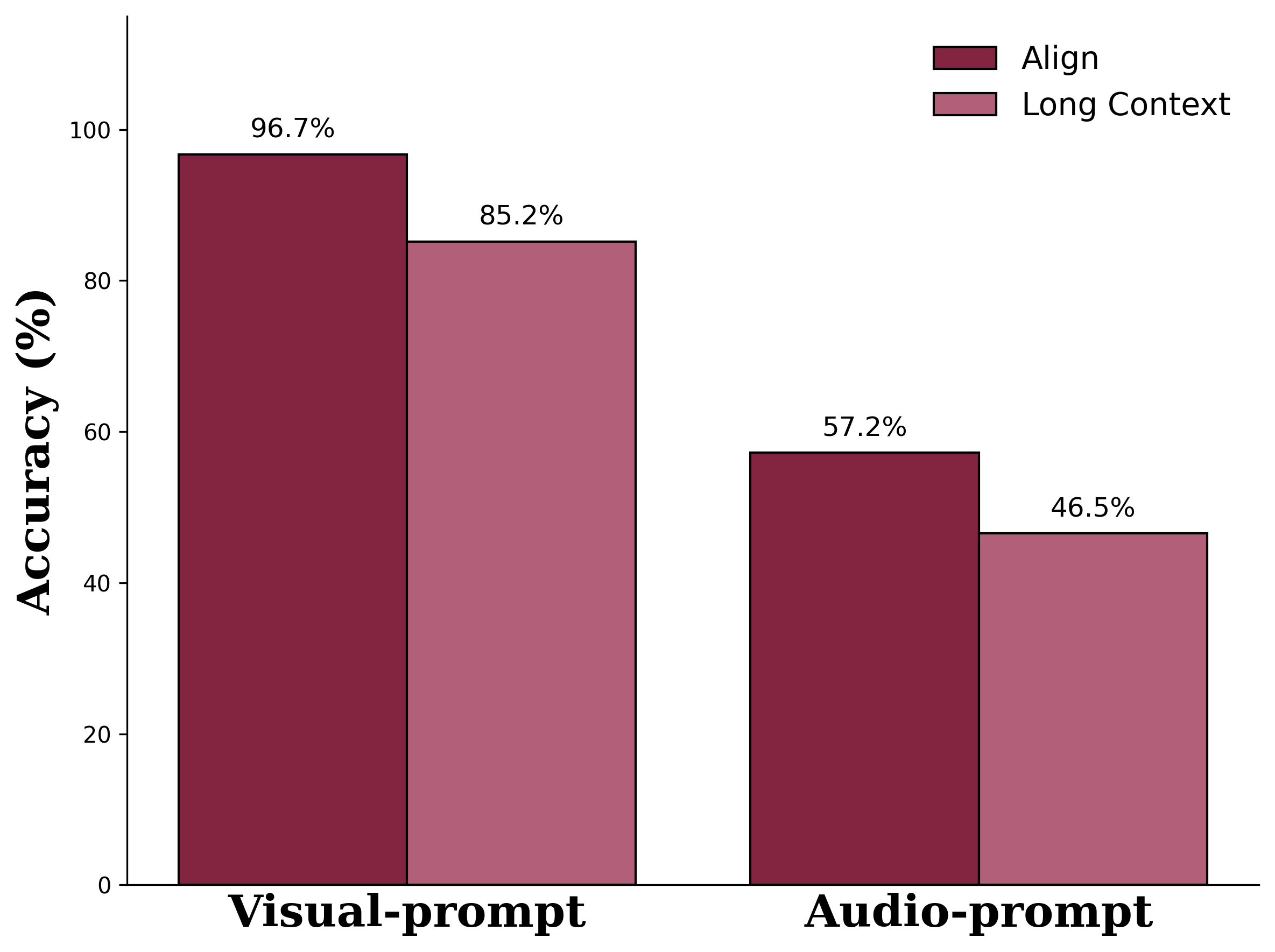}
        \caption{Gemini-2.0-Flash}
        \label{fig:supp_long_gemini2}
    \end{subfigure}

    \vspace{1mm}
    \begin{subfigure}[t]{0.48\linewidth}
        \centering
        \includegraphics[width=\linewidth]{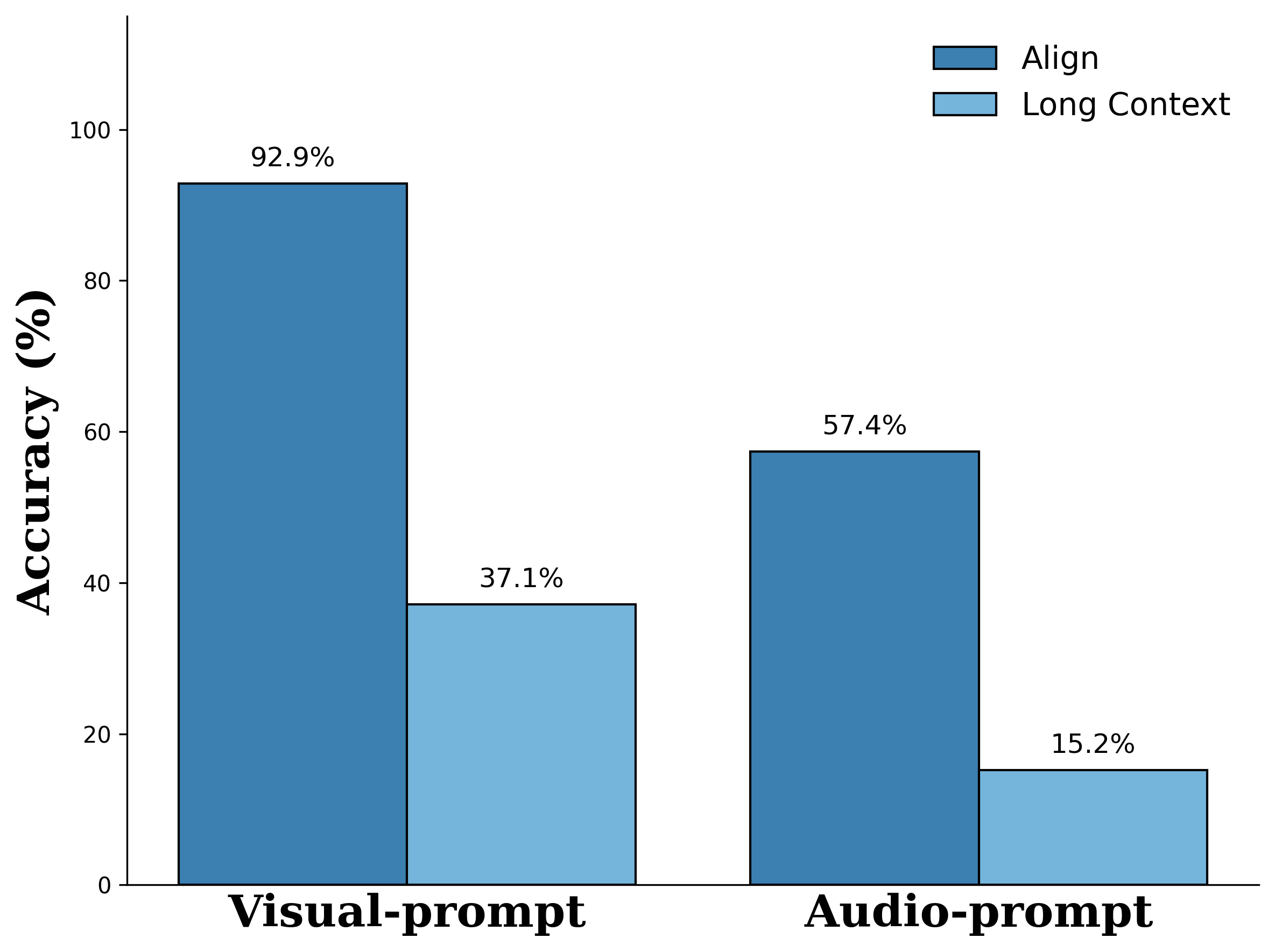}
        \caption{Qwen3-Omni-30B}
        \label{fig:supp_long_qwen3}
    \end{subfigure}
    \hfill
    \begin{subfigure}[t]{0.48\linewidth}
        \centering
        \includegraphics[width=\linewidth]{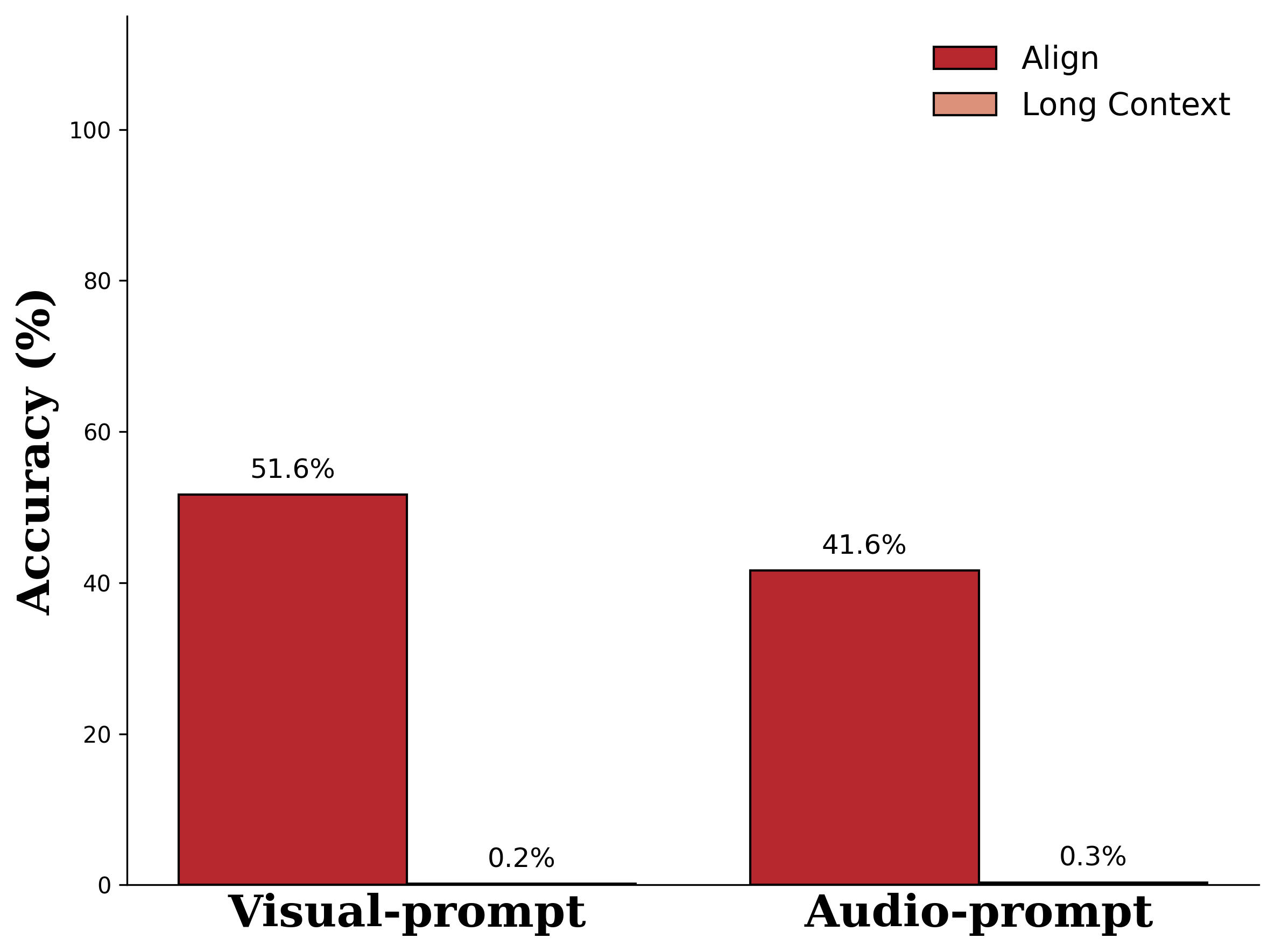}
        \caption{ChatBridge}
        \label{fig:supp_long_chatbridge}
    \end{subfigure}

    \vspace{1mm}
    \caption{
    \textbf{Performance under long-context interference.} 
    }
    \label{fig:supp_long_context}
    \vspace{-2mm}
\end{figure}
\subsection{Irrelevant long context caption}
To examine the robustness of different model families to long-range distractors, we prepend large amounts of irrelevant text (“garbage context”) far before the multimodal query and evaluate performance under aligned versus long-context conditions (Fig.~\ref{fig:supp_long_context}). The results reveal a clear divide between closed-source and open-source architectures in their ability to resist interference from distant textual noise. Gemini-2.5-Pro and Gemini-2.0-Flash show relatively mild degradation when exposed to long, irrelevant context. Although accuracy decreases across both prompt types, the drops remain moderate and the models preserve much of their original performance(the biggest drop among both the settings between both models is $10.4\%$). This suggests that these closed-source models maintain strong grounding mechanisms that prevent distant, semantically unrelated text from dominating the multimodal encoding. Their internal routing of cross-modal attention appears to effectively constrain where language context can influence downstream reasoning.

In contrast, Qwen3-Omni-30B and ChatBridge suffer severe declines in accuracy under long-context interference. Qwen experiences major reductions for both visual- and audio-prompt settings, and ChatBridge undergoes complete collapse in both cases. These sharp failures indicate that open-source models lack robust long-range filtering: irrelevant early-context is able to hijack the model’s hidden representations, overwhelming the true multimodal signal. The vulnerability is especially pronounced for audio-conditioned inference, where the already weaker audio representations are easily overwritten by distractor text.

Taken together, these results demonstrate that closed-source architectures exhibit significantly stronger resilience to irrelevant long-context noise, while open-source models are far more susceptible to interference, especially when relying on fragile auditory cues.




\begin{table*}[h]
    \centering
    \small
    \resizebox{\linewidth}{!}{
    \begin{tabular}{l|cc|c|cc|c}
    \toprule
    & \multicolumn{3}{c|}{\textbf{Visual Prompt}} & \multicolumn{3}{c}{\textbf{Audio Prompt}} \\
    & \multicolumn{2}{c|}{\textit{Standard}} & \multicolumn{1}{c|}{\textit{Abstention Test}} & \multicolumn{2}{c|}{\textit{Standard}} & \multicolumn{1}{c}{\textit{Abstention Test}} \\
    \textbf{Model} & \textbf{Align} & \textbf{Audio Removed} & \textbf{Frames Zeroed} & \textbf{Align} & \textbf{Frames Zeroed} & \textbf{Audio Removed} \\
    \midrule
    Gemini-1.5-Pro & 97.90 & 95.28 & 47.42 & 60.37 & 24.95 & 3.79 \\
    Gemini-2.0-Flash & 96.71 & 91.91 & 3.04 & 57.21 & 9.42 & 1.06 \\
    Qwen3-Omni-30B & 92.88 & 83.73 & 15.05 & 57.39 & 14.58 & 11.71 \\
    {\qwenSB} & 76.68 & 58.72 & 10.94 & 46.60 & 25.16 & 9.86 \\
    ChatBridge & 51.64 & 54.71 & 55.77 & 41.61 & 7.07 & 37.69 \\
    PandaGPT & 28.75 & 29.79 & 11.25 & 13.12 & 1.18 & 0.61 \\
    \midrule
    \rowcolor{gray!15} \textbf{{\qwenSB} + Ours} & 94.68 & 94.37 & \textbf{90.27} & 88.14 & 79.79 & 0.00 \\
    \bottomrule
    \end{tabular}
    }
    \caption{\textbf{Zero-Shot Abstention Performance.} We add ``None of the above'' to the options and remove one modality (Frame Zeroed or Audio Removed). A high score indicates the model correctly abstains from answering when the data is missing. Our fine-tuned model (bottom row) shows exceptional zero-shot abstention in the visual domain (90.27\%), proving it no longer hallucinates visual answers from audio cues.}
    \label{tab:abstention_test}
\end{table*}
\section{Unimodal Abstention Evaluation: The ``None of the Above'' Experiment}
\label{sec:abstention}
In the unimodal study presented in the main paper (Section 4.1.2), models were forced to select a class from a predefined list even when the relevant modality was removed (e.g., answering ``Which class best describes the visual content of this video?'' given a black video). This setting is inherently ill-posed, as it forces the model to either guess randomly or hallucinate based on the remaining modality.

To provide a more rigorous evaluation of modality dependence, we introduce a Zero-Shot Abstention Test. Inspired by the methodology in AVTrustBench~\cite{chowdhury2025avtrustbench}, we append the option ``None of the above'' to the candidate list. In this setting, if a model is asked to describe the visual content of a black frame, the only correct behavior is to reject the semantic classes and select ``None of the above.'' A failure to do so indicates that the model is hallucinating information from the remaining modality (e.g., ``hearing'' the visual content). We evaluate all baselines and our alignment-tuned model in this setting. \textit{Importantly, our fine-tuned model was never exposed to ``None of the above'' labels or unimodal data during training, making this a zero-shot stability test.} The results are summarized in Table.~\ref{tab:abstention_test}.

We observe a strong resistance to abstention across SOTA baselines. Gemini-2.0-Flash, for example, records near-zero accuracy in missing-modality settings (3.04\% for Frames Zeroed, 1.06\% for Audio Removed). Instead of signaling ignorance, the model forces a prediction based on the single available modality in $>95\%$ of trials. 
\textbf{Our fine-tuned model ({\qwenSB}+Ours) demonstrates a remarkable emergence of abstention capability in the visual domain. }In the Frames Zeroed setting, our model achieves an accuracy of 90.27\%, significantly outperforming the base {\qwenSB} (10.94\%) and the 30B-parameter Qwen3-Omni (15.05\%). 
This indicates that our modality-aware fine-tuning successfully taught the model that visual questions require visual evidence. By learning to distinguish between aligned and misaligned pairs during training, the model effectively learned to disregard audio cues when performing visual reasoning. Consequently, when visual evidence is absent (black frames), it refuses to let the audio track dictate the visual answer, correctly defaulting to ``None of the above.''

On the other hand, in the Audio Removed setting, our model scores 0\%, similar to several baselines. This suggests that while we successfully blocked the Audio $\rightarrow$ Visual leakage, the Visual $\rightarrow$ Audio shortcut remains strong. When the audio is silent, the model likely still perceives the visible object (e.g., a dog) and is compelled to predict the associated sound (e.g., ``Bark''), illustrating the extreme difficulty of overcoming visual dominance in MLLMs. 

\section{Can Reasoning Traces Fix Misalignment? (Chain-of-Thought Evaluation)} 
\label{app:cot}
\begin{figure}[t]
    \centering
    \fcolorbox{purple}{promptcolor!10}{
    \parbox{\dimexpr0.9\linewidth-2\fboxsep-2\fboxrule}{
        \scriptsize 
        \noindent\textbf{Visual-focused prompt:} 
        \begin{quote}
        \texttt{"Think step by step about what the visual content shows. First provide your thinking process, then give the final answer in the format: Final Answer: <class>. "}
        \end{quote}
        \vspace{-1mm}
        \noindent\textbf{Audio-focused prompt:} 
        \begin{quote}
        \texttt{"Think step by step about what the audio content shows. First provide your thinking process, then give the final answer in the format: Final Answer: <class>. "}
        \end{quote}
        \vspace{-1mm}
    }}
    \caption{\textbf{Chain-of-Thought (CoT) Prompting Strategy.} We explicitly instruct the model to articulate its thinking process before providing the final classification, aiming to force a logical separation of modalities.}
    \label{fig:cot_prompt}
    \vspace{-2mm}
\end{figure}

\begin{table}[h]
    \centering
    \small
    \setlength{\tabcolsep}{4pt}
    \resizebox{\linewidth}{!}{
    \begin{tabular}{l|cc|cc}
    \toprule
    & \multicolumn{2}{c|}{\textbf{Visual Prompt (\%)}} & \multicolumn{2}{c}{\textbf{Audio Prompt (\%)}} \\
    \textbf{Method} & \textbf{Align} & \textbf{Misalign} & \textbf{Align} & \textbf{Misalign} \\
    \midrule
     {\qwenSB} (Standard) & 76.68 & 58.72 & 46.60 & 25.16 \\
     {\qwenSB} + CoT & 66.18 & 53.25 & 45.14 & 31.16 \\
    \midrule
    \rowcolor{gray!15} \textbf{{\qwenSB} + Ours (Standard)} & \textbf{94.68} & \textbf{94.37} & \textbf{88.14} & \textbf{79.79} \\
    {\qwenSB} + Ours + CoT & 94.71 & 94.66 & 58.46 & 55.14 \\
    \bottomrule
    \end{tabular}
    }
    \caption{\textbf{Impact of Chain-of-Thought (CoT) Prompting.} Comparing standard inference vs. CoT. While CoT provides marginal gains for the base model in specific niches, it degrades the robust audio grounding of our fine-tuned model, likely by re-introducing visual priors during the reasoning generation step.}
    \label{tab:cot_results}
\end{table}
Recent literature~\cite{wang2025multimodal} suggests that prompting MLLMs to ``think step-by-step'' (Chain-of-Thought or CoT) can potentially strengthen reasoning capabilities and improve interpretability. To investigate whether inference-time reasoning can resolve sensory conflict without parameter updates, we evaluated both the base Qwen2.5-Omni-7B and our alignment-tuned variant using the CoT prompt structure illustrated in Figure~\ref{fig:cot_prompt}.
The results, summarized in Table~\ref{tab:cot_results}, reveal two counter-intuitive findings that challenge the assumption that CoT is beneficial for multimodal misalignment.
Contrary to the expectation that reasoning traces would filter noise, applying CoT to the base Qwen model resulted in a performance regression on visual tasks (Visual Base: $76.68\% \rightarrow 66.18\%$). This aligns with recent findings that encouraging the model to ``think'' might not always help~\cite{lithink}. In our misalignment setting, the model likely uses the reasoning steps to describe the conflicting audio or visual priors, effectively confusing itself rather than resolving the conflict.

The most striking result is the impact of CoT on our fine-tuned model ({\qwenSB}+Ours). While visual performance remains stable, auditory performance collapses (Audio prompt with aligned samples: $88.14\% \rightarrow 58.46\%$). 
We hypothesize that 
during the ``thinking'' phase, the model likely defaults to describing the dominant visual input (visual dominance), which re-contaminates the context window and overrides the learned audio alignment. This confirms that robust modality alignment requires intrinsic parameter optimization rather than extrinsic prompt engineering, and that CoT might not help with perceptual grounding.

\section{Qualitative Analysis of Improved Modality Grounding}
\label{app:qualitative}
\begin{figure*}[t]
    \centering
    \setlength{\tabcolsep}{1pt} 
    
    \begin{subfigure}[b]{0.49\linewidth}
        \centering
        \includegraphics[width=\linewidth]{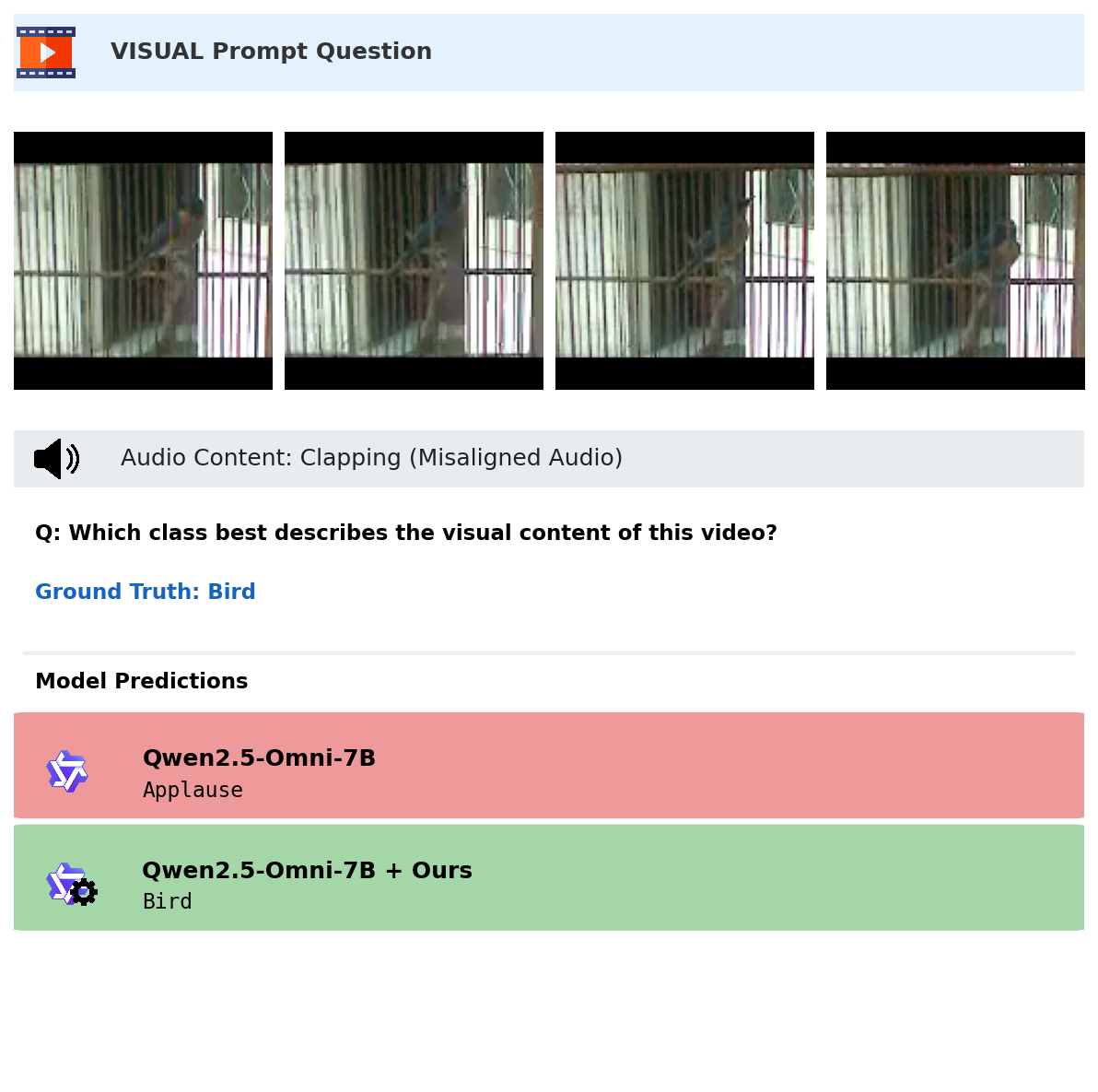}
    \end{subfigure}
    \hfill
    \begin{subfigure}[b]{0.49\linewidth}
        \centering
        \includegraphics[width=\linewidth]{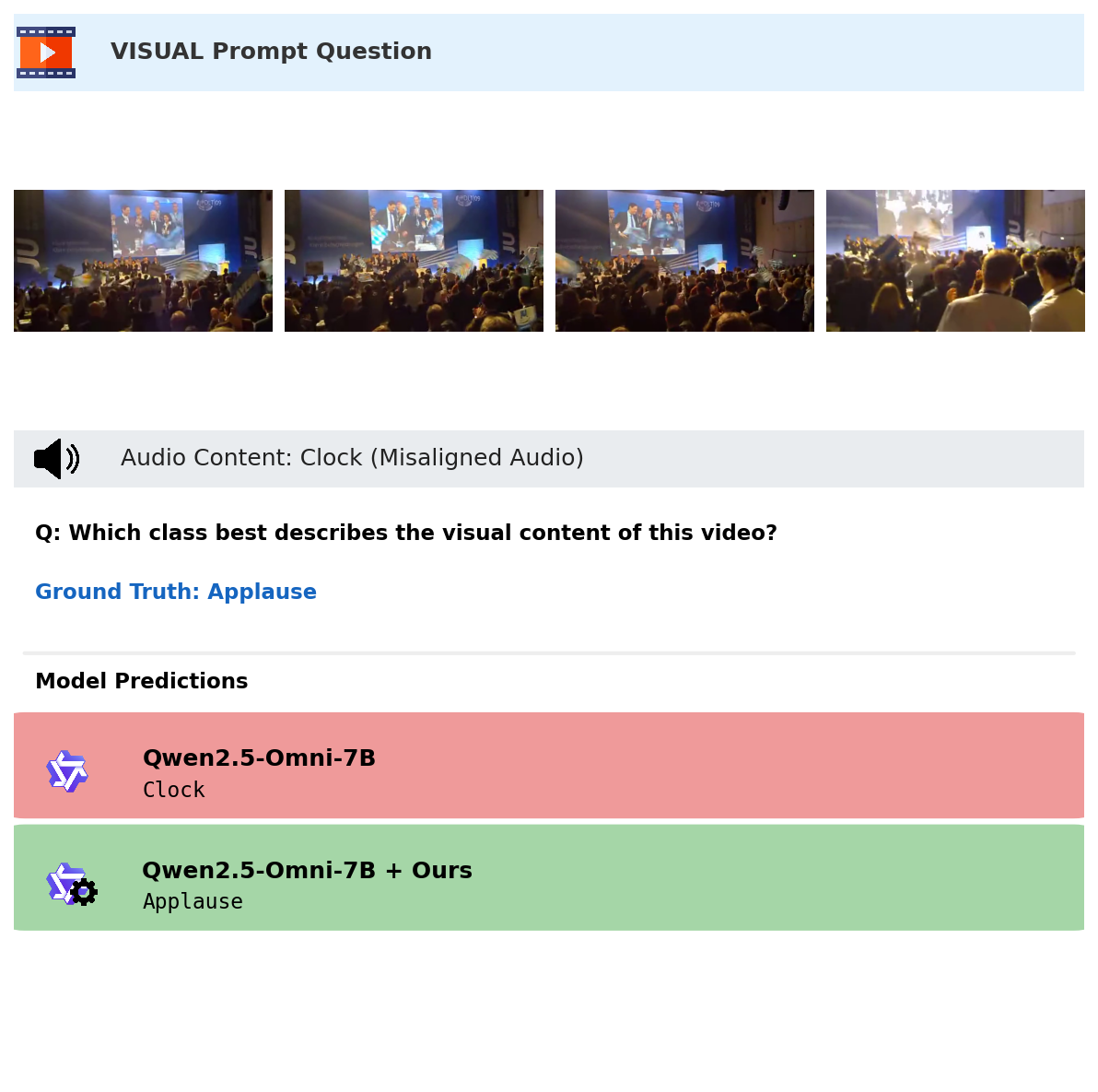}
    \end{subfigure}
    
    \vspace{3mm} 

    \begin{subfigure}[b]{0.49\linewidth}
        \centering
        \includegraphics[width=\linewidth]{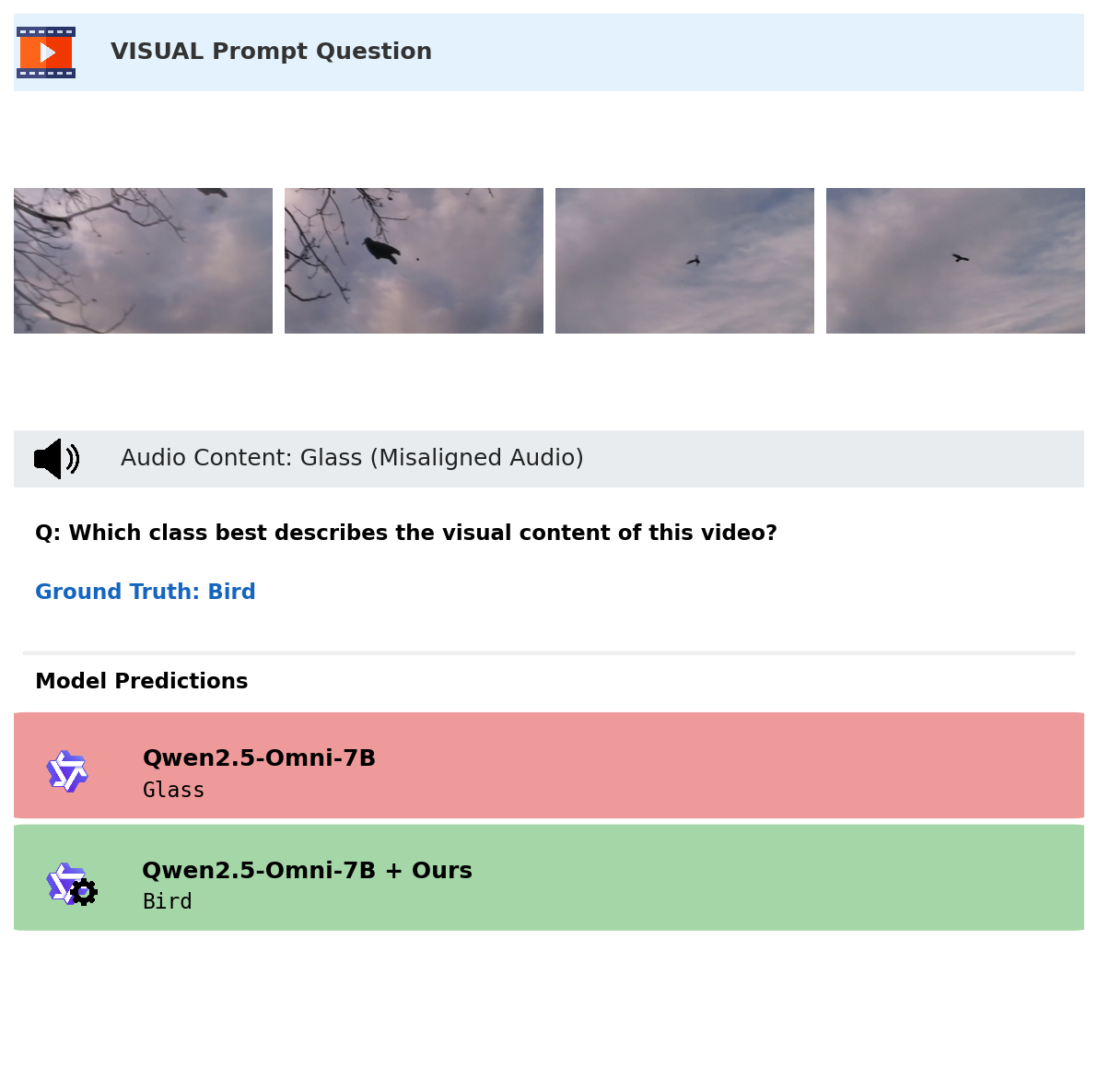}
    \end{subfigure}
    \hfill
    \begin{subfigure}[b]{0.49\linewidth}
        \centering
        \includegraphics[width=\linewidth]{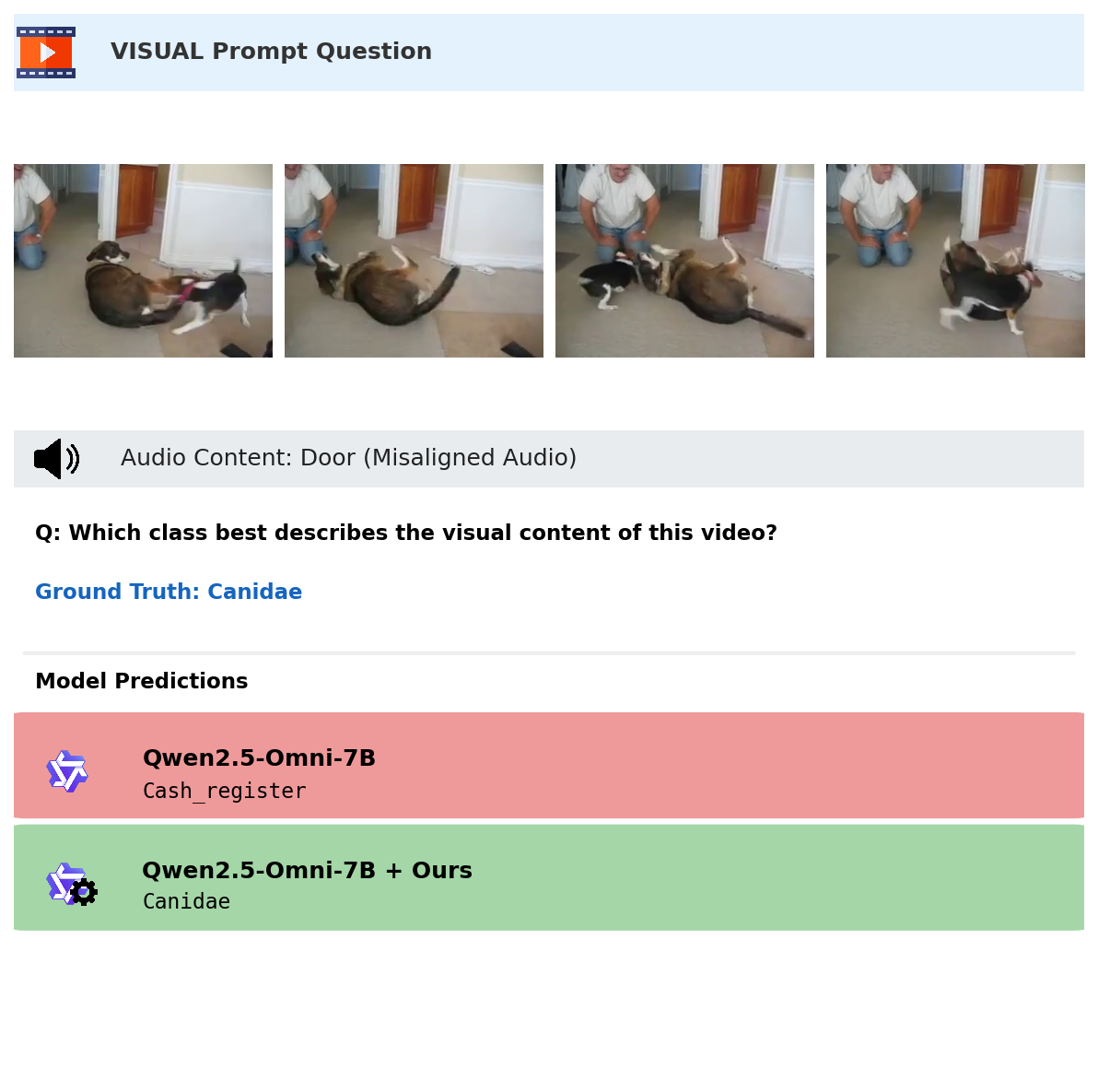}
    \end{subfigure}
    
    \vspace{3mm} 
    

    \caption{\textbf{Qualitative Results Gallery.} Red indicates incorrect baseline predictions, Green indicates our correct predictions. While the baseline consistently suffers from hallucinations driven by conflicting modalities, our model demonstrates robust grounding in the requested sensory input.}
    \label{fig:qual_gallery}
\end{figure*}

\begin{figure*}[t]
    \centering
    \setlength{\tabcolsep}{1pt} 
    
    \begin{subfigure}[b]{0.49\linewidth}
        \centering
        \includegraphics[width=\linewidth]{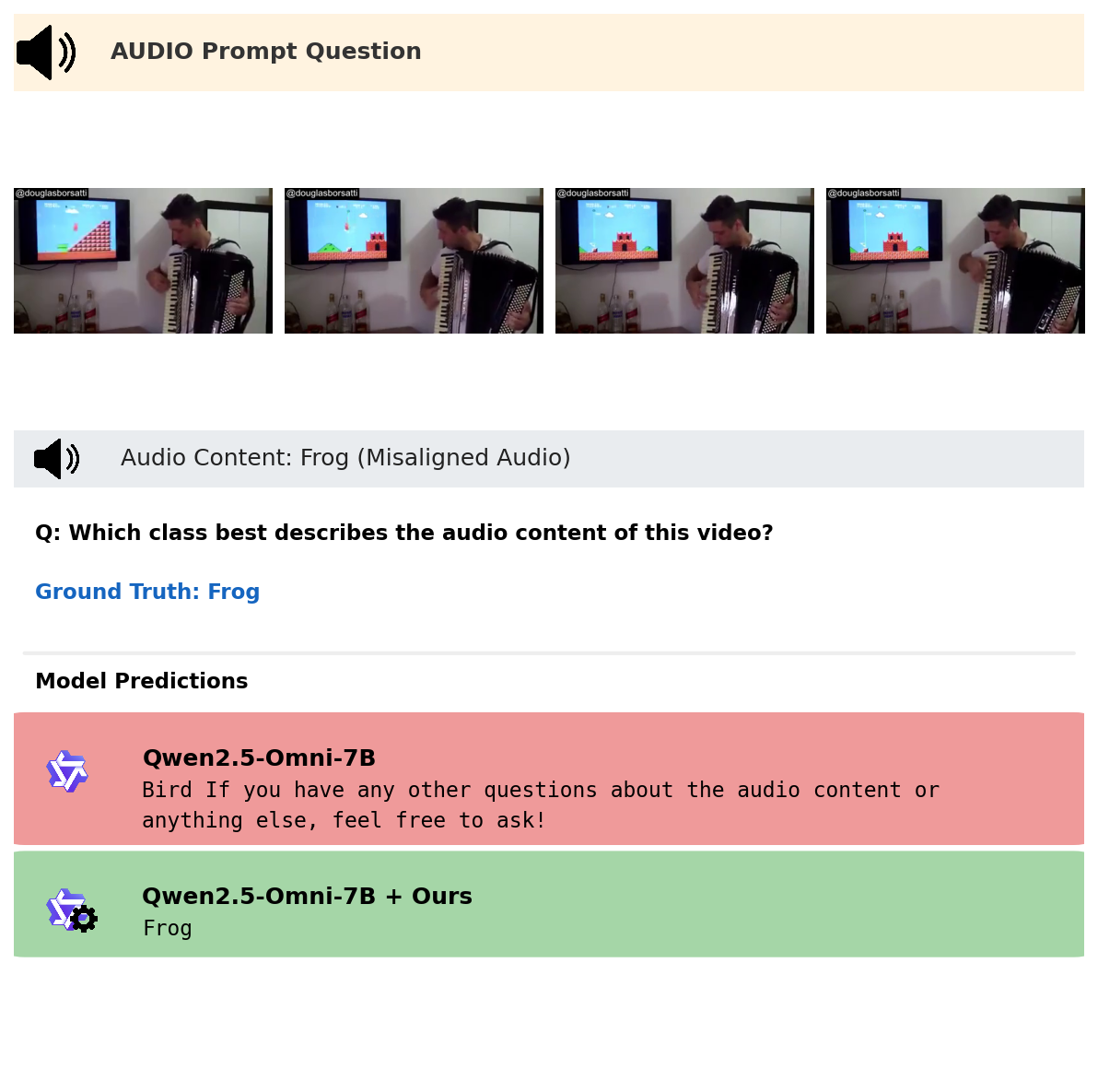}
    \end{subfigure}
    \hfill
    \begin{subfigure}[b]{0.49\linewidth}
        \centering
        \includegraphics[width=\linewidth]{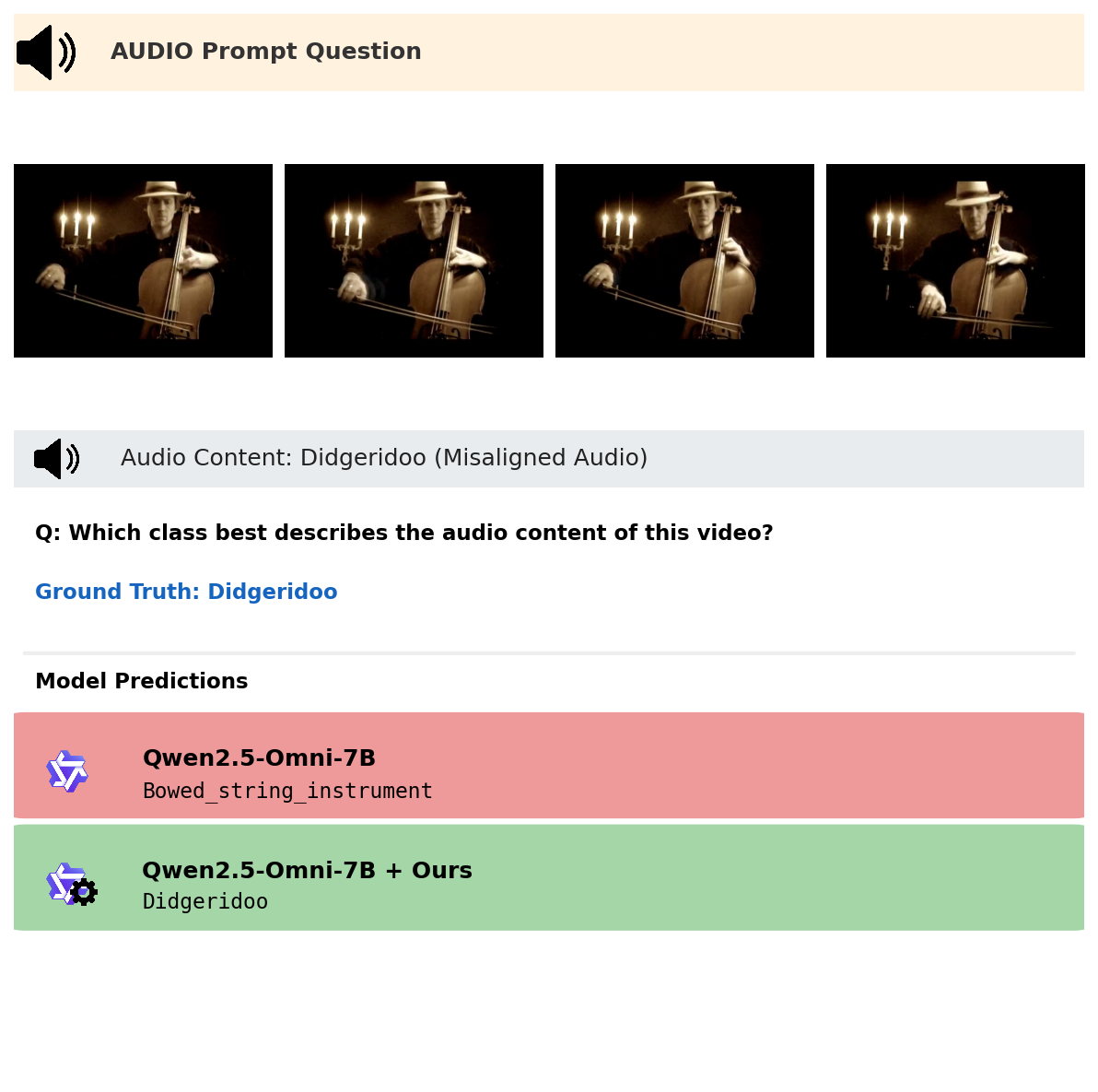}
    \end{subfigure}
    
    \vspace{3mm} 

    \begin{subfigure}[b]{0.49\linewidth}
        \centering
        \includegraphics[width=\linewidth]{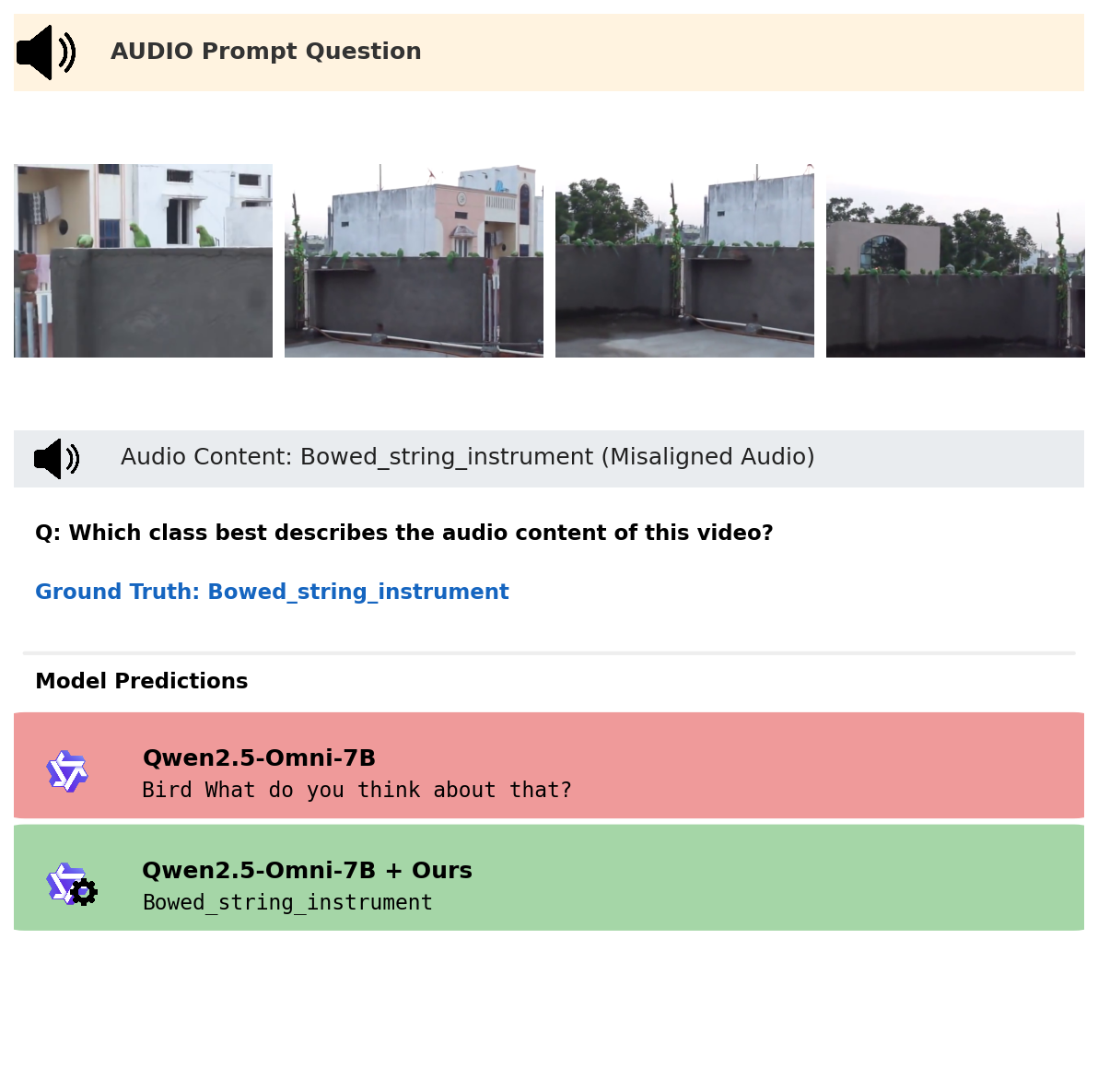}
    \end{subfigure}
    \hfill
    \begin{subfigure}[b]{0.49\linewidth}
        \centering
        \includegraphics[width=\linewidth]{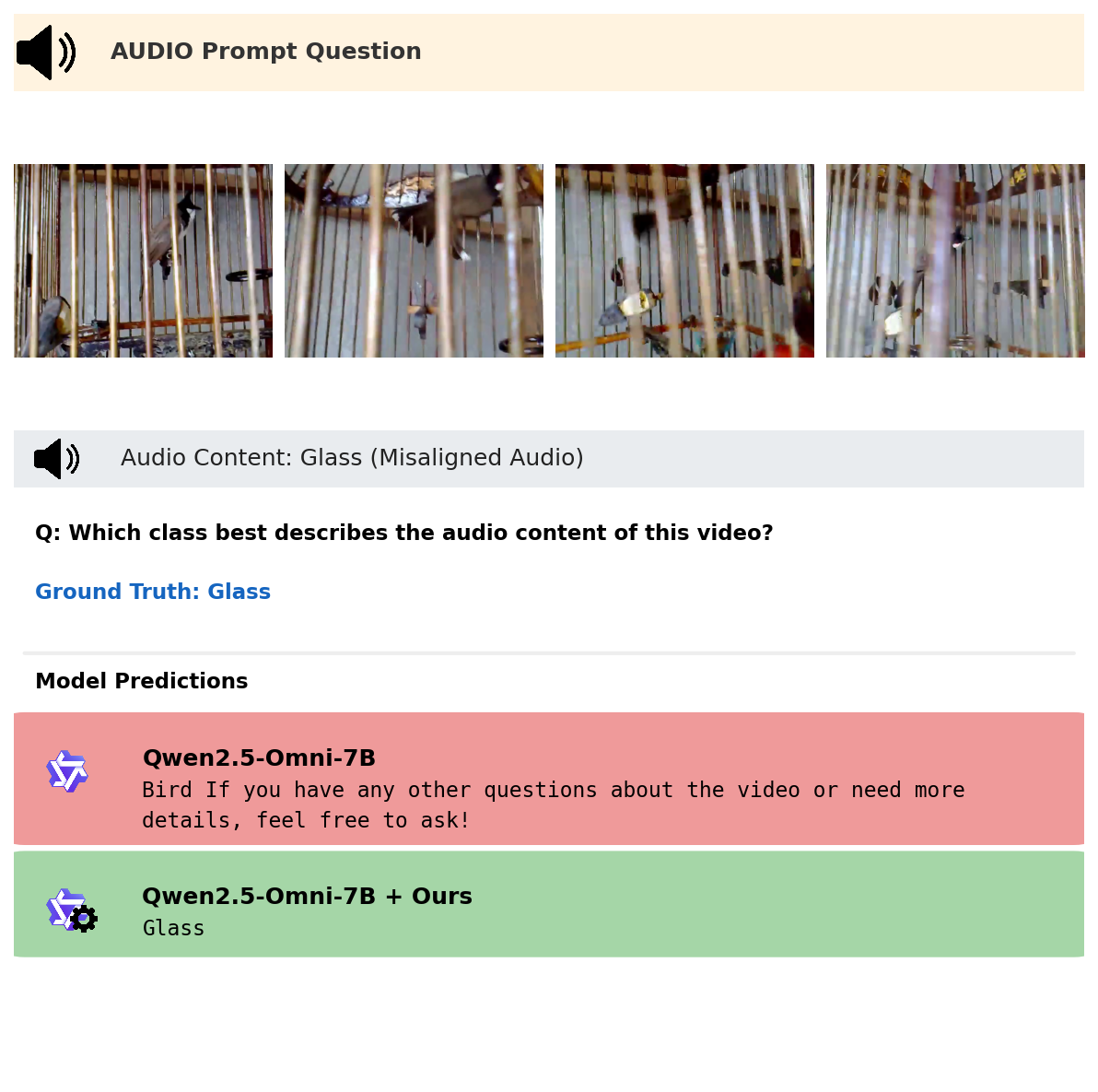}
    \end{subfigure}
    
    \vspace{3mm} 
    

    \caption{\textbf{Qualitative Results Gallery.} Red indicates incorrect baseline predictions, Green indicates our correct predictions. While the baseline consistently suffers from hallucinations driven by conflicting modalities, our model demonstrates robust grounding in the requested sensory input.}
    \label{fig:qual_gallery2}
\end{figure*}

\begin{figure*}[t]
    \centering
    \setlength{\tabcolsep}{1pt} 
    
    \begin{subfigure}[b]{0.49\linewidth}
        \centering
        \includegraphics[width=\linewidth]{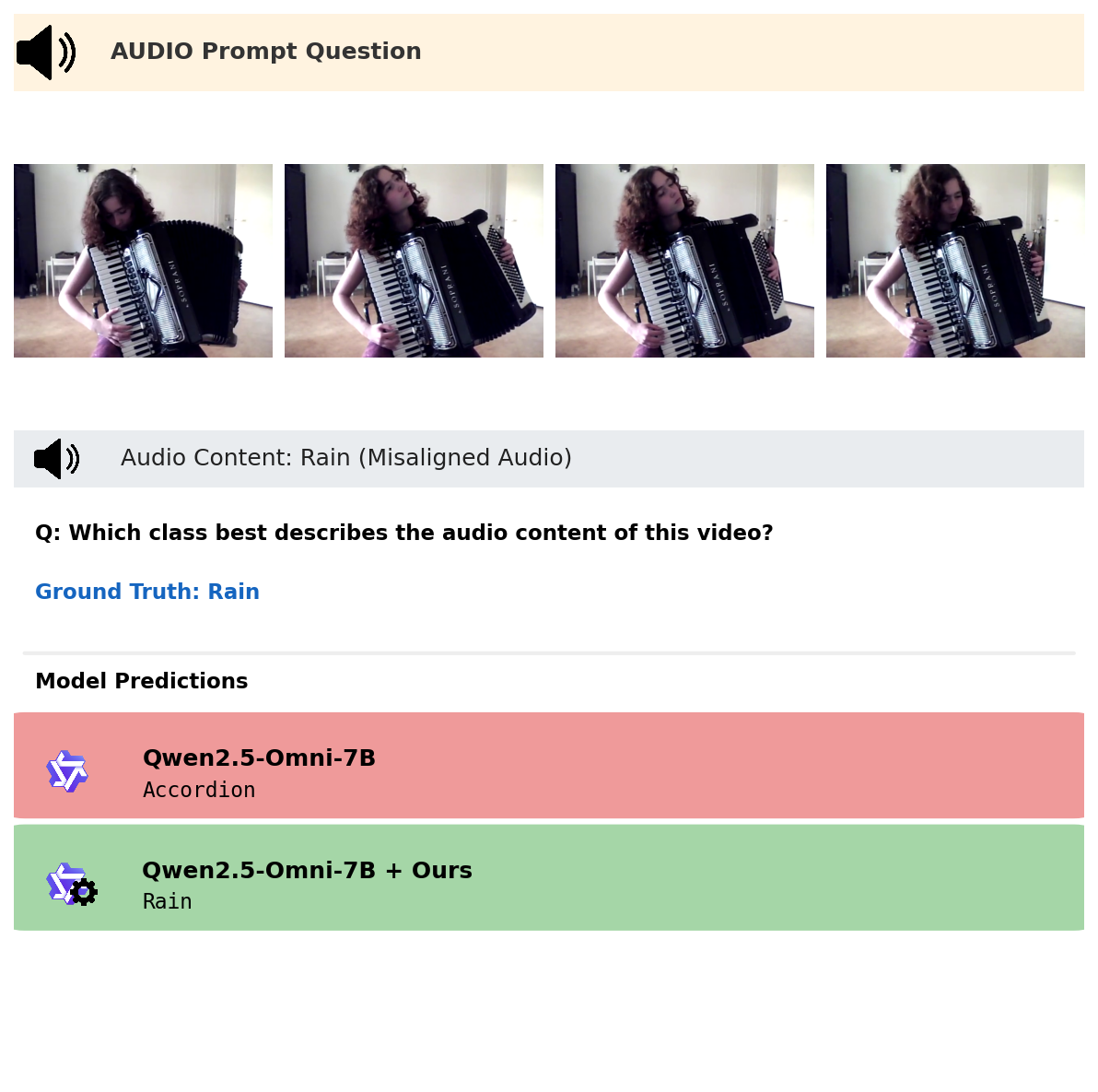}
    \end{subfigure}
    \hfill
    \begin{subfigure}[b]{0.49\linewidth}
        \centering
        \includegraphics[width=\linewidth]{images/quality/190.png}
    \end{subfigure}
    
    \vspace{3mm} 

    \begin{subfigure}[b]{0.49\linewidth}
        \centering
        \includegraphics[width=\linewidth]{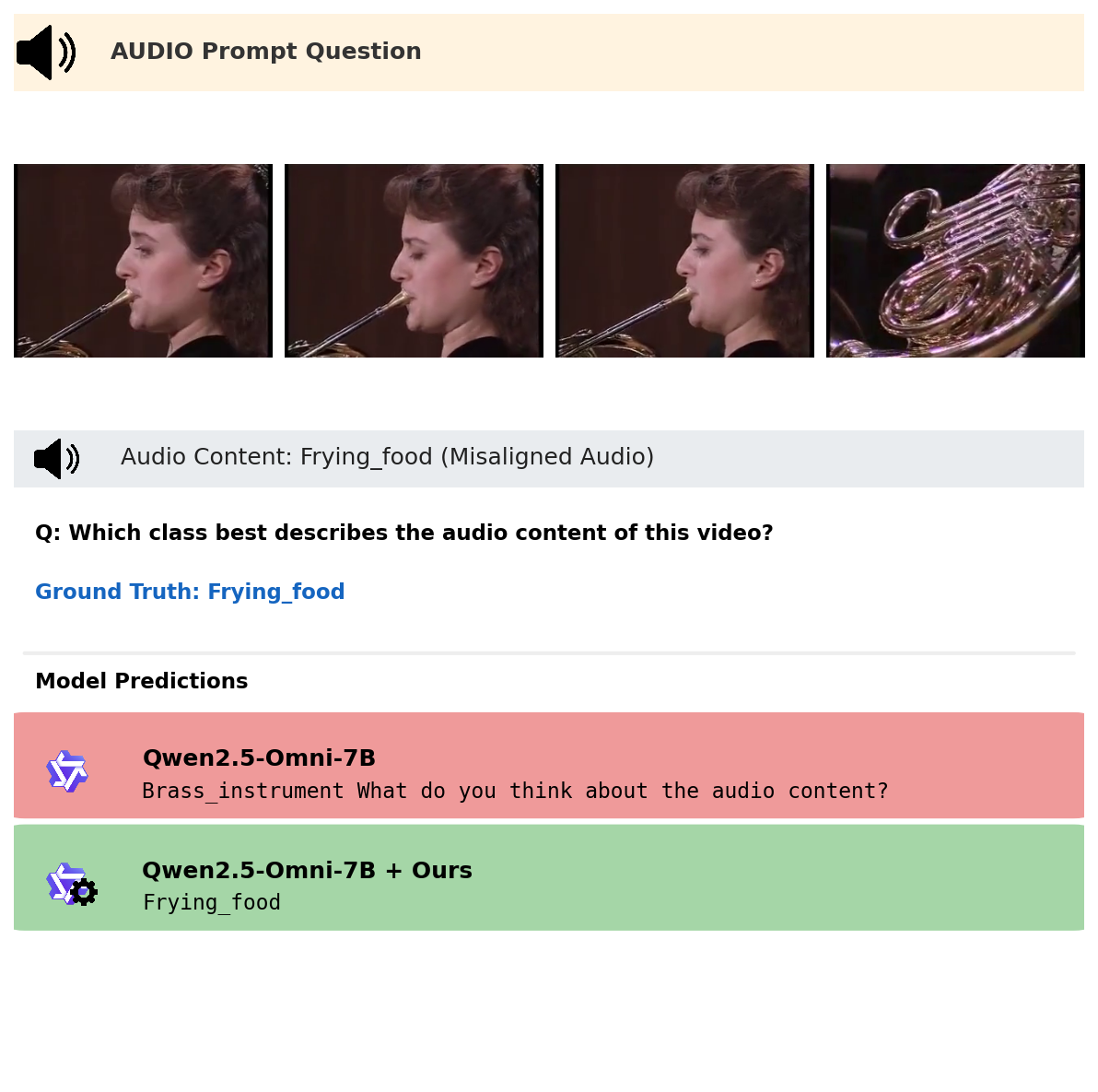}
    \end{subfigure}
    \hfill
    \begin{subfigure}[b]{0.49\linewidth}
        \centering
        \includegraphics[width=\linewidth]{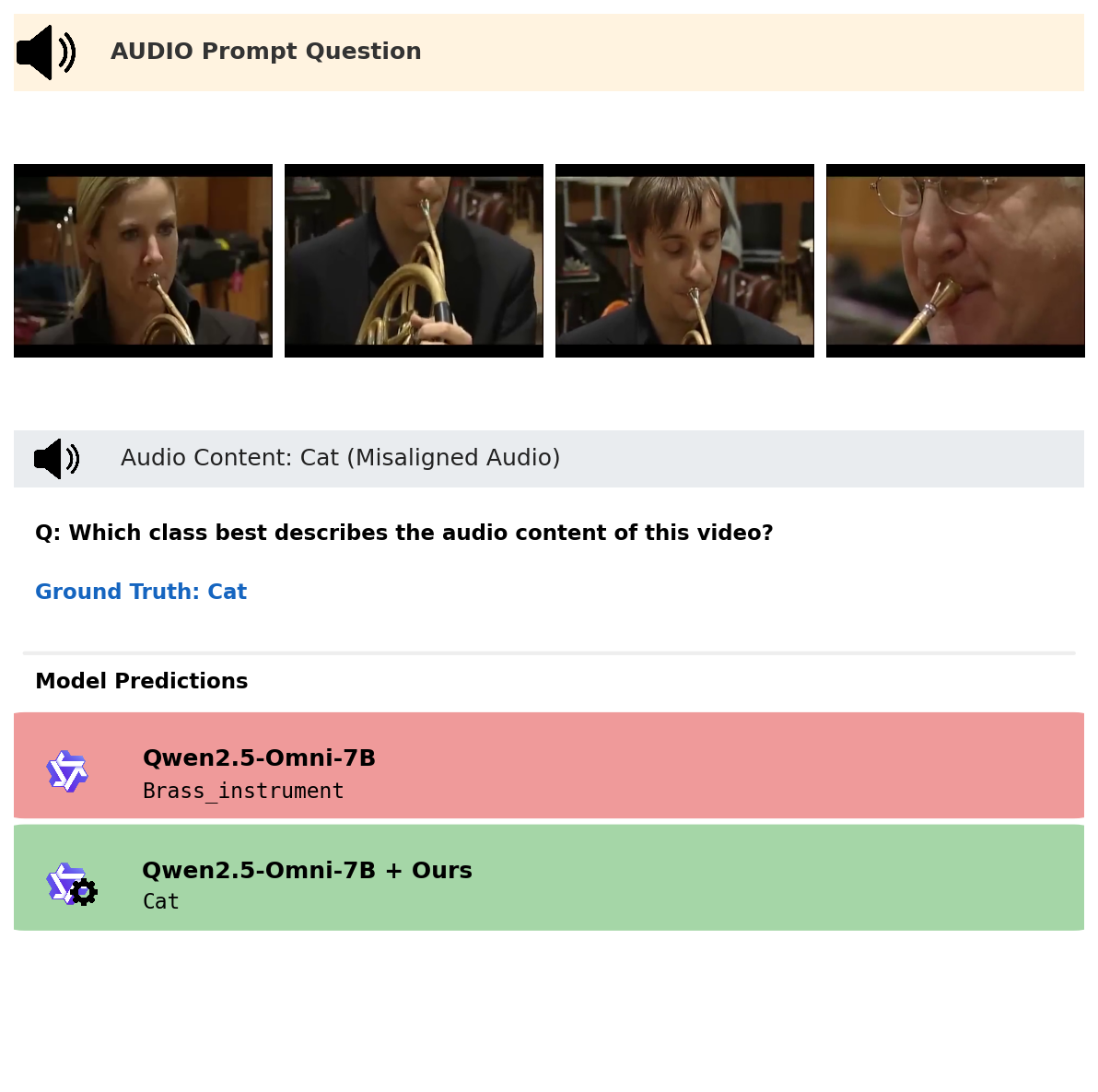}
    \end{subfigure}
    
    \vspace{3mm} 
    

    \caption{\textbf{Qualitative Results Gallery.} Red indicates incorrect baseline predictions, Green indicates our correct predictions. While the baseline consistently suffers from hallucinations driven by conflicting modalities, our model demonstrates robust grounding in the requested sensory input.}
    \label{fig:qual_gallery3}
\end{figure*}

We present a selection of qualitative examples in Figure ~\ref{fig:qual_gallery}--~\ref{fig:qual_gallery3} to visualize the practical improvement of alignment-aware fine-tuning compared to standard baselines. In scenarios characterized by semantic misalignment, such as a video depicting a dog paired with the sound of a phone ringing, baseline models frequently succumb to visual dominance and incorrectly predict a barking sound. Our model successfully decouples these conflicting sensory streams, correctly attending to the auditory signal despite the visual contradiction. Furthermore, the improved model demonstrates significant resilience to textual interference. When provided with misleading captions that contradict the actual sensory content, our model ignores the linguistic hallucination and grounds its response in the verified audio-visual evidence, confirming that the training process effectively reduces the over-reliance on priors from both the visual and textual modalities. 

\clearpage
\clearpage
\end{document}